\pdfoutput=1

\documentclass[11pt]{article}

\usepackage[final]{acl}

\usepackage{times}
\usepackage{latexsym}

\usepackage[T1]{fontenc}
\usepackage[utf8]{inputenc}
\usepackage[russian,english]{babel}

\usepackage{amsmath}
\usepackage{microtype}

\usepackage{inconsolata}

\usepackage{graphicx}
\usepackage{tcolorbox}

\usepackage{mdframed}
\usepackage{enumitem}
\mdfsetup{nobreak=true} 
\usepackage{xspace}
\usepackage{wrapfig}
\usepackage{booktabs}
\usepackage{float}
\usepackage{listings} 
\usepackage{soul}
\usepackage{multirow}
\usepackage{tabularx}
\usepackage{array}
\usepackage{makecell}
\usepackage{longtable}
\usepackage{adjustbox}
\usepackage{pifont}
\usepackage{algorithm}
\usepackage{algpseudocode}
\usepackage{manyfoot}
\usepackage{hyperref}

\usepackage{listings}
\usepackage{xcolor}
\definecolor{codebg}{HTML}{F7F7F7}
\lstdefinestyle{myyaml}{
  basicstyle=\ttfamily\small,
  backgroundcolor=\color{codebg},
  frame=leftline,               
  rulecolor=\color{blue},
  numbers=none,
  breaklines=true,
  showstringspaces=false
}

\newcolumntype{T}[1]{>{\raggedright\arraybackslash\begin{minipage}[t]{#1}}p{#1}<{\end{minipage}}}

\addto\extrasenglish{%
}

\newcommand{\meramulti}{MERA Multi\xspace}

\newcommand{\aquaria}{\hyperref[sec:dataset_aquaria]{AQUARIA}\xspace}
\newcommand{\commonvideoqa}{\hyperref[sec:dataset_commonvideoqa]{CommonVideoQA}\xspace}
\newcommand{\labtabvqa}{\hyperref[sec:dataset_labtabvqa]{LabTabVQA}\xspace}
\newcommand{\realvqa}{\hyperref[sec:dataset_realvqa]{RealVQA}\xspace}
\newcommand{\realvideoqa}{\hyperref[sec:dataset_realvideoqa]{RealVideoQA}\xspace}
\newcommand{\schoolsciencevqa}{\hyperref[sec:dataset_schoolsciencevqa]{SchoolScienceVQA}\xspace}
\newcommand{\unisciencevqa}{\hyperref[sec:dataset_unisciencevqa]{UniScienceVQA}\xspace}
\newcommand{\weird}{\hyperref[sec:dataset_weird]{WEIRD}\xspace}
\newcommand{\ruclevr}{\hyperref[sec:dataset_ruclevr]{ruCLEVR}\xspace}
\newcommand{\rucommonvqa}{\hyperref[sec:dataset_rucommonvqa]{ruCommonVQA}\xspace}
\newcommand{\ruenvaqa}{\hyperref[sec:dataset_ruenvaqa]{ruEnvAQA}\xspace}
\newcommand{\ruhhhimage}{\hyperref[sec:dataset_ruhhhimage]{ruHHH-Image}\xspace}
\newcommand{\ruhhhvideo}{\hyperref[sec:dataset_ruhhhvideo]{ruHHH-Video}\xspace}
\newcommand{\rumathvqa}{\hyperref[sec:dataset_rumathvqa]{ruMathVQA}\xspace}
\newcommand{\runaturalsciencevqa}{\hyperref[sec:dataset_runaturalsciencevqa]{ruNaturalScienceVQA}\xspace}
\newcommand{\rutieaudio}{\hyperref[sec:dataset_rutieaudio]{ruTiE-Audio}\xspace}
\newcommand{\rutieimage}{\hyperref[sec:dataset_rutieimage]{ruTiE-Image}\xspace}
\newcommand{\ruslun}{\hyperref[sec:dataset_ruslun]{RuSLUn}\xspace}
\newcommand{\skillname}[1]{\textbf{#1}}

\newcommand{\skillPerceptionTable}{
\begin{table*}[!htbp]
\centering
\tiny
\setlength{\extrarowheight}{2pt}
\renewcommand{\arraystretch}{1.2}
\begin{tabularx}{\textwidth}{@{}l| p{1.5cm}| p{1.5cm}| p{1.7cm}| >{\raggedright\arraybackslash}X| l| l| l@{}}
\toprule
\multicolumn{5}{c|}{\textbf{Taxonomy Levels}} & \multicolumn{3}{c}{\textbf{Modalities}} \\
\cline{1-5} \cline{6-8}
\textbf{L1} & \textbf{L2} & \textbf{L3} & \textbf{L4} & \textbf{L5} & \textbf{Image} & \textbf{Audio} & \textbf{Video} \\
\cline{1-8}
\multirow[c]{61}{*}{\rotatebox[origin=c]{90}{Perception}} & \multirow[c]{22}{*}{\parbox[c]{1cm}{Fine-grained cross-instance perception}} & \multicolumn{2}{>{\raggedright\arraybackslash}p{3.4cm}|}{\multirow[c]{3}{*}{Overlapping object differentiation}} & \skillname{Overlapping image differentiation} & {} & {X} & {} \\
\cline{5-8}
 &  & \multicolumn{2}{p{3.4cm}|}{} & \multirow[c]{2}{*}{\skillname{Speaker diarization}} & {X} & ruTiE-Audio & {} \\
 &  & \multicolumn{2}{p{3.4cm}|}{} &  &  & AQUARIA &  \\
\cline{3-8}
 &  & \multicolumn{2}{>{\raggedright\arraybackslash}p{3.4cm}|}{\multirow[c]{7}{*}{Mutual object localization}} & \multirow[c]{5}{*}{\skillname{Spatial object relationship}} & LabTabVQA & X & CommonVideoQA \\
 &  & \multicolumn{2}{p{3.4cm}|}{} &  & ruCommonVQA &  & RealVideoQA \\
 &  & \multicolumn{2}{p{3.4cm}|}{} &  & ruCLEVR &  &  \\
 &  & \multicolumn{2}{p{3.4cm}|}{} &  & ruCLEVR &  &  \\
 &  & \multicolumn{2}{p{3.4cm}|}{} &  & RealVQA &  &  \\
\cline{5-8}
 &  & \multicolumn{2}{p{3.4cm}|}{} & \multirow[c]{2}{*}{\skillname{Temporal object relationship}} &  & ruTiE-Audio & CommonVideoQA \\
 &  & \multicolumn{2}{p{3.4cm}|}{} &  &  & AQUARIA & RealVideoQA \\
\cline{3-8}
 &  & \multicolumn{2}{>{\raggedright\arraybackslash}p{3.4cm}|}{\multirow[c]{2}{*}{Repeating pattern recognition}} & \skillname{Visual pattern recognition} &  & X &  \\
\cline{5-8}
 &  & \multicolumn{2}{p{3.4cm}|}{} & \skillname{Temporal pattern recognition} & X &  &  \\
\cline{3-8}
 &  & \multicolumn{2}{>{\raggedright\arraybackslash}p{3.4cm}|}{\multirow[c]{10}{*}{Event recognition}} & \multirow[c]{3}{*}{\skillname{Object-object interaction}} & ruCommonVQA & ruEnvAQA & CommonVideoQA \\
 &  & \multicolumn{2}{p{3.4cm}|}{} &  & RealVQA & AQUARIA & RealVideoQA \\
 &  & \multicolumn{2}{p{3.4cm}|}{} &  & ruHHH-Image &  & ruHHH-Video \\
\cline{5-8}
 &  & \multicolumn{2}{p{3.4cm}|}{} & \multirow[c]{4}{*}{\skillname{Human-object interaction}} & ruCommonVQA & ruTiE-Audio & CommonVideoQA \\
 &  & \multicolumn{2}{p{3.4cm}|}{} &  & RealVQA & AQUARIA & RealVideoQA \\
 &  & \multicolumn{2}{p{3.4cm}|}{} &  & ruHHH-Image &  & ruHHH-Video \\
 &  & \multicolumn{2}{p{3.4cm}|}{} &  & ruTiE-Image &  &  \\
\cline{5-8}
 &  & \multicolumn{2}{p{3.4cm}|}{} & \multirow[c]{3}{*}{\skillname{Human-human interaction}} & ruCommonVQA & ruTiE-Audio & ruHHH-Video \\
 &  & \multicolumn{2}{p{3.4cm}|}{} &  & ruHHH-Image & AQUARIA &  \\
 &  & \multicolumn{2}{p{3.4cm}|}{} &  & ruTiE-Image &  &  \\
\cline{2-8}
 & \multirow[c]{20}{*}{\parbox[c]{1cm}{Fine-grained single-instance perception}} & \multicolumn{2}{>{\raggedright\arraybackslash}p{3.4cm}|}{\multirow[c]{8}{*}{Object recognition}} & \multirow[c]{3}{*}{\skillname{Object localization}} & ruCommonVQA & X & CommonVideoQA \\
 &  & \multicolumn{2}{p{3.4cm}|}{} &  & ruCLEVR &  & RealVideoQA \\
 &  & \multicolumn{2}{p{3.4cm}|}{} &  & RealVQA &  &  \\
\cline{5-8}
 &  & \multicolumn{2}{p{3.4cm}|}{} & \multirow[c]{5}{*}{\skillname{Object recognition}} & ruCommonVQA & ruTiE-Audio & CommonVideoQA \\
 &  & \multicolumn{2}{p{3.4cm}|}{} &  & ruCLEVR & ruEnvAQA & RealVideoQA \\
 &  & \multicolumn{2}{p{3.4cm}|}{} &  & RealVQA & AQUARIA & ruHHH-Video \\
 &  & \multicolumn{2}{p{3.4cm}|}{} &  & ruHHH-Image &  &  \\
 &  & \multicolumn{2}{p{3.4cm}|}{} &  & ruTiE-Image &  &  \\
\cline{3-8}
 &  & \multicolumn{2}{>{\raggedright\arraybackslash}p{3.4cm}|}{\multirow[c]{6}{*}{Event recognition}} & \multirow[c]{3}{*}{\skillname{Object motion recognition}} & ruCommonVQA & ruEnvAQA & CommonVideoQA \\
 &  & \multicolumn{2}{p{3.4cm}|}{} &  & RealVQA & AQUARIA & RealVideoQA \\
 &  & \multicolumn{2}{p{3.4cm}|}{} &  & ruHHH-Image &  & ruHHH-Video \\
\cline{5-8}
 &  & \multicolumn{2}{p{3.4cm}|}{} & \multirow[c]{3}{*}{\skillname{Living things motion recognition}} & ruCommonVQA & AQUARIA &  \\
 &  & \multicolumn{2}{p{3.4cm}|}{} &  & RealVQA &  & ruHHH-Video \\
 &  & \multicolumn{2}{p{3.4cm}|}{} &  & ruHHH-Image &  &  \\
\cline{3-8}
 &  & \multirow[c]{4}{*}{\parbox[c]{1.7cm}{Pose recognition}} & \parbox[c]{1.7cm}{Non-human pose recognition} & \skillname{Animal body pose recognition} &  & X &  \\
\cline{4-8}
 &  &  & \multirow[c]{3}{*}{\parbox[c]{1.7cm}{Human pose recognition}} & \skillname{Human body pose recognition} &  & X &  \\
\cline{5-8}
 &  &  &  & \skillname{Facial expression recognition} &  & X &  \\
\cline{5-8}
 &  &  &  & \skillname{Hand gesture recognition} &  & X &  \\
\cline{2-8}
 & \multirow[c]{19}{*}{\parbox[c]{1cm}{Textual grounding}} & \multicolumn{2}{>{\raggedright\arraybackslash}p{3.4cm}|}{\multirow[c]{11}{*}{Image-to-text grounding}} & \multirow[c]{4}{*}{\skillname{Scheme recognition}} & ruMathVQA & X &  \\
 &  & \multicolumn{2}{p{3.4cm}|}{} &  & ruNaturalScienceVQA &  &  \\
 &  & \multicolumn{2}{p{3.4cm}|}{} &  & UniScienceVQA &  &  \\
 &  & \multicolumn{2}{p{3.4cm}|}{} &  & SchoolScienceVQA &  &  \\
\cline{5-8}
 &  & \multicolumn{2}{p{3.4cm}|}{} & \skillname{Plot recognition} &  & X &  \\
\cline{5-8}
 &  & \multicolumn{2}{p{3.4cm}|}{} & \skillname{Table recognition} & LabTabVQA & X &  \\
\cline{5-8}
 &  & \multicolumn{2}{p{3.4cm}|}{} & \multirow[c]{5}{*}{\skillname{Text recognition (OCR)}} & LabTabVQA & X &  \\
 &  & \multicolumn{2}{p{3.4cm}|}{} &  & ruMathVQA &  &  \\
 &  & \multicolumn{2}{p{3.4cm}|}{} &  & ruNaturalScienceVQA &  &  \\
 &  & \multicolumn{2}{p{3.4cm}|}{} &  & UniScienceVQA &  &  \\
 &  & \multicolumn{2}{p{3.4cm}|}{} &  & SchoolScienceVQA &  &  \\
\cline{3-8}
 &  & \multirow[c]{7}{*}{\parbox[c]{1.7cm}{Audio-to-text grounding}} & \parbox[c]{1.7cm}{Prosody \& stress recognition} & \skillname{Prosody \& stress recognition} & X &  &  \\
\cline{4-8}
 &  &  & \multirow[c]{4}{*}{\parbox[l]{1.7cm}{Speech recognition}} & \skillname{Onomatopoeia} & X &  &  \\
\cline{5-8}
 &  &  &  & \multirow[c]{2}{*}{\skillname{Speech recognition}} & X & ruTiE-Audio &  \\
 &  &  &  &  &  & AQUARIA &  \\
 &  &  &  &  &  & RuSLUn &  \\
\cline{5-8}
 &  &  &  & \skillname{Song lyrics recognition} & X &  &  \\
\cline{3-8}
 &  & \multicolumn{2}{>{\raggedright\arraybackslash}p{3.4cm}|}{\multirow[c]{3}{*}{Media grounding}} & \skillname{Visual media grounding} &  & X &  \\
\cline{5-8}
 &  & \multicolumn{2}{p{3.4cm}|}{} & \multirow[c]{2}{*}{\skillname{Temporal media grounding}} & X &  & CommonVideoQA \\
 &  & \multicolumn{2}{p{3.4cm}|}{} &  &  &  & RealVideoQA \\
\bottomrule
\end{tabularx}
\caption{Skill taxonomy coverage of \meramulti tasks (perception part). Columns \textbf{L1-L5} show the skill hierarchy levels, while \textbf{Image}, \textbf{Audio} and \textbf{Video} columns indicate which tasks cover each perception type across different modalities.}\label{tab:skill_perception}
\end{table*}
}

\newcommand{\skillKnowledgeTable}{
\begin{table*}[t]
\tiny
\setlength{\extrarowheight}{2pt}
\renewcommand{\arraystretch}{1.2}
\begin{tabularx}{\textwidth}{@{}l| p{1.5cm}| p{1.5cm}| p{1.7cm}| >{\raggedright\arraybackslash}X| l| l| l@{}}

\toprule
\multicolumn{5}{c|}{\textbf{Taxonomy Level}} & \multicolumn{3}{c}{\textbf{Modality}} \\
\cline{1-5} \cline{6-8}
\textbf{L1} & \textbf{L2} & \textbf{L3} & \textbf{L4} & \textbf{L5} & \textbf{Image} & \textbf{Audio} & \textbf{Video}\\
\cline{1-8}
\multirow[c]{13}{*}{\rotatebox[origin=c]{90}{Knowledge}} & \multicolumn{2}{>{\raggedright\arraybackslash}p{2cm}|}{\multirow[c]{13}{*}{Knowledge}} & \multirow[c]{6}{*}{\parbox[c]{1.7cm}{Common everyday knowledge}} & \multirow[c]{5}{*}{\skillname{Common everyday knowledge}} & ruCommonVQA &  &  \\
 & \multicolumn{2}{p{2cm}|}{} &  &  & RealVQA & ruTiE-Audio & CommonVideoQA \\
 & \multicolumn{2}{p{2cm}|}{} &  &  & ruHHH-Image & ruEnvAQA & RealVideoQA \\
 & \multicolumn{2}{p{2cm}|}{} &  &  & WEIRD & AQUARIA & ruHHH-Video \\
 & \multicolumn{2}{p{2cm}|}{} &  &  & ruTiE-Image &  &  \\
\cline{5-8}
 & \multicolumn{2}{p{2cm}|}{} &  & \skillname{Ethics} & ruHHH-Image &  & ruHHH-Video \\
\cline{4-8}
 & \multicolumn{2}{p{2cm}|}{} & \multirow[c]{7}{*}{\parbox[c]{1.7cm}{Domain knowledge}} & \multirow[c]{3}{*}{\skillname{Common domain knowledge}} & ruMathVQA & ruTiE-Audio & CommonVideoQA \\
 & \multicolumn{2}{p{2cm}|}{} &  &  & RealVQA & ruEnvAQA & RealVideoQA \\
 & \multicolumn{2}{p{2cm}|}{} &  &  & ruTiE-Image & AQUARIA &  \\
\cline{5-8}
 & \multicolumn{2}{p{2cm}|}{} &  & \multirow[c]{4}{*}{\skillname{Expert domain knowledge}} & ruMathVQA &  &  \\
 & \multicolumn{2}{p{2cm}|}{} &  &  & ruNaturalScienceVQA &  &  \\
 & \multicolumn{2}{p{2cm}|}{} &  &  & UniScienceVQA &  &  \\
 & \multicolumn{2}{p{2cm}|}{} &  &  & SchoolScienceVQA &  &  \\
\bottomrule
\end{tabularx}\caption{Knowledge taxonomy structure and multimodal task distribution in \meramulti. Columns \textbf{L1-L5} show the skill hierarchy levels, while \textbf{Image}, \textbf{Audio} and \textbf{Video} columns indicate which tasks cover each knowledge type across different modalities.}\label{tab:skill_knowledge}
\end{table*}
}

\newcommand{\skillReasoningTable}{
\begin{table*}[!htbp]
\setlength{\tabcolsep}{6pt}
\setlength{\extrarowheight}{1pt}
\tiny
\renewcommand{\arraystretch}{1.2}
\begin{tabularx}{\textwidth}{@{}l| p{1.5cm}| p{1.5cm}| p{1.7cm}| >{\raggedright\arraybackslash}X| l| l| l@{}}
\toprule
\multicolumn{5}{c|}{\textbf{Taxonomy Level}} & \multicolumn{3}{c}{\textbf{Modality}} \\
\cline{1-5} \cline{6-8}
\textbf{L1} & \textbf{L2} & \textbf{L3} & \textbf{L4} & \textbf{L5} & \textbf{Image} & \textbf{Audio} & \textbf{Video}\\
\cline{1-8}
\multirow[c]{58}{*}{\rotatebox[origin=c]{90}{Reasoning}} & \multirow[c]{25}{*}{\parbox[c]{1cm}{Inductive reasoning}} & \multirow[c]{24}{*}{\parbox[c]{1cm}{Attribute recognition}} & \multirow[c]{12}{*}{\parbox[c]{1cm}{Coarse attribute recognition}} & \skillname{Generated content detection} &  &  &  \\
\cline{5-8}
 &  &  &  & \skillname{Source characterization} &  &  &  \\
\cline{5-8}
 &  &  &  & \skillname{Media characteristic understanding} &  &  &  \\
\cline{5-8}
 &  &  &  & \skillname{Speech emotion recognition} &  & AQUARIA &  \\
\cline{5-8}
 &  &  &  & \skillname{Music emotion recognition} &  & AQUARIA &  \\
\cline{5-8}
 &  &  &  & \skillname{Melodic structure interpretation} &  &  &  \\
\cline{5-8}
 &  &  &  & \skillname{Topic understanding} & ruTiE-Image & ruTiE-Audio &  \\
\cline{5-8}
 &  &  &  & \skillname{Style \& genre understanding} & RealVQA & AQUARIA &  \\
\cline{5-8}
 &  &  &  & \multirow[c]{4}{*}{\skillname{Scene understanding}} & ruCommonVQA & ruTiE-Audio & CommonVideoQA \\
 &  &  &  &  & RealVQA & ruEnvAQA & RealVideoQA \\
 &  &  &  &  & ruHHH-Image & AQUARIA & ruHHH-Video \\
 &  &  &  &  & ruTiE-Image &  &  \\
\cline{4-8}
 &  &  & \multirow[c]{12}{*}{\parbox[c]{1cm}{Object attribute recognition}} & \multirow[c]{6}{*}{\skillname{Physical property understanding}} & ruCommonVQA & ruEnvAQA & CommonVideoQA \\
 &  &  &  &  & ruCLEVR & AQUARIA & RealVideoQA \\
 &  &  &  &  & RealVQA &  &  \\
 &  &  &  &  & UniScienceVQA &  &  \\
 &  &  &  &  & SchoolScienceVQA &  &  \\
 &  &  &  &  & WEIRD &  &  \\
\cline{5-8}
 &  &  &  & \multirow[c]{4}{*}{\skillname{Object function understanding}} & ruCommonVQA & ruEnvAQA & CommonVideoQA \\
 &  &  &  &  & RealVQA &  & RealVideoQA \\
 &  &  &  &  & ruHHH-Image &  & ruHHH-Video \\
 &  &  &  &  & WEIRD &  &  \\
\cline{5-8}
 &  &  &  & \multirow[c]{2}{*}{\skillname{Identity \& emotion understanding}} & ruCommonVQA & AQUARIA &  \\
 &  &  &  &  & WEIRD &  &  \\
\cline{3-8}
 &  & \multicolumn{2}{>{\raggedright\arraybackslash}p{2.7cm}|}{Other inductive reasoning} & \skillname{Other inductive reasoning} &  &  &  \\
\cline{2-8}
 & \multicolumn{3}{>{\raggedright\arraybackslash}p{3.7cm}|}{\multirow[c]{3}{*}{Deductive reasoning}} & \skillname{Weirdness understanding} & WEIRD &  &  \\
 \cline{5-8}
 & \multicolumn{3}{p{3.7cm}|}{} & \skillname{Analogical reasoning} & ruTiE-Image & ruTiE-Audio &  \\
 \cline{5-8}
 & \multicolumn{3}{p{3.7cm}|}{} & \skillname{Other deductive reasoning} &  &  &  \\
\cline{2-8}
 & \multicolumn{3}{>{\raggedright\arraybackslash}p{3.7cm}|}{\multirow[c]{2}{*}{\raggedright Abductive reasoning}} & {\skillname{Hypothetical reasoning}} & RealVQA & AQUARIA & CommonVideoQA \\

\cline{5-8}
 & \multicolumn{3}{p{3.7cm}|}{} & \multirow[c]{2}{*}{\skillname{Cause \& effect understanding}} & RealVQA & AQUARIA & CommonVideoQA \\
 & \multicolumn{3}{p{3.7cm}|}{} &  &  &  & RealVideoQA \\
\cline{2-8}
 & \multicolumn{2}{>{\raggedright\arraybackslash}p{2cm}|}{\multirow[c]{17}{*}{Quantitative reasoning}} & \multirow[c]{11}{*}{\parbox[c]{1.7cm}{Counting}} & \multirow[c]{8}{*}{\skillname{Static counting}} & LabTabVQA & X & CommonVideoQA \\
 & \multicolumn{2}{p{2cm}|}{} &  &  & ruCommonVQA &  & RealVideoQA \\
 & \multicolumn{2}{p{2cm}|}{} &  &  & ruCLEVR &  &  \\
 & \multicolumn{2}{p{2cm}|}{} &  &  & ruNaturalScienceVQA &  &  \\
 & \multicolumn{2}{p{2cm}|}{} &  &  & RealVQA &  &  \\
 & \multicolumn{2}{p{2cm}|}{} &  &  & UniScienceVQA &  &  \\
 & \multicolumn{2}{p{2cm}|}{} &  &  & SchoolScienceVQA &  &  \\
 & \multicolumn{2}{p{2cm}|}{} &  &  & ruTiE-Image &  &  \\
\cline{5-8}
 & \multicolumn{2}{p{2cm}|}{} &  & \multirow[c]{3}{*}{\skillname{Temporal counting}} & X & ruTiE-Audio & CommonVideoQA \\
 & \multicolumn{2}{p{2cm}|}{} &  &  &  & ruEnvAQA & RealVideoQA \\
 & \multicolumn{2}{p{2cm}|}{} &  &  &  & AQUARIA &  \\
\cline{4-8}
 & \multicolumn{2}{p{2cm}|}{} & \multirow[c]{6}{*}{\parbox[c]{1cm}{Mathematical reasoning}} & \multirow[c]{6}{*}{\skillname{Mathematical reasoning}} & ruMathVQA & ruTiE-Audio & CommonVideoQA \\
 & \multicolumn{2}{p{2cm}|}{} &  &  & ruNaturalScienceVQA &  & RealVideoQA \\
 & \multicolumn{2}{p{2cm}|}{} &  &  & RealVQA &  &  \\
 & \multicolumn{2}{p{2cm}|}{} &  &  & UniScienceVQA &  &  \\
 & \multicolumn{2}{p{2cm}|}{} &  &  & SchoolScienceVQA &  &  \\
 & \multicolumn{2}{p{2cm}|}{} &  &  & ruTiE-Image &  &  \\
\cline{2-8}
 & \multicolumn{3}{>{\raggedright\arraybackslash}p{3.7cm}|}{\multirow[c]{10}{*}{Other reasoning}} & \skillname{Critical thinking} &  &  &  \\
\cline{5-8}
 & \multicolumn{3}{p{3.7cm}|}{} & \skillname{Counterfactual robustness} & RealVQA &  &  \\
\cline{5-8}
 & \multicolumn{3}{p{3.7cm}|}{} & \multirow[c]{5}{*}{\skillname{Problem decomposition}} & ruMathVQA &  &  \\
 & \multicolumn{3}{p{3.7cm}|}{} &  & ruNaturalScienceVQA &  &  \\
 & \multicolumn{3}{p{3.7cm}|}{} &  & RealVQA &  &  \\
 & \multicolumn{3}{p{3.7cm}|}{} &  & UniScienceVQA &  &  \\
 & \multicolumn{3}{p{3.7cm}|}{} &  & SchoolScienceVQA &  &  \\
\cline{5-8}
 & \multicolumn{3}{p{3.7cm}|}{} & \multirow[c]{3}{*}{\skillname{Comparative reasoning}} & RealVQA & ruEnvAQA &  \\
 & \multicolumn{3}{p{3.7cm}|}{} &  & UniScienceVQA & AQUARIA &  \\
 & \multicolumn{3}{p{3.7cm}|}{} &  & SchoolScienceVQA &  &  \\
\bottomrule
\end{tabularx}
\caption{Skill taxonomy coverage of \meramulti tasks (reasoning part). Columns \textbf{L1-L5} show the skill hierarchy levels, while \textbf{Image}, \textbf{Audio} and \textbf{Video} columns indicate which tasks cover each reasoning type across different modalities.}\label{tab:skill_reasoning}
\end{table*}
}

\title{Multimodal Evaluation of Russian-language Architectures}

\author{
 \textbf{Artem Chervyakov\textsuperscript{*}\textsuperscript{1}},
 \textbf{Ulyana Isaeva\textsuperscript{*}\textsuperscript{1}},
 \textbf{Anton Emelyanov\textsuperscript{1}},
 \textbf{Artem Safin\textsuperscript{1}},
 \textbf{Maria Tikhonova\textsuperscript{1,4}},
\\
 \textbf{Alexander Kharitonov\textsuperscript{1}},
 \textbf{Yulia Lyakh\textsuperscript{1}},
 \textbf{Petr Surovtsev\textsuperscript{1}},
 \textbf{Denis Shevelev\textsuperscript{1}},
 \textbf{Vildan Saburov\textsuperscript{1,8}},
\\
 \textbf{Vasily Konovalov\textsuperscript{3,5,6}},
 \textbf{Elisei Rykov\textsuperscript{5,7}},
 \textbf{Ivan Sviridov\textsuperscript{2}},
 \textbf{Amina Miftakhova\textsuperscript{2}},
 \textbf{Ilseyar Alimova\textsuperscript{5}},
 \\
 \textbf{Alexander Panchenko\textsuperscript{5,3}},
 \textbf{Alexander Kapitanov\textsuperscript{1}},
 \textbf{Alena Fenogenova\textsuperscript{*}\textsuperscript{1}}
\\
\textsuperscript{1}SberAI,
\textsuperscript{2}Sber AI Lab,
\textsuperscript{3}AIRI,
\textsuperscript{4}HSE University,
\textsuperscript{5}Skoltech,
\textsuperscript{6}MIRAI,\\
\textsuperscript{7}T-Tech,
\textsuperscript{8}Moscow Center for Advanced Studies
\\
 \small{
   \textbf{Correspondence:} \href{mailto:mera@a-ai.ru}{mera@a-ai.ru}
 }
}

\begin{document}
\maketitle
\begin{abstract}

\let\thefootnote\relax\footnotetext{* Core contributors}

Multimodal large language models (MLLMs) are currently at the center of research attention, showing rapid progress in scale and capabilities, yet their intelligence, limitations, and risks remain insufficiently understood. To address these issues, particularly in the context of the Russian language, where no multimodal benchmarks currently exist, we introduce \textbf{\meramulti}, an open multimodal evaluation framework for Russian-spoken architectures. 
The benchmark is instruction-based and encompasses default text, image, audio, and video modalities, comprising 18 newly constructed evaluation tasks for both general-purpose models and modality-specific architectures (image-to-text, video-to-text, and audio-to-text). Our contributions include: (i) a universal taxonomy of multimodal abilities; (ii) 18 datasets created entirely from scratch with attention to Russian cultural and linguistic specificity, unified prompts, and metrics; (iii) baseline results for both closed-source and open-source models; (iv) a methodology for preventing benchmark leakage, including watermarking for private sets. While our current focus is on Russian, the proposed benchmark provides a replicable methodology for constructing multimodal benchmarks in typologically diverse languages, particularly within the Slavic language family.
\end{abstract}


\begin{figure*}
    \centering
    \includegraphics[width=\linewidth]{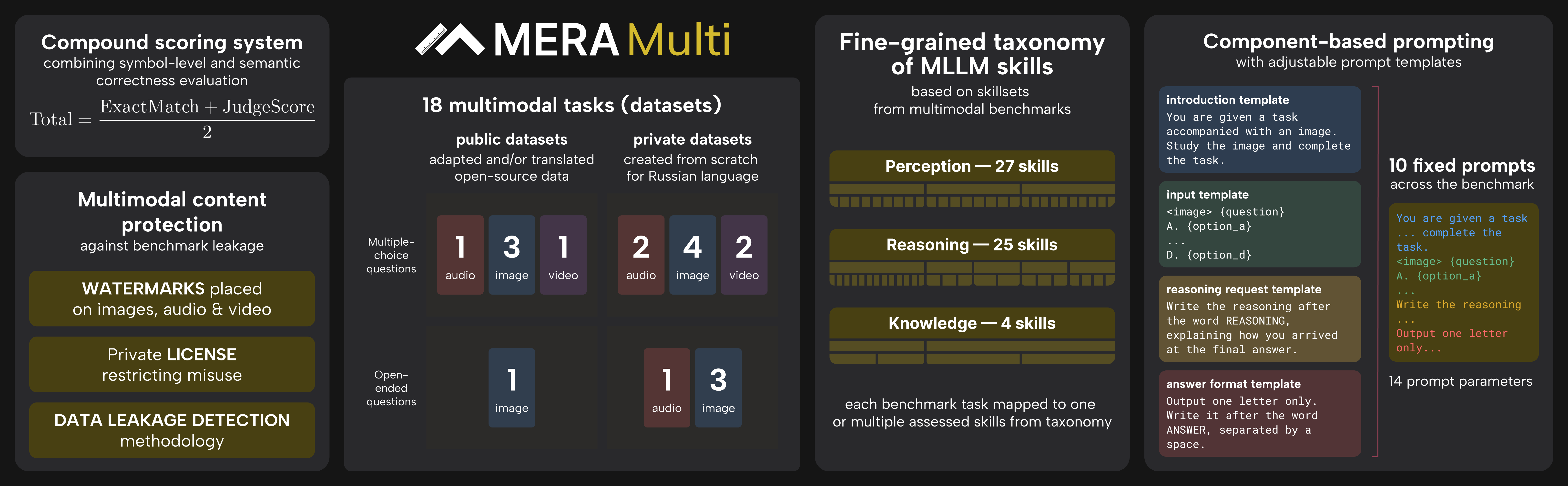}
    \caption{Overview of \meramulti.
The benchmark unites multimodal evaluation, taxonomy-based skill assessment, and data leakage protection across 18 tasks covering (default) text, image, audio, and video modalities. It employs standardized block-prompting, compound scoring, and integrates methods for multimodal content protection, forming a transparent and robust methodology for culturally grounded multimodal evaluation in Russian.}
    \label{fig:main}
\end{figure*}

\vspace{-5px}
\section{Introduction}
\label{sec:introduction}
\vspace{-5px}

Recent breakthroughs in generative AI, including models like GPT-5\footnote{\url{https://openai.com/index/introducing-gpt-5}}
, ImageBind~\cite{girdhar2023imagebind}, and LLaVa~\cite{DBLP:conf/cvpr/LiuLLL24}, have significantly advanced the state of the art across multiple modalities. This accelerated progress created a growing need for comprehensive multimodal benchmarks capable of rigorous evaluation of the full spectrum of versatile capabilities of such models.
Although several benchmarks have been proposed for English and general-domain evaluation, such as MultiBench~\cite{liang2021multibench}, MMBench~\cite{liu2024mmbench}, and General-Bench~\cite{fei2025path}, they predominantly neglect the linguistic and cultural nuances of Slavic languages, particularly Russian. Beyond its Cyrillic script, Russian possesses a rich cultural context where concepts familiar to native speakers (e.g., folklore, Soviet media) are foreign to others, creating a challenge for automatic understanding.

Existing Russian-specific benchmarks, including  TAPE~\cite{taktasheva2022tape}, Russian SuperGLUE~\cite{fenogenova2022russian}, and MERA~\cite{fenogenova2024mera}, focus exclusively on text-based tasks, leaving a critical gap in multimodal evaluation. To address this, we introduce \textbf{\meramulti}\footnote{\url{https://mera.a-ai.ru/en/multi}}, the first multimodal benchmark for MLLM evaluation in Russian. It comprises 18 tasks spanning (default) text\footnote{Since textual input is an inherent component of all tasks in our benchmark, we treat it as the default modality. Consequently, we do not explicitly list it when describing or tabulating tasks, unless required for clarity.}, image, audio, and video modalities, built upon a unified taxonomy of multimodal abilities. Beyond Russian \meramulti offers a blueprint for developing multimodal benchmarks across other Slavic and morphologically rich languages. These languages share structural complexity, typological proximity, and cultural specificity that make direct translation or adaptation of English-centric benchmarks inadequate. Thus, it serves not only as the first multimodal benchmark for Russian, but also as a scalable methodology for culturally and linguistically aware evaluation in underrepresented settings.

More specifically, our contributions are fourfold:
\begin{itemize}[nosep]
    \item We propose a unified taxonomy and evaluation methodology designed for MLLMs assessment;
    \item We create 18 novel datasets \footnote{\url{https://hf.co/collections/MERA-evaluation/mera-multimodality}} incorporating Russian cultural and linguistic specificities, unified prompts, and metrics. 
    \item We provide baseline performance results for both open- and closed-source models;
    \item We establish a data leakage analysis and watermarking strategy to protect private evaluation datasets.
\end{itemize}

Additionally, we provide a standardized codebase for full reproducibility and a submission platform with automated scoring and a public leaderboard\footnote{\url{https://mera.a-ai.ru/en/multi/leaderboard}}. All datasets and code are made available under the \meramulti license, which allows the use of the benchmark sets for non-commercial purposes, provided that the data is not used for model training of any kind. These elements establish a foundation for transparent and culturally aware multimodal evaluation in Russian, while positioning \meramulti as a reference point for developing similar frameworks across Slavic and other non-English languages, thereby promoting broader community development and cross-linguistic benchmarking.

\begin{table*}[t]
\centering
\tiny
\begin{tabular}{lcccccc}
\toprule
\textbf{Benchmark} & \textbf{Modalities} & \textbf{\# Tasks/Skills} & \textbf{Primary Focus} & \textbf{Language} &\textbf{Cultural Focus} \\
\midrule
MultiBench~\cite{liang2021multibench} & 10 (Text, Image, Audio, Video, Table, Set, ...) & 20 & General-purpose representation & EN  & General \\
MMBench~\cite{liu2024mmbench} & Image, Text & 20 & Fine-grained visual reasoning & EN, ZH  & General \\
SUPERB~\cite{yang2021superb} & Speech, Audio & 10  & Speech processing & EN  & General \\
STAR~\cite{wu2024star} & Video, Text & 4  & VideoQA, temporal reasoning & EN  & General \\
InfiniBench~\cite{ataallah2024infinibench} & Video, Text & 8  & Long-context video understanding & EN & General \\
Video-MME~\cite{fu2025video} & Video, Text & 12  & Fine-grained video analysis & EN  & General \\
General-Bench~\cite{fei2025path} & Text, Image, Video, Audio, Tabule & 700+ & Large-scale ability coverage & EN  & General \\
OmniDialog~\cite{razzhigaev2024omnidialog} & Text, Image, Audio & 8  & Multimodal dialogue & EN  & General \\
MMMU~\cite{yue2023mmmu} & Image, Text & 30 & Expert-level reasoning & EN  & General \\
SEED-Bench~\cite{li2023seed} & Image, Video, Text & 12 & Generative comprehension & EN  & General \\
\midrule
{\meramulti} (ours) & Text, Image, Audio, Video & 18 & Comprehensive understanding & RU & General + Russian \\
\bottomrule
\end{tabular}
\caption{Comparison of major multimodal evaluation benchmarks and \meramulti.}
\label{tab:benchmark_comparison}
\end{table*}

\section{Related Work}
\label{sec:rel_work}
\paragraph{Text-Based Benchmarks}
Evaluation of language models has historically relied on Natural Language Understanding benchmarks such as GLUE~\cite{wang2018glue} and SuperGLUE~\cite{wang2019superglue}, which set initial standards for English. As these  benchmarks became saturated, newer instruction-oriented and reasoning-focused benchmarks emerged such as BIG-bench~\cite{DBLP:conf/acl/SuzgunSSGTCCLCZ23} and HELM~\cite{perlitz-etal-2024-efficient}.
Further efforts such as MMLU~\cite{hendrycks2020measuring}, AGIEval~\cite{zhong2023agieval}, and C-Eval~\cite{huang2023c} extended evaluation to academic and professional domains, though primarily in English and Chinese.
For Russian, several text-based benchmarks were introduced. Among them are Russian SuperGLUE~\cite{shavrina2020russiansuperglue,fenogenova2022russian}, TAPE~\cite{taktasheva2022tape}, and RuCoLA~\cite{mikhailov2022rucola}.
The instruction-based benchmark MERA~\cite{fenogenova2024mera} advanced this line of work but remained limited to text-only evaluation. Our work extends this direction toward multimodal, instruction-following assessment of MLLMs in Russian.

\paragraph{Multimodal Benchmarks}
The rapid progress of multimodal models has led to numerous benchmarks extending evaluation beyond text. Early large-scale efforts as MultiBench~\cite{liang2021multibench} covered 10 modalities and emphasized general-purpose representation learning. MMBench~\cite{liu2024mmbench} focused on fine-grained visual reasoning and bilingual (English–Chinese) evaluation, while SUPERB~\cite{yang2021superb} unified diverse audio tasks under a single framework. For video understanding, STAR~\cite{wu2024star}, InfiniBench~\cite{ataallah2024infinibench}, and Video-MME~\cite{fu2025video} advanced evaluation toward temporal and long-context reasoning.
More comprehensive setups such as General-Bench~\cite{fei2025path}, OmniDialog~\cite{razzhigaev2024omnidialog}, MMMU~\cite{yue2023mmmu}, and SEED-Bench~\cite{li2023seed} assess multimodal and reasoning abilities at scale. 
All in all, despite these advances, most existing multimodal benchmarks are English-centric. To our knowledge, there is currently no existing multimodal benchmark for the Russian language. \meramulti addresses the gap by providing a unified and culturally adapted evaluation of understanding across several modalities (see~\autoref{tab:benchmark_comparison} for the detailed comparison between \meramulti and other benchmarks).

\section{Overview of \meramulti Benchmark}
\label{sec:mera}
\begin{table}[t]
\centering
\scriptsize
\begin{tabular}{@{}p{0.2cm}|p{2cm}llp{0.52cm}p{1.6cm}}
\toprule
\textbf{} & \textbf{Dataset / task} & \textbf{Size} & \textbf{HB} & \textbf{Answer} & \textbf{License}\\
\midrule
\multirow{4}{*}{\rotatebox[origin=c]{90}{audio}} &  \ruenvaqa & 596 & 0.95 & MC & CC BY-NC 4.0 \\
&  \ruslun & 741 & 0.90 & OE & CC-BY-4.0 \\
& *\aquaria & 738 & 0.98 & MC & \meramulti \\
& *\rutieaudio & 1500 & 0.75 & MC & \meramulti \\
\midrule
\multirow{11}{*}{\rotatebox[origin=c]{90}{image}} &  \ruclevr & 1148 & 0.96 & OE & CC-BY-4.0 \\
&  \rucommonvqa & 3015 & 0.84 & OE & CC-BY-4.0 \\
&  \runaturalsciencevqa & 363 & 0.99 & MC & CC-BY-SA-4.0 \\
&  \weird & 814 & 0.85 & MC & CC-BY-4.0 \\
& *\labtabvqa & 339 & 0.91 & MC & \meramulti \\
& *\realvqa & 773 & 0.63 & OE &  \meramulti \\
& *\ruhhhimage & 595 & 0.89 & MC & \meramulti \\
& *\rumathvqa & 502 & 0.95 & OE & \meramulti \\
& *\rutieimage & 1500 & 0.77 & MC & \meramulti \\
& *\schoolsciencevqa & 4227 & 0.82 & MC & \meramulti \\
& *\unisciencevqa & 7432 & 0.13 & OE & \meramulti \\
\midrule
\multirow{3}{*}{\rotatebox[origin=c]{90}{video}} &  \commonvideoqa & 1200 & 0.96 & MC & CC-BY-4.0 \\
& *\realvideoqa & 671 & 0.96 & MC & \meramulti \\
& *\ruhhhvideo & 911 & 0.84 & MC & \meramulti \\

\bottomrule
\end{tabular}
\caption{Overview of datasets in \meramulti. Those marked with an asterisk are \textit{private datasets} collected from scratch, while the others are \textit{public datasets} compiled from open-source datasets. \textbf{Size} column shows the number of samples in the dataset. \textbf{HB} is the human baseline value (basic / expert or basic only, see \ref{sec:appendix_hb} for details). \textbf{Answer} column is the task format (MC and OE stand for multiple-choice and open-ended, respectively). \textit{\meramulti} license refers to the benchmark license anonymized for the review period.}
\label{tab:datasets}
\end{table}

Figure~\ref{fig:main} presents an overview of \meramulti and its methodological parts.

\subsection{Benchmark General Structure}
\label{sec:structure}

The proposed benchmark is designed to evaluate the capabilities of Russian MLLMs. 
Besides the omnipresent textual modality, it incorporates tasks from three other modalities: image (11 datasets), audio (4 datasets), and video (3 datasets). 
The tasks are formulated in two primary formats: multiple-choice questions and open-ended questions requiring short free-form answers. 

To balance reproducibility and novelty, the benchmark integrates both publicly available datasets (7 tasks), curated from open data sources, and private datasets (11 tasks), collected from scratch specifically for this study. The latter are designed to incorporate Russian cultural nuances, target underexplored skill categories, and mitigate potential data contamination issues (see details in section~\ref{sec:dataleakeage}).
A complete list of datasets is provided in~\autoref{tab:datasets}. Task examples and dataset creation details are in \ref{sec:appendix_datasets}.

\subsection{Skill Taxonomy}
\label{sec:taxonomy}

Contemporary studies on multimodal benchmarks often rely on custom skill sets during the dataset design process \cite{liu2024mmbench,fu2024mmecomprehensiveevaluationbenchmark}.
We performed a comprehensive analysis of such systems and synthesized them into a consolidated MLLM skill taxonomy, which underpins the foundation of \meramulti.
Such a system functions as a comprehensive map for skill coverage, which, when coupled with an alignment of existing datasets to corresponding skills, can spotlight deficiencies in benchmark task diversity.
 
The taxonomy is aligned with three broad categories of human-like cognition: \textbf{knowledge}, \textbf{perception}, and \textbf{reasoning}, a division also adopted by recent multimodal benchmarks. The knowledge taxonomy is shown in~\autoref{tab:skill_knowledge}, and the perception and reasoning parts are provided in the Appendix in~\autoref{tab:skill_perception} and~\autoref{tab:skill_reasoning} respectively, due to space constraints.
To comply with the emerging MLLM capabilities, this taxonomy is designed to be extendable and this paper provides the initial version of the unified taxonomy and encourages its adoption and extension in future research.

Each \meramulti\ dataset is systematically mapped to a predefined set of multimodal skills. Rather than assigning a single skill per task, each task is designed to be multifaceted, evaluating a distinctive combination of abilities.
For example, a single task might require visual perception (to identify objects), OCR (to read text in the image), and reasoning (to answer a why-question) all together. 

Further details on skill taxonomy are in~\autoref{sec:appendix_skill_taxonomy}.

\skillKnowledgeTable

\subsection{Evaluation Methodology}
\label{sec:methodology}
\meramulti\ evaluation methodology is designed to systematically assess the multimodal reasoning, perception, and knowledge abilities of MLLMs. It measures both general-purpose and modality-specific competence across image, audio, and video inputs. To achieve this, we propose the benchmark that integrates four complementary components: (i) a block-prompting structure that ensures consistency and diversity of task formulations across modalities (\autoref{sec:prompts}); (ii) dual-level evaluation metrics combining symbolic and semantic correctness through a dedicated LLM-as-a-judge model (\autoref{sec:scoring}); (iii) an evaluation pipeline including scoring aggregation and cross-modality weighting (\autoref{sec:evaluation}); and (iv) a submission protocol enabling automated, reproducible leaderboard updates (\autoref{sec:submission}). These components provide a coherent methodology that balances rigor, interpretability, and cross-model comparability in multimodal evaluation.

\subsubsection{Evaluation Framework Implementation}
\label{sec:harness}
We build the benchmark code base on the \texttt{lm-eval}\footnote{\url{https://github.com/EleutherAI/lm-evaluation-harness}} framework~\citep{eval-harness,eval-harness-paper}, extending it to support multimodal inputs while preserving its core structure for texts. The codebase introduces comprehensive vision, audio, and video evaluation through, and chat-template formatting tailored for instruction-tuned and API-based models. Multimodal data is integrated either by passing it separately to the model's processor or by embedding it directly within a chat template alongside the text. All evaluations are strictly generative: models produce free-form answers until a stop condition is met. All outputs are used directly except \ruslun which requires minimal parsing for structured output.

To balance rigor and semantic understanding, we employ two primary metrics across most datasets. $Exact Match$ ($EM$) serves as a generative analog of accuracy, using normalized string comparison to assess both factual correctness and strict adherence to the specified output format. Complementing this, the $Judge Score$ ($JS$) measures semantic similarity via an LLM-as-a-judge (trained for this purpose) that scores an answer as 1 for substantive agreement with the reference and 0 otherwise. This dual approach provides a nuanced view: EM is sensitive to format, while the JS focuses on meaning. For the datasets where this format is unsuitable we employ task-specific metrics or report only the JS.

The $Final Score$ of the task (for Total Score calculation purposes) is defined as follows:
\begin{equation}
    FS(task) = \frac{\text{EM}(task) + \text{JS}(task)}{2};
\end{equation}

It should be noted that as long as some of our prompts suggest that the model return the answer for the question after a specific phrase (``ANSWER''), we consider $EM$ to be the maximum of two scores: (i) $EM$ of the full model answer, (ii) $EM$ of the last part of the model's answer split by the previously mentioned specific phrase. If the model breaches the instruction, we assess the full answer. Otherwise, we consider the part that is after the specific phrase to be the model's answer as required by the instruction.

\subsubsection{Prompt Structure}
\label{sec:prompts}

Following the approach of~\citet{voronov2024mind}, we avoid hard-coding task-specific prompts by employing a block-prompting scheme that constructs universal prompts aligned with each task’s structure. For every dataset, we instantiate a predefined set of 10 prompts drawn from fixed block layouts that we initially designed and realized under different surface modalities (e.g., formal request, command, technical statement). This design preserves a uniform formulation across benchmark tasks while allowing for controlled variation, such as the presence or absence of a reasoning request, to mitigate benchmark saturation.

Following the methodology of~\citet{fenogenova2024mera}, we uniformly assign prompts for each task across dataset samples to ensure that the aggregated performance metric reflects an average over prompt variants rather than the idiosyncrasies of a single prompt. A post-hoc statistical analysis (see~\autoref{sec:appendix_prompts_analysis} for the details) confirmed that our prompt set is not invariant, as some prompts produce statistically significant shifts in model scores. This finding directly motivates and validates our multi-prompt design as a necessary guard against evaluation bias. 

\subsubsection{Judge Scoring}
\label{sec:scoring}

Evaluating open-ended model generations requires moving beyond strict heuristics. We frame this as a task of assessing semantic equivalence, conditioned on the question, as devising universal regex patterns to extract answers from free-form text, particularly from multi-step reasoning chains -- proves infeasible. The practically infinite space of possible valid generations necessitates a more flexible and scalable assessment approach. We therefore introduce a learned LLM-as-judge model, framing evaluation as a question-conditioned semantic equivalence task (binary classification -- determine the correctness of a model's prediction).

This judge is trained on a diverse, human-annotated dataset of model outputs collected from various Russian-language instruction-based benchmarks. The final model, based on the RuModernBERT\footnote{\url{https://hf.co/deepvk/RuModernBERT-base}}~\cite{deepvk2025rumodernbert} encoder, achieves F1 score of 0.96 on a held-out test set and shows 99.6\% agreement with EM on identical answers, confirming its reliability. We deploy this judge as our primary metric, providing a robust, scalable score for correctness across the benchmark. Comprehensive details on data, training, and model comparisons are in the~\autoref{sec:judge_appendix}.

\subsubsection{Evaluation Pipeline}
\label{sec:evaluation}

Let \meramulti comprise $M$ modalities (e.g., image, audio, video) not counting the omnipresent text one, with modality $m$ containing $n_m$ tasks. We treat modalities as equally important and split each modality's weight evenly across its tasks. Let $k_m$ be the number of tasks a model attempted in modality $m$ with $s_{m,i} \in [0; 1]$ being the score (average of all task $i$ metrics).

\begin{itemize}[nosep]
\item The \textbf{Attempted Score}. Quality over the tasks the model attempted, with the original equal-per-modality weighting renormalized over the attempted subset:
\begin{equation}\label{eqn:precision}
    A = \frac{\sum_{m=1}^{M} \frac{1}{n_m} \sum_{i \in A_m} s_{m,i}}{\sum_{m=1}^{M} \frac{k_m}{n_m}}
\end{equation}
\item The \textbf{Coverage}. Breadth of evaluation is the fraction of the benchmark the model actually attempted: 
\begin{equation}\label{eqn:precision}
    C = \frac{1}{M} \sum_{m=1}^{M} \frac{k_m}{n_m}
\end{equation}
\item The \textbf{Total Score}. We combine quality and breadth as
\begin{equation}\label{eqn:precision}
    T = A \cdot C
\end{equation}
\end{itemize}

This approach has the following positive effects:
\begin{itemize}[nosep]
    \item \textbf{Separation of concerns}. $A$ reports how well a model performs where it runs; $C$ reports how much of \meramulti it covers.
    \item \textbf{Fair, single leaderboard}. $T$ enables joint ranking without imputing arbitrary zeros for missing modalities; specialists still excel on modality-specific boards (higher $A$, lower $C$).
    \item \textbf{Stable under growth}. Adding/removing tasks adjusts $C$ (breadth), but leaves $A$ (quality on attempted tasks) invariant; equal-per-modality weighting prevents any modality from dominating due to task count.
\end{itemize}

\subsubsection{Submission}
\label{sec:submission}
The private dataset test answers are available only for the organizers and experts supporting the benchmark. The scoring system is automatic and is available on the benchmark platform.

The process of submission is the following:
\begin{itemize}[nosep]
    \item First, users clone \meramulti benchmark repository and form submission files using shell script and the provided code. 
    \item Second, they upload the submission files via the platform interface\footnote{For this step the registration on the platform is required.} in their personal account for automatic assessment.
    \item The evaluation result is then displayed in the user’s account and kept private unless they request publication using the ``Publish'' function. 
    \item In this case, it undergoes an expert verification of its reproducibility, which includes checking log files automatically formed by the evaluation script and the provided submission information. Once approved, the model’s score is shown publicly on the leaderboard (user can choose on which leaderbord image/audio/video/multi to add the results), while its specific outputs remain private.
\end{itemize}

\subsection{Data Protection}
\label{sec:dataleakeage}

As pre-training datasets grow, benchmark leakage is becoming more common. This issue is exacerbated by opaque training processes and the undisclosed use of supervised data in modern MLLMs, which undermines the validity of benchmarks and fosters unfair comparisons, ultimately slowing progress.
To protect the multimodal data in our private benchmark suite from unauthorized use, we employ: (i) data watermarking (\autoref{sec:watermarking}), (ii) leakage detection (\autoref{sec:data_leakage}). 
We also introduce the license, which explicitly permits the use of the benchmark data for research and non-commercial testing purposes, but strictly prohibits its incorporation into any model training process. Together, these measures provide robust technical and legal safeguards.

\subsubsection{Data Watermarking}
\label{sec:watermarking}

We embed imperceptible yet identifiable watermarks into our benchmark data to trace its provenance and detect unauthorized use in training corpora. Our approach is tailored to each modality:
\begin{itemize}[nosep]
    \item \textbf{Audio}: We use AudioSeal~\cite{roman2024proactivedetectionvoicecloning_watermarking} for localized detection, which employs a neural, inaudible watermark.
    \item \textbf{Images/Video}: a simple overlay of the \meramulti\ watermark on images and every video frame (same code across frames).
\end{itemize}

\begin{table}[t]
\scriptsize
\centering
\begin{tabular}{@{}p{0.2cm}|ll@{}}
\toprule
 & \textbf{Model} & \textbf{Confidence Interval} \\
\midrule
\multirow{9}{*}{\rotatebox[origin=c]{90}{image}} & Qwen2-VL-2B-Instruct & (0, 0.352) \\
& Qwen2-VL-7B-Instruct & (0, 3.920) \\
& Qwen2.5-VL-3B-Instruct & (0.259, 2.180) \\
& Qwen2.5-VL-7B-Instruct & (0.444, 1.389) \\
& llava-1.5-7b-hf & (0, 0.937) \\
& llava-1.5-13b-hf & (0, 0.839) \\
& llama3-llava-next-8b-hf & (0, 1.024) \\
& gemma-3-12b-it & (1.183, 3.120) \\
& gemma-3-12b-it & (0, 1.756) \\
\midrule
\multirow{7}{*}{\rotatebox[origin=c]{90}{video}} & LLaVA-NeXT-Video & (0.811, 3.314) \\
& LLaVA-NeXT-Video-DPO & (1.367, 2.125) \\
& LLaVA-NeXT-Video-34B & (0, 1.373) \\
& Qwen2-VL-2B-Instruct & (0, 0.645) \\
& Qwen2.5-VL-7B-Instruct & (0, 0.781) \\
& Qwen2.5-VL-3B-Instruct & (0, 0.330) \\
& Qwen2-VL-7B-Instruct & (0, 0.105) \\
\midrule
\multirow{3}{*}{\rotatebox[origin=c]{90}{audio}} & ultravox-v0\_2 & (0, 0.940) \\
& ultravox-v0\_3-llama-3\_2-1b & (0, 1.566) \\
& ultravox-v0\_5-llama-3\_2-1b & (0, 0.195) \\
& ultravox-v0\_6-llama-3\_1-8b & (0, 0.875) \\
& ultravox-v0\_4 & (0, 0.401) \\
& ultravox-v0\_4\_1-mistral-nemo & (0, 0.079) \\
& Qwen2-Audio-7B-Instruct & (0.0, 0.0) \\
& MiniCPM-o-2\_6 & (0.0, 0.0) \\
\bottomrule
\end{tabular}
\caption{Confidence intervals of Judge Score (JS) differences between watermarked and clear data (\%).}
\label{tab:watermarking_results}
\end{table}

Table \ref{tab:watermarking_results} demonstrates the Confidential Intervals (CI) for differences of the models' metrics on data with watermarks and without them. The CI are computed for differences in per cents for convenience. The results demonstrate that with 95\% probability the difference is less than 5\% (usually less than 1\%) which leads to the conclusion: our watermarking strategy does not significantly affect the evaluation results.

\subsubsection{Data Leakage Detection}
\label{sec:data_leakage}

We detect training-set data leakage using membership inference attacks~\cite{shokri2017membership}. Our approach~\cite{emelyanov2025fimmia} extends the Semantic Membership Inference Attack (SMIA) \cite{mozaffari2024semanticmembershipinferenceattack_smia} to MLLMs, calculating the loss for a data point by considering its text in conjunction with its paired image, video, or audio data. The Multimodal SMIA (MSMIA) method identifies leakage by comparing a model's behavior on original examples versus their semantically perturbed "neighbors" (masked, removed, doubled, switched text tokens); models that have been trained on the data (models with data leak) are to exhibit systematically different loss patterns.

\begin{table}[t]
\footnotesize
\centering
\begin{tabular}{lc}
\toprule
\textbf{Modality} & \textbf{AUC-ROC} \\
\midrule
Image & 88.658 \\ 
Video & 88.388 \\ 
Audio & 81.250 \\ 
\bottomrule
\end{tabular}
\caption{Average AUC-ROC of MSMIA per modality. Averaging over the models used for training and evaluating MSMIA.}
\label{tab:msmia_results_overall}
\end{table}

Concretely, we train the MSMIA detector by comparing two versions of a model: the original and a version we fine-tune on candidate data (simulating a leak). The detector learns to distinguish between them by analyzing the differences in loss and text embeddings when processing original data points versus their perturbed neighbors. Once trained, this detector can be applied to any other model to output a probability that a specific data sample was part of that model's training set. Overall results of the MSMIA detection capabilities are presented in \autoref{tab:msmia_results_overall}. Following the original methodology, we evaluate detection performance using AUC-ROC. The table demonstrates relatively high scores, which means that the MSMIA method tends to be capable of detecting whether a model has been trained on some multimodal data samples or not, with high probability and a rather low false-positive rate. The details on the MSMIA training and the metrics are provided in~\autoref{sec:leakage_appendix}.

\section{Baselines}
\label{sec:baselines}

\subsection{Model Baselines}
We evaluate over 50 publicly available multimodal models from the most trending model families on HuggingFace, varying in size from 1B to 110B parameters. Also we evaluate proprietary GPT 4.1 (OpenAI) to make a comparison between open and closed source models.

See Appendix \ref{app:model_baselines_details} for the baseline details. We evaluate models in the same environments and scenarios by the procedure described in Section \ref{sec:evaluation} and the submission procedure described in Section \ref{sec:submission}. We also provide examples of particular submissions (the model evaluated on part of the tasks of a modality).

\subsection{Human Baselines}\label{sec:hb}
To estimate human-level performance across \meramulti tasks, we compute human baseline (HB) values based on the aggregated results of crowd annotators. For datasets requiring additional domain expertise, we also establish expert HB obtained from qualified expert annotators. Annotation quality is ensured through honeypot tasks with automated correctness verification and post-hoc filtering of low-performing annotators. All crowd annotations are collected via the ABC Elementary platform\footnote{\url{https://app.elementary.center}}, which guarantees data anonymity, fair compensation, and ethical compliance. See Appendix~\ref{sec:appendix_hb} for detailed methodology and cost breakdown .

\begin{table*}
    \centering
    \scriptsize
    \small
    \begin{tabular}{l|c|ccccc}
        \toprule
        \multirow{2}*{\textbf{Model}} & \multirow{2}*{\textbf{\begin{tabular}{@{}c@{}}\textbf{Total} \\ \textbf{Score}\end{tabular}}} & \multirow{2}*{\textbf{\begin{tabular}{@{}c@{}}\textbf{Attempted} \\ \textbf{Score}\end{tabular}}} & \multirow{2}*{\textbf{Coverage}} & \multirow{2}*{\textbf{\begin{tabular}{@{}c@{}}\textbf{Image} \\ \textbf{Total}\end{tabular}}} & \multirow{2}*{\textbf{\begin{tabular}{@{}c@{}}\textbf{Audio} \\ \textbf{Total}\end{tabular}}} & \multirow{2}*{\textbf{\begin{tabular}{@{}c@{}}\textbf{Video} \\ \textbf{Total}\end{tabular}}} \\
         & & & & & & \\
        
        \midrule
        Qwen3-Omni-30B-A3B-Inst & \textbf{0.500} & \textbf{0.563} & 0.889 & \textbf{0.554} & \textbf{0.561} & 0.410 \\
        Qwen2.5-Omni-7B & 0.317 & 0.317 & \textbf{1.000} & 0.226 & 0.474 & 0.442 \\
        Qwen2.5-VL-72B-Inst & 0.302 & 0.453 & 0.667 & 0.406 & 0.000 & \textbf{0.625} \\
        MiniCPM-o-2\_6 & 0.255 & 0.255 & \textbf{1.000} & 0.182 & 0.369 & 0.373 \\
        Qwen2-VL-72B-Inst & 0.254 & 0.381 & 0.667 & 0.333 & 0.000 & 0.557 \\
        Qwen2.5-Omni-3B & 0.251 & 0.251 & \textbf{1.000} & 0.180 & 0.380 & 0.337 \\
        Qwen2.5-VL-7B-Inst & 0.209 & 0.313 & 0.667 & 0.256 & 0.000 & 0.523 \\
        GPT 4.1 & 0.159 & 0.478 & 0.333 & 0.478 & 0.000 & 0.000 \\
        Qwen2-VL-7B-Inst & 0.145 & 0.218 & 0.667 & 0.195 & 0.000 & 0.301 \\
        Qwen2.5-VL-3B-Inst & 0.136 & 0.203 & 0.667 & 0.142 & 0.000 & 0.427 \\
        InternVL3-9B & 0.135 & 0.203 & 0.667 & 0.172 & 0.000 & 0.316 \\
        Qwen3-VL-2B-Inst & 0.125 & 0.187 & 0.667 & 0.125 & 0.000 & 0.416 \\
        ultravox-v0\_5-llama-3\_1-8b & 0.104 & 0.311 & 0.333 & 0.000 & 0.311 & 0.000 \\
        ultravox-v0\_4\_1-llama-3\_1-8b & 0.102 & 0.307 & 0.333 & 0.000 & 0.307 & 0.000 \\
        ultravox-v0\_4 & 0.101 & 0.304 & 0.333 & 0.000 & 0.304 & 0.000 \\
        ultravox-v0\_6-llama-3\_1-8b & 0.100 & 0.300 & 0.333 & 0.000 & 0.300 & 0.000 \\
        Phi-4-multimodal-instruct & 0.098 & 0.147 & 0.667 & 0.185 & 0.042 & 0.000 \\
        ultravox-v0\_4\_1-mistral-nemo & 0.088 & 0.265 & 0.333 & 0.000 & 0.265 & 0.000 \\
        audio-flamingo-3-hf & 0.086 & 0.259 & 0.333 & 0.000 & 0.259 & 0.000 \\
        llava-next-110b-hf & 0.079 & 0.236 & 0.333 & 0.236 & 0.000 & 0.000 \\
        Phi-3.5-vision-inst & 0.076 & 0.228 & 0.333 & 0.228 & 0.000 & 0.000 \\
        Qwen2-Audio-7B-Inst & 0.074 & 0.223 & 0.333 & 0.000 & 0.223 & 0.000 \\
        SmolVLM-Inst & 0.064 & 0.192 & 0.333 & 0.192 & 0.000 & 0.000 \\
        gemma-3-27b-it & 0.050 & 0.151 & 0.333 & 0.151 & 0.000 & 0.000 \\
        granite-vision-3.3-2b & 0.048 & 0.143 & 0.333 & 0.143 & 0.000 & 0.000 \\
        deepseek-vl2-small & 0.044 & 0.163 & 0.273 & 0.133 & 0.000 & 0.000 \\

    \bottomrule
    \end{tabular}
    \caption{Overall baselines information over three modalities (vision/image, audio, video). All scores are aggregated. Modality total score is the attempted score multiplied by coverage of the modality.}
    \label{tab:baselines_multi}
\end{table*}

\section{Results}
\label{sec:results}

The leaderboard is designed in such a way that the more modalities the model covers, the higher the Total Score could be. 
The top performer, Qwen3-Omni-30B-A3B-Instruct, leads with a Total Score of 0.5, driven by its high Attempted Score (0.563) and rather high Coverage (0.889), showing strong image, audio, and video capabilities. Notably, the models from Qwen families obtain larger scores for image and video modalities (first 3 places of the overall leaderboard are taken by those models). GPT 4.1 still leads in image modality while having low Coverage (0.333), which leads to a lower Total Score (0.159 compared to 0.5 of the top performer). The main trend is defined by the metrics used: broader coverage leads to higher Total Score. Thus omni-models occupy the first places. But strong uni- or bimodal capabilities may gain advantage over middle-performing models with high Coverage (e.g. Qwen2.5-VL-72B-Instruct and Qwen2-VL-72B-Instruct with 0.302 and 0.254 Total Scores respectively). 

This pattern is consistent across all modalities. In audio, the specialists from \textit{ultravox} family tend to display poorer performance compared to omni-models like Qwen2.5-Omni-7B (0.311 vs 0.474 for Audio Total Score) even though \textit{ultravox} models use other LLM's from \textit{Mistral}, \textit{Llama}, \textbf{Qwen} families as backbones. 
Considering the video modality, vision models are usually trained with video-inputs or may slice the video into frames and use ``regular'' vision encoders for them, which explains why Qwen2-VL-72B-Instruct shows the best Video Total Score (0.625) while the models that specialize specifically on video modality like those from \textbf{LLaVA-NeXT-Video} family show poorer metrics.

Consistently, \texttt{Judge Score (JS)} $>$ \texttt{EM} across models, indicating that many responses are semantically correct but violate output format; whenever \texttt{JS} $\approx$ \texttt{EM}, the model followed instructions closely. This gap justifies reporting \texttt{JS} alongside EM. When \texttt{JS} $<$ \texttt{EM}, this means that we can extract the answer from model's generation but the entire generation may be misleading (e.g. wrong rationale, reasoning conclusion mismatches the selected answer).

Tables with separate datasets metrics and analysis may be found in Appendix~\ref{app:model_baselines_details}.



\begin{tcolorbox}
\textbf{Takeaway 1}: There is still a gap between modalities. Omni-models partially bridge it. Specialist models, however, show that while image understanding is a relatively mature field, audio and video understanding are underrepresented in terms of both models and datasets, which is reflected in lower scores on benchmark tasks.
\end{tcolorbox}

\begin{tcolorbox}
\textbf{Takeaway 2}: Overall metrics are robust to missing task scores (unfinished submission) and multiple modalities. 
\end{tcolorbox}

\section*{Conclusion}
\label{sec:conclusion}

The rapid progress of generative AI has introduced new challenges for evaluating models in multimodal contexts.
We present \textbf{\meramulti}, the first comprehensive framework for transparent and culturally grounded multimodal evaluation in Russian. 
The benchmark encompasses 18 tasks across four modalities (default text, image, audio, and video), covering diverse domains and scenarios. It systematizes diverse multimodal abilities via a proposed taxonomy and evaluates them through methodologically verified prompts and metrics. 
We also provide a standardized code base\footnote{\url{https://github.com/MERA-Evaluation/MERA_MULTIMODAL}} that guarantees reproducibility and a submission platform offering automated evaluation, scoring, and open leaderboards.

In the future, we plan to expand the benchmark to encompass additional scenarios and actively encourage community contributions. We envision \meramulti as a collaborative initiative that promotes transparent evaluation practices and provides a methodological foundation for developing culturally aware multimodal benchmarks across non-English languages such as Slavic, ultimately advancing the creation of more robust and reliable multimodal models.

\section*{Limitations}

First of all, despite the fact that our benchmark covers 18 tasks spanning multiple domains, aiming to represent complementary semantic abilities of the models, this set may be underrepresenting some abilities of the model or some domains which may be crucial for certain tasks and applications. Namely, it is not impossible that a model which excels at our benchmark will perform poorly on a specialized domain or task.

Even with fixed prompts and decoding settings, \meramulti scores can vary because the entire hardware–software stack affects inference: GPU model, drivers/CUDA/cuDNN, PyTorch, vLLM/transformers (and commit hashes), flash-attention kernels, tokenizers/checkpoints, precision/quantization, and batching --- some of which are non-deterministic. We therefore request public submissions adhere to the same parameters and, in submission information, specify the GPUs and libraries versions they used for reproducibility purposes. 

\section*{Ethical Statement}
While the presented benchmark is able to comprehensively evaluate ``semantic'' abilities of the model, i.e., the capacity of individual models to understand and reason about data in different modalities, we did not perform explicit evaluation of any bias of these models, e.g., toward any kind of underrepresented minorities. In our opinion, this is an extremely important direction of future work, yet being outside the scope of our current contribution.

For the creation of novel datasets, the work of human annotators is used. We state that their work was adequately paid or compensated (see~\autoref{sec:appendix_hb} for the details).

Researchers participating in the
benchmark will be encouraged to adhere to ethical research practices, including proper citation,
acknowledgment of data sources, and responsible
reporting of results. Regular ethical reviews will assess the benchmark’s impact, identify potential ethical concerns, and implement necessary adjustments
to uphold the highest ethical standards throughout
the usage usage.

We proofread the text of this article using Overleaf Writefull assistant\footnote{\url{https://www.writefull.com}}, GPT-4o\footnote{\url{https://chatgpt.com}}, Grammarly\footnote{\url{https://app.grammarly.com}} to correct grammatical, spelling, and style errors and paraphrase sentences. We emphasize that these tools are used solely to enhance the quality of English writing, in full compliance with the ACL policies on the responsible use of AI writing assistance. Nevertheless, some segments of our publication can be potentially detected as AI-generated, AI-edited, or human-AI-generated.

\section*{Acknowledgments}
MERA is a collaborative project, thoughtfully designed to serve the needs of both industry and academia. The authors extend their sincere gratitude to our partners from the AI Alliance Russia, whose invaluable collaboration made an undertaking of this scale possible.
Special thanks are due to Ekaterina Morgunova, Yegor Nizamov, and Uliana Plugina for their significant contributions in coordinating our benchmark partners and contractors for the website development.

We also express our deep appreciation to the entire team dedicated to developing the website platform and maintaining the scoring system.

We are profoundly grateful to the following individuals for their dedication and hard work: 
Yaroslav Grebnyak, 
Anna Kostikova, 
Aleksandr Sautin, 
Artem Goryainov, 
Aleksandra Rak, 
Artem Goryainov, 
Albina Akhmetgareeva, 
Igor Churin, 
Leonid Sinev, 
Yulia Lazareva,
Ksenia Biryukova, 
Jamilya Erkenova,
Valentina Khlebutina, 
Maria Slabunova, 
Sergei Markov and to many others whom we may have inadvertently missed, but who supported us with their ideas, collaborated on dataset creation, validated results, and helped organize the workflow — your contributions are sincerely appreciated.

We also thank Zaryana Damashova and Ekaterina Artemova for their contributions to the creation of the RuSLUn dataset.

The work of Alexander Panchenko was supported by the RSF project 25-71-30008 ``Laboratory for reliable, adaptive, and trustworthy Artificial Intelligence''.

\bibliography{custom}

@inproceedings{yue2023mmmu,
  title={Mmmu: A massive multi-discipline multimodal understanding and reasoning benchmark for expert agi},
  author={Yue, Xiang and Ni, Yuansheng and Zhang, Kai and Zheng, Tianyu and Liu, Ruoqi and Zhang, Ge and Stevens, Samuel and Jiang, Dongfu and Ren, Weiming and Sun, Yuxuan and others},
  booktitle={Proceedings of the IEEE/CVF Conference on Computer Vision and Pattern Recognition},
  pages={9556--9567},
  year={2024}
}

@article{emelyanov2025fimmia,
  title={FiMMIA: scaling semantic perturbation-based membership inference across modalities},
  author={Emelyanov, Anton and Kudriashov, Sergei and Fenogenova, Alena},
  journal={arXiv preprint arXiv:2512.02786},
  year={2025}
}

@article{DBLP:journals/corr/abs-2503-15948,
  author       = {Elisei Rykov and
                  Kseniia Petrushina and
                  Kseniia Titova and
                  Alexander Panchenko and
                  Vasily Konovalov},
  title        = {Don't Fight Hallucinations, Use Them: Estimating Image Realism
                  using {NLI} over Atomic Facts},
  journal      = {CoRR},
  volume       = {abs/2503.15948},
  year         = {2025},
  url          = {https://doi.org/10.48550/arXiv.2503.15948},
  doi          = {10.48550/ARXIV.2503.15948},
  eprinttype    = {arXiv},
  eprint       = {2503.15948},
  bibsource    = {dblp computer science bibliography, https://dblp.org}
}

@inproceedings{rykov-etal-2025-looking,
    title = "Through the Looking Glass: Common Sense Consistency Evaluation of Weird Images",
    author = "Rykov, Elisei  and
      Petrushina, Kseniia  and
      Titova, Kseniia  and
      Razzhigaev, Anton  and
      Panchenko, Alexander  and
      Konovalov, Vasily",
    editor = "Ebrahimi, Abteen  and
      Haider, Samar  and
      Liu, Emmy  and
      Haider, Sammar  and
      Leonor Pacheco, Maria  and
      Wein, Shira",
    booktitle = "Proceedings of the 2025 Conference of the Nations of the Americas Chapter of the Association for Computational Linguistics: Human Language Technologies (Volume 4: Student Research Workshop)",
    month = apr,
    year = "2025",
    address = "Albuquerque, USA",
    publisher = "Association for Computational Linguistics",
    url = "https://aclanthology.org/2025.naacl-srw.28/",
    doi = "10.18653/v1/2025.naacl-srw.28",
    pages = "279--293",
    ISBN = "979-8-89176-192-6"
}

@misc{deepvk2025rumodernbert,
    title={RuModernBERT: Modernized BERT for Russian},
    author={Spirin, Egor and Malashenko, Boris and Sokolov Andrey},
    url={https://huggingface.co/deepvk/rumodernbert-base},
    publisher={Hugging Face},
    year={2025},
}

@inproceedings{voronov2024mind,
  title={Mind Your Format: Towards Consistent Evaluation of In-Context Learning Improvements},
  author={Voronov, Anton and Wolf, Lena and Ryabinin, Max},
  booktitle={Findings of the Association for Computational Linguistics ACL 2024},
  pages={6287--6310},
  year={2024}
}

@inproceedings{shokri2017membership,
  title={Membership inference attacks against machine learning models},
  author={Shokri, Reza and Stronati, Marco and Song, Congzheng and Shmatikov, Vitaly},
  booktitle={2017 IEEE symposium on security and privacy (SP)},
  pages={3--18},
  year={2017},
  organization={IEEE}
}

@inproceedings{li2023seed,
  title={SEED-Bench: Benchmarking Multimodal LLMs with Generative Comprehension},
  author={Li, Bohao and Wang, Rui and Wang, Guangzhi and Ge, Yuying and Ge, Yixiao and Shan, Ying},
  booktitle={Proceedings of the IEEE/CVF Conference on Computer Vision and Pattern Recognition},
  pages={19314--19327},
  year={2023}
}

@inproceedings{girdhar2023imagebind,
  author       = {Rohit Girdhar and
                  Alaaeldin El{-}Nouby and
                  Zhuang Liu and
                  Mannat Singh and
                  Kalyan Vasudev Alwala and
                  Armand Joulin and
                  Ishan Misra},
  title        = {ImageBind One Embedding Space to Bind Them All},
  booktitle    = {{IEEE/CVF} Conference on Computer Vision and Pattern Recognition,
                  {CVPR} 2023, Vancouver, BC, Canada, June 17-24, 2023},
  pages        = {15180--15190},
  publisher    = {{IEEE}},
  year         = {2023},
  url          = {https://doi.org/10.1109/CVPR52729.2023.01457},
  doi          = {10.1109/CVPR52729.2023.01457},
  bibsource    = {dblp computer science bibliography, https://dblp.org}
}

@inproceedings{DBLP:conf/cvpr/LiuLLL24,
  author       = {Haotian Liu and
                  Chunyuan Li and
                  Yuheng Li and
                  Yong Jae Lee},
  title        = {Improved Baselines with Visual Instruction Tuning},
  booktitle    = {{IEEE/CVF} Conference on Computer Vision and Pattern Recognition,
                  {CVPR} 2024, Seattle, WA, USA, June 16-22, 2024},
  pages        = {26286--26296},
  publisher    = {{IEEE}},
  year         = {2024},
  url          = {https://doi.org/10.1109/CVPR52733.2024.02484},
  doi          = {10.1109/CVPR52733.2024.02484},
  bibsource    = {dblp computer science bibliography, https://dblp.org}
}

@inproceedings{perlitz-etal-2024-efficient,
    title = "Efficient Benchmarking (of Language Models)",
    author = "Perlitz, Yotam  and
      Bandel, Elron  and
      Gera, Ariel  and
      Arviv, Ofir  and
      Ein-Dor, Liat  and
      Shnarch, Eyal  and
      Slonim, Noam  and
      Shmueli-Scheuer, Michal  and
      Choshen, Leshem",
    editor = "Duh, Kevin  and
      Gomez, Helena  and
      Bethard, Steven",
    booktitle = "Proceedings of the 2024 Conference of the North American Chapter of the Association for Computational Linguistics: Human Language Technologies (Volume 1: Long Papers)",
    month = jun,
    year = "2024",
    address = "Mexico City, Mexico",
    publisher = "Association for Computational Linguistics",
    url = "https://aclanthology.org/2024.naacl-long.139/",
    doi = "10.18653/v1/2024.naacl-long.139",
    pages = "2519--2536"
}

@inproceedings{DBLP:conf/acl/SuzgunSSGTCCLCZ23,
  author       = {Mirac Suzgun and
                  Nathan Scales and
                  Nathanael Sch{\"{a}}rli and
                  Sebastian Gehrmann and
                  Yi Tay and
                  Hyung Won Chung and
                  Aakanksha Chowdhery and
                  Quoc V. Le and
                  Ed H. Chi and
                  Denny Zhou and
                  Jason Wei},
  editor       = {Anna Rogers and
                  Jordan L. Boyd{-}Graber and
                  Naoaki Okazaki},
  title        = {Challenging BIG-Bench Tasks and Whether Chain-of-Thought Can Solve
                  Them},
  booktitle    = {Findings of the Association for Computational Linguistics: {ACL} 2023,
                  Toronto, Canada, July 9-14, 2023},
  pages        = {13003--13051},
  publisher    = {Association for Computational Linguistics},
  year         = {2023},
  url          = {https://doi.org/10.18653/v1/2023.findings-acl.824},
  doi          = {10.18653/V1/2023.FINDINGS-ACL.824},
  bibsource    = {dblp computer science bibliography, https://dblp.org}
}

@inproceedings{hendrycks2020measuring,
  author    = {Dan Hendrycks and Collin Burns and Steven Basart and Andy Zou and Mantas Mazeika and Dawn Song and Jacob Steinhardt},
  booktitle = {9th International Conference on Learning Representations, {ICLR} 2021, Virtual Event, Austria, May 3-7, 2021},
  title     = {{Measuring Massive Multitask Language Understanding}},
  year      = {2021},
  url       = {https://openreview.net/forum?id=d7KBjmI3GmQ},
}

@inproceedings{razzhigaev2024omnidialog,
  title={OmniDialog: A Multimodal Benchmark for Generalization Across Text, Visual, and Audio Modalities},
  author={Razzhigaev, Anton and Kurkin, Maxim and Goncharova, Elizaveta and Abdullaeva, Irina and Lysenko, Anastasia and Panchenko, Alexander and Kuznetsov, Andrey and Dimitrov, Denis},
  booktitle={Proceedings of the 2nd GenBench Workshop on Generalisation (Benchmarking) in NLP},
  pages={183--195},
  year={2024}
}

@inproceedings{fei2025path,
title={On Path to Multimodal Generalist: General-Level and General-Bench},
author={Hao Fei and Yuan Zhou and Juncheng Li and Xiangtai Li and Qingshan Xu and Bobo Li and Shengqiong Wu and Yaoting Wang and Junbao Zhou and Jiahao Meng and Qingyu Shi and Zhiyuan Zhou and Liangtao Shi and Minghe Gao and Daoan Zhang and Zhiqi Ge and Siliang Tang and Kaihang Pan and Yaobo Ye and Haobo Yuan and Tao Zhang and Weiming Wu and Tianjie Ju and Zixiang Meng and Shilin Xu and Liyu Jia and Wentao Hu and Meng Luo and Jiebo Luo and Tat-Seng Chua and Shuicheng YAN and Hanwang Zhang},
booktitle={Forty-second International Conference on Machine Learning},
year={2025},
url={https://openreview.net/forum?id=VsJ1K2HV3k}
}

@inproceedings{fu2025video,
  author       = {Chaoyou Fu and
                  Yuhan Dai and
                  Yongdong Luo and
                  Lei Li and
                  Shuhuai Ren and
                  Renrui Zhang and
                  Zihan Wang and
                  Chenyu Zhou and
                  Yunhang Shen and
                  Mengdan Zhang and
                  Peixian Chen and
                  Yanwei Li and
                  Shaohui Lin and
                  Sirui Zhao and
                  Ke Li and
                  Tong Xu and
                  Xiawu Zheng and
                  Enhong Chen and
                  Caifeng Shan and
                  Ran He and
                  Xing Sun},
  title        = {Video-MME: The First-Ever Comprehensive Evaluation Benchmark of Multi-modal
                  LLMs in Video Analysis},
  booktitle    = {{IEEE/CVF} Conference on Computer Vision and Pattern Recognition,
                  {CVPR} 2025, Nashville, TN, USA, June 11-15, 2025},
  pages        = {24108--24118},
  publisher    = {Computer Vision Foundation / {IEEE}},
  year         = {2025},
  url          = {https://openaccess.thecvf.com/content/CVPR2025/html/Fu\_Video-MME\_The\_First-Ever\_Comprehensive\_Evaluation\_Benchmark\_of\_Multi-modal\_LLMs\_in\_CVPR\_2025\_paper.html},
  doi          = {10.1109/CVPR52734.2025.02245},
  bibsource    = {dblp computer science bibliography, https://dblp.org}
}

@misc{ataallah2024infinibench,
      title={{InfiniBench}: A Comprehensive Benchmark for Large Multimodal Models in Very Long Video Understanding}, 
      author={Kirolos Ataallah and Chenhui Gou and Eslam Abdelrahman and Khushbu Pahwa and Jian Ding and Mohamed Elhoseiny},
      year={2024},
      eprint={2406.19875},
      archivePrefix={arXiv},
      primaryClass={cs.CV},
      url={https://arxiv.org/abs/2406.19875}, 
}

@inproceedings{wu2024star,
title={{STAR}: A Benchmark for Situated Reasoning in Real-World Videos},
 author = {Wu, Bo and Yu, Shoubin and Chen, Zhenfang and Tenenbaum, Josh and Gan, Chuang},
 booktitle = {Proceedings of the Neural Information Processing Systems Track on Datasets and Benchmarks},
 editor = {J. Vanschoren and S. Yeung},
 url = {https://datasets-benchmarks-proceedings.neurips.cc/paper_files/paper/2021/file/5ef059938ba799aaa845e1c2e8a762bd-Paper-round2.pdf},
 volume = {1},
 year = {2021}
}

@inproceedings{yang2021superb,
 author = {Shu{-}Wen Yang and
Po{-}Han Chi and
Yung{-}Sung Chuang and
Cheng{-}I Jeff Lai and
Kushal Lakhotia and
Yist Y. Lin and
Andy T. Liu and
Jiatong Shi and
Xuankai Chang and
Guan{-}Ting Lin and
Tzu{-}Hsien Huang and
Wei{-}Cheng Tseng and
Ko{-}tik Lee and
Da{-}Rong Liu and
Zili Huang and
Shuyan Dong and
Shang{-}Wen Li and
Shinji Watanabe and
Abdelrahman Mohamed and
Hung{-}yi Lee},
 booktitle = {Interspeech 2021, 22nd Annual Conference of the International Speech
Communication Association, Brno, Czechia, 30 August - 3 September
2021},
 doi = {10.21437/Interspeech.2021-1775},
 editor = {Hynek Hermansky and
Honza Cernock{\'{y}} and
Luk{\'{a}}s Burget and
Lori Lamel and
Odette Scharenborg and
Petr Motl{\'{\i}}cek},
 pages = {1194--1198},
 publisher = {{ISCA}},
 title = {{SUPERB:} Speech Processing Universal PERformance Benchmark},
 year = {2021}
}

@inproceedings{liu2024mmbench,
  author       = {Yuan Liu and
                  Haodong Duan and
                  Yuanhan Zhang and
                  Bo Li and
                  Songyang Zhang and
                  Wangbo Zhao and
                  Yike Yuan and
                  Jiaqi Wang and
                  Conghui He and
                  Ziwei Liu and
                  Kai Chen and
                  Dahua Lin},
  editor       = {Ales Leonardis and
                  Elisa Ricci and
                  Stefan Roth and
                  Olga Russakovsky and
                  Torsten Sattler and
                  G{\"{u}}l Varol},
  title        = {MMBench: Is Your Multi-modal Model an All-Around Player?},
  booktitle    = {Computer Vision - {ECCV} 2024 - 18th European Conference, Milan, Italy,
                  September 29-October 4, 2024, Proceedings, Part {VI}},
  series       = {Lecture Notes in Computer Science},
  volume       = {15064},
  pages        = {216--233},
  publisher    = {Springer},
  year         = {2024},
  url          = {https://doi.org/10.1007/978-3-031-72658-3\_13},
  doi          = {10.1007/978-3-031-72658-3\_13},
  bibsource    = {dblp computer science bibliography, https://dblp.org}
}

@inproceedings{liang2021multibench,
  author       = {Paul Pu Liang and
                  Yiwei Lyu and
                  Xiang Fan and
                  Zetian Wu and
                  Yun Cheng and
                  Jason Wu and
                  Leslie Chen and
                  Peter Wu and
                  Michelle A. Lee and
                  Yuke Zhu and
                  Ruslan Salakhutdinov and
                  Louis{-}Philippe Morency},
  editor       = {Joaquin Vanschoren and
                  Sai{-}Kit Yeung},
  title        = {MultiBench: Multiscale Benchmarks for Multimodal Representation Learning},
  booktitle    = {Proceedings of the Neural Information Processing Systems Track on
                  Datasets and Benchmarks 1, NeurIPS Datasets and Benchmarks 2021, December
                  2021, virtual},
  year         = {2021},
  url          = {https://datasets-benchmarks-proceedings.neurips.cc/paper/2021/hash/37693cfc748049e45d87b8c7d8b9aacd-Abstract-round1.html},
  bibsource    = {dblp computer science bibliography, https://dblp.org}
}

@inproceedings{taktasheva2022tape,
  author    = {Taktasheva, Ekaterina and Shavrina, Tatiana and Fenogenova, Alena and Shevelev, Denis and Katricheva, Nadezhda and Tikhonova, Maria and Akhmetgareeva, Albina and Zinkevich, Oleg and Bashmakova, Anastasiia and Iordanskaia, Svetlana and Spiridonova, Alena and Kurenshchikova, Valentina and Artemova, Ekaterina and Mikhailov, Vladislav},
  booktitle = {Findings of the Association for Computational Linguistics: EMNLP 2022},
  title     = {{TAPE}: Assessing Few-shot {R}ussian Language Understanding},
  year      = {2022},
  address   = {Abu Dhabi, United Arab Emirates},
  editor    = {Goldberg, Yoav and Kozareva, Zornitsa and Zhang, Yue},
  month     = dec,
  pages     = {2472--2497},
  publisher = {Association for Computational Linguistics},
  doi       = {10.18653/v1/2022.findings-emnlp.183},
}

@inproceedings{mikhailov2022rucola,
    title = "{R}u{C}o{LA}: {R}ussian Corpus of Linguistic Acceptability",
    author = "Mikhailov, Vladislav  and
      Shamardina, Tatiana  and
      Ryabinin, Max  and
      Pestova, Alena  and
      Smurov, Ivan  and
      Artemova, Ekaterina",
    editor = "Goldberg, Yoav  and
      Kozareva, Zornitsa  and
      Zhang, Yue",
    booktitle = "Proceedings of the 2022 Conference on Empirical Methods in Natural Language Processing",
    month = dec,
    year = "2022",
    address = "Abu Dhabi, United Arab Emirates",
    publisher = "Association for Computational Linguistics",
    url = "https://aclanthology.org/2022.emnlp-main.348/",
    doi = "10.18653/v1/2022.emnlp-main.348",
    pages = "5207--5227",
}

@misc{fenogenova2022russian,
      title={Russian {SuperGLUE} 1.1: Revising the Lessons not Learned by {R}ussian {NLP} models}, 
      author={Alena Fenogenova and Maria Tikhonova and Vladislav Mikhailov and Tatiana Shavrina and Anton Emelyanov and Denis Shevelev and Alexandr Kukushkin and Valentin Malykh and Ekaterina Artemova},
      year={2022},
      eprint={2202.07791},
      archivePrefix={arXiv},
      primaryClass={cs.CL},
      url={https://arxiv.org/abs/2202.07791}, 
}

@inproceedings{shavrina2020russiansuperglue,
  author    = {Shavrina, Tatiana and Fenogenova, Alena and Anton, Emelyanov and Shevelev, Denis and Artemova, Ekaterina and Malykh, Valentin and Mikhailov, Vladislav and Tikhonova, Maria and Chertok, Andrey and Evlampiev, Andrey},
  booktitle = {Proceedings of the 2020 Conference on Empirical Methods in Natural Language Processing (EMNLP)},
  title     = {{R}ussian{S}uper{GLUE}: A {R}ussian Language Understanding Evaluation Benchmark},
  year      = {2020},
  address   = {Online},
  pages     = {4717--4726},
  publisher = {Association for Computational Linguistics},
  doi       = {10.18653/v1/2020.emnlp-main.381},
}

@inproceedings{zhong2023agieval,
 address = {Mexico City, Mexico},
 author = {Zhong, Wanjun  and
Cui, Ruixiang  and
Guo, Yiduo  and
Liang, Yaobo  and
Lu, Shuai  and
Wang, Yanlin  and
Saied, Amin  and
Chen, Weizhu  and
Duan, Nan},
 booktitle = {Findings of the Association for Computational Linguistics: NAACL 2024},
 editor = {Duh, Kevin  and
Gomez, Helena  and
Bethard, Steven},
 pages = {2299--2314},
 publisher = {Association for Computational Linguistics},
 title = {{AGIE}val: A Human-Centric Benchmark for Evaluating Foundation Models},
 url = {https://aclanthology.org/2024.findings-naacl.149},
 year = {2024}
}

@inproceedings{huang2023c,
 author = {Yuzhen Huang and
Yuzhuo Bai and
Zhihao Zhu and
Junlei Zhang and
Jinghan Zhang and
Tangjun Su and
Junteng Liu and
Chuancheng Lv and
Yikai Zhang and
Jiayi Lei and
Yao Fu and
Maosong Sun and
Junxian He},
 booktitle = {Advances in Neural Information Processing Systems 36: Annual Conference
on Neural Information Processing Systems 2023, NeurIPS 2023, New Orleans,
LA, USA, December 10 - 16, 2023},
 editor = {Alice Oh and
Tristan Naumann and
Amir Globerson and
Kate Saenko and
Moritz Hardt and
Sergey Levine},
 title = {C-Eval: {A} Multi-Level Multi-Discipline Chinese Evaluation Suite
for Foundation Models},
 url = {http://papers.nips.cc/paper\_files/paper/2023/hash/c6ec1844bec96d6d32ae95ae694e23d8-Abstract-Datasets\_and\_Benchmarks.html},
 year = {2023}
}

@inproceedings{wang2018glue,
 author = {Alex Wang and
Amanpreet Singh and
Julian Michael and
Felix Hill and
Omer Levy and
Samuel R. Bowman},
 booktitle = {7th International Conference on Learning Representations, {ICLR} 2019,
New Orleans, LA, USA, May 6-9, 2019},
 publisher = {OpenReview.net},
 title = {{GLUE:} {A} Multi-Task Benchmark and Analysis Platform for Natural
Language Understanding},
 url = {https://openreview.net/forum?id=rJ4km2R5t7},
 year = {2019}
}

@inproceedings{wang2019superglue,
 author = {Alex Wang and
Yada Pruksachatkun and
Nikita Nangia and
Amanpreet Singh and
Julian Michael and
Felix Hill and
Omer Levy and
Samuel R. Bowman},
 booktitle = {Advances in Neural Information Processing Systems 32: Annual Conference
on Neural Information Processing Systems 2019, NeurIPS 2019, December
8-14, 2019, Vancouver, BC, Canada},
 editor = {Hanna M. Wallach and
Hugo Larochelle and
Alina Beygelzimer and
Florence d'Alch{\'{e}}{-}Buc and
Emily B. Fox and
Roman Garnett},
 pages = {3261--3275},
 title = {SuperGLUE: {A} Stickier Benchmark for General-Purpose Language Understanding
Systems},
 url = {https://proceedings.neurips.cc/paper/2019/hash/4496bf24afe7fab6f046bf4923da8de6-Abstract.html},
 year = {2019}
}

@inproceedings{fenogenova2024mera,
  author    = {Fenogenova, Alena and Chervyakov, Artem and Martynov, Nikita and Kozlova, Anastasia and Tikhonova, Maria and Akhmetgareeva, Albina and Emelyanov, Anton and Shevelev, Denis and Lebedev, Pavel and Sinev, Leonid and Isaeva, Ulyana and Kolomeytseva, Katerina and Moskovskiy, Daniil and Goncharova, Elizaveta and Savushkin, Nikita and Mikhailova, Polina and Minaeva, Anastasia and Dimitrov, Denis and Panchenko, Alexander and Markov, Sergey},
  booktitle = {Proceedings of the 62nd Annual Meeting of the Association for Computational Linguistics (Volume 1: Long Papers)},
  title     = {{MERA}: A Comprehensive {LLM} Evaluation in {R}ussian},
  year      = {2024},
  address   = {Bangkok, Thailand},
  editor    = {Ku, Lun-Wei and Martins, Andre and Srikumar, Vivek},
  month     = aug,
  pages     = {9920--9948},
  publisher = {Association for Computational Linguistics},
  doi       = {10.18653/v1/2024.acl-long.534},
}

@inproceedings{yang2024airbenchbenchmarkinglargeaudiolanguage,
    title = "{AIR}-Bench: Benchmarking Large Audio-Language Models via Generative Comprehension",
    author = "Yang, Qian  and
      Xu, Jin  and
      Liu, Wenrui  and
      Chu, Yunfei  and
      Jiang, Ziyue  and
      Zhou, Xiaohuan  and
      Leng, Yichong  and
      Lv, Yuanjun  and
      Zhao, Zhou  and
      Zhou, Chang  and
      Zhou, Jingren",
    editor = "Ku, Lun-Wei  and
      Martins, Andre  and
      Srikumar, Vivek",
    booktitle = "Proceedings of the 62nd Annual Meeting of the Association for Computational Linguistics (Volume 1: Long Papers)",
    month = aug,
    year = "2024",
    address = "Bangkok, Thailand",
    publisher = "Association for Computational Linguistics",
    url = "https://aclanthology.org/2024.acl-long.109/",
    doi = "10.18653/v1/2024.acl-long.109",
    pages = "1979--1998"
}

@inproceedings{wang2025audiobenchuniversalbenchmarkaudio,
    title = "{A}udio{B}ench: A Universal Benchmark for Audio Large Language Models",
    author = "Wang, Bin  and
      Zou, Xunlong  and
      Lin, Geyu  and
      Sun, Shuo  and
      Liu, Zhuohan  and
      Zhang, Wenyu  and
      Liu, Zhengyuan  and
      Aw, AiTi  and
      Chen, Nancy F.",
    editor = "Chiruzzo, Luis  and
      Ritter, Alan  and
      Wang, Lu",
    booktitle = "Proceedings of the 2025 Conference of the Nations of the Americas Chapter of the Association for Computational Linguistics: Human Language Technologies (Volume 1: Long Papers)",
    month = apr,
    year = "2025",
    address = "Albuquerque, New Mexico",
    publisher = "Association for Computational Linguistics",
    url = "https://aclanthology.org/2025.naacl-long.218/",
    doi = "10.18653/v1/2025.naacl-long.218",
    pages = "4297--4316",
    ISBN = "979-8-89176-189-6"
}

@inproceedings{
    sakshi2024mmaumassivemultitaskaudio,
    title={{MMAU}: A Massive Multi-Task Audio Understanding and Reasoning Benchmark},
    author={S Sakshi and Utkarsh Tyagi and Sonal Kumar and Ashish Seth and Ramaneswaran Selvakumar and Oriol Nieto and Ramani Duraiswami and Sreyan Ghosh and Dinesh Manocha},
    booktitle={The Thirteenth International Conference on Learning Representations},
    year={2025},
    url={https://openreview.net/forum?id=TeVAZXr3yv}
}

@inproceedings{johnson2016clevrdiagnosticdatasetcompositional,
  author       = {Justin Johnson and
                  Bharath Hariharan and
                  Laurens van der Maaten and
                  Li Fei{-}Fei and
                  C. Lawrence Zitnick and
                  Ross B. Girshick},
  title        = {{CLEVR:} {A} Diagnostic Dataset for Compositional Language and Elementary
                  Visual Reasoning},
  booktitle    = {2017 {IEEE} Conference on Computer Vision and Pattern Recognition,
                  {CVPR} 2017, Honolulu, HI, USA, July 21-26, 2017},
  pages        = {1988--1997},
  publisher    = {{IEEE} Computer Society},
  year         = {2017},
  url          = {https://doi.org/10.1109/CVPR.2017.215},
  doi          = {10.1109/CVPR.2017.215},
  bibsource    = {dblp computer science bibliography, https://dblp.org}
}

@inproceedings{balanced_vqa_v2,
  author       = {Yash Goyal and
                  Tejas Khot and
                  Douglas Summers{-}Stay and
                  Dhruv Batra and
                  Devi Parikh},
  title        = {Making the {V} in {VQA} Matter: Elevating the Role of Image Understanding
                  in Visual Question Answering},
  booktitle    = {2017 {IEEE} Conference on Computer Vision and Pattern Recognition,
                  {CVPR} 2017, Honolulu, HI, USA, July 21-26, 2017},
  pages        = {6325--6334},
  publisher    = {{IEEE} Computer Society},
  year         = {2017},
  url          = {https://doi.org/10.1109/CVPR.2017.670},
  doi          = {10.1109/CVPR.2017.670},
  bibsource    = {dblp computer science bibliography, https://dblp.org}
}

@inproceedings{lin2015microsoftcococommonobjects,
  author       = {Tsung{-}Yi Lin and
                  Michael Maire and
                  Serge J. Belongie and
                  James Hays and
                  Pietro Perona and
                  Deva Ramanan and
                  Piotr Doll{\'{a}}r and
                  C. Lawrence Zitnick},
  editor       = {David J. Fleet and
                  Tom{\'{a}}s Pajdla and
                  Bernt Schiele and
                  Tinne Tuytelaars},
  title        = {Microsoft {COCO:} Common Objects in Context},
  booktitle    = {Computer Vision - {ECCV} 2014 - 13th European Conference, Zurich,
                  Switzerland, September 6-12, 2014, Proceedings, Part {V}},
  series       = {Lecture Notes in Computer Science},
  volume       = {8693},
  pages        = {740--755},
  publisher    = {Springer},
  year         = {2014},
  url          = {https://doi.org/10.1007/978-3-319-10602-1\_48},
  doi          = {10.1007/978-3-319-10602-1\_48},
  bibsource    = {dblp computer science bibliography, https://dblp.org}
}

@inproceedings{lipping2022clothoaqacrowdsourceddatasetaudio,
  title={Clotho-aqa: A crowdsourced dataset for audio question answering},
  author={Lipping, Samuel and Sudarsanam, Parthasaarathy and Drossos, Konstantinos and Virtanen, Tuomas},
  booktitle={2022 30th European Signal Processing Conference (EUSIPCO)},
  pages={1140--1144},
  year={2022},
  organization={IEEE}
}

@inproceedings{li2022learninganswerquestionsdynamic,
  author       = {Guangyao Li and
                  Yake Wei and
                  Yapeng Tian and
                  Chenliang Xu and
                  Ji{-}Rong Wen and
                  Di Hu},
  title        = {Learning to Answer Questions in Dynamic Audio-Visual Scenarios},
  booktitle    = {{IEEE/CVF} Conference on Computer Vision and Pattern Recognition,
                  {CVPR} 2022, New Orleans, LA, USA, June 18-24, 2022},
  pages        = {19086--19096},
  publisher    = {{IEEE}},
  year         = {2022},
  url          = {https://doi.org/10.1109/CVPR52688.2022.01852},
  doi          = {10.1109/CVPR52688.2022.01852},
  bibsource    = {dblp computer science bibliography, https://dblp.org}
}

@inproceedings{lu2022learn,
  author       = {Pan Lu and
                  Swaroop Mishra and
                  Tanglin Xia and
                  Liang Qiu and
                  Kai{-}Wei Chang and
                  Song{-}Chun Zhu and
                  Oyvind Tafjord and
                  Peter Clark and
                  Ashwin Kalyan},
  editor       = {Sanmi Koyejo and
                  S. Mohamed and
                  A. Agarwal and
                  Danielle Belgrave and
                  K. Cho and
                  A. Oh},
  title        = {Learn to Explain: Multimodal Reasoning via Thought Chains for Science
                  Question Answering},
  booktitle    = {Advances in Neural Information Processing Systems 35: Annual Conference
                  on Neural Information Processing Systems 2022, NeurIPS 2022, New Orleans,
                  LA, USA, November 28 - December 9, 2022},
  year         = {2022},
  url          = {http://papers.nips.cc/paper\_files/paper/2022/hash/11332b6b6cf4485b84afadb1352d3a9a-Abstract-Conference.html},
  bibsource    = {dblp computer science bibliography, https://dblp.org}
}

@inproceedings{bastianelli2020slurpspokenlanguageunderstanding,
    title = "{SLURP}: A Spoken Language Understanding Resource Package",
    author = "Bastianelli, Emanuele  and
      Vanzo, Andrea  and
      Swietojanski, Pawel  and
      Rieser, Verena",
    editor = "Webber, Bonnie  and
      Cohn, Trevor  and
      He, Yulan  and
      Liu, Yang",
    booktitle = "Proceedings of the 2020 Conference on Empirical Methods in Natural Language Processing (EMNLP)",
    month = nov,
    year = "2020",
    address = "Online",
    publisher = "Association for Computational Linguistics",
    url = "https://aclanthology.org/2020.emnlp-main.588/",
    doi = "10.18653/v1/2020.emnlp-main.588",
    pages = "7252--7262"
}

@inproceedings{vandergoot2021maskedlanguagemodelingtranslation,
    title = "From Masked Language Modeling to Translation: Non-{E}nglish Auxiliary Tasks Improve Zero-shot Spoken Language Understanding",
    author = {van der Goot, Rob  and
      Sharaf, Ibrahim  and
      Imankulova, Aizhan  and
      {\"U}st{\"u}n, Ahmet  and
      Stepanovi{\'c}, Marija  and
      Ramponi, Alan  and
      Khairunnisa, Siti Oryza  and
      Komachi, Mamoru  and
      Plank, Barbara},
    editor = "Toutanova, Kristina  and
      Rumshisky, Anna  and
      Zettlemoyer, Luke  and
      Hakkani-Tur, Dilek  and
      Beltagy, Iz  and
      Bethard, Steven  and
      Cotterell, Ryan  and
      Chakraborty, Tanmoy  and
      Zhou, Yichao",
    booktitle = "Proceedings of the 2021 Conference of the North American Chapter of the Association for Computational Linguistics: Human Language Technologies",
    month = jun,
    year = "2021",
    address = "Online",
    publisher = "Association for Computational Linguistics",
    url = "https://aclanthology.org/2021.naacl-main.197/",
    doi = "10.18653/v1/2021.naacl-main.197",
    pages = "2479--2497"
}

@inproceedings{bittonguetta2023breakingcommonsensewhoops,
  title={Breaking common sense: Whoops! a vision-and-language benchmark of synthetic and compositional images},
  author={Bitton-Guetta, Nitzan and Bitton, Yonatan and Hessel, Jack and Schmidt, Ludwig and Elovici, Yuval and Stanovsky, Gabriel and Schwartz, Roy},
  booktitle={Proceedings of the IEEE/CVF International Conference on Computer Vision},
  pages={2616--2627},
  year={2023}
}

@inproceedings{roman2024proactivedetectionvoicecloning_watermarking,
author = {Roman, Robin San and Fernandez, Pierre and Elsahar, Hady and D\'{e}fossez, Alexandre and Furon, Teddy and Tran, Tuan},
title = {Proactive detection of voice cloning with localized watermarking},
year = {2024},
publisher = {JMLR.org},
booktitle = {Proceedings of the 41st International Conference on Machine Learning},
articleno = {1759},
numpages = {17},
location = {Vienna, Austria},
series = {ICML'24}
}

@inproceedings{
mozaffari2024semanticmembershipinferenceattack_smia,
title={Semantic Membership Inference Attack against Large Language Models},
author={Hamid Mozaffari and Virendra Marathe},
booktitle={Neurips Safe Generative AI Workshop 2024},
year={2024},
url={https://openreview.net/forum?id=I7S3Pf7Idl}
}

@misc{bai2025qwen25vltechnicalreport,
      title={Qwen2.5-VL Technical Report}, 
      author={Shuai Bai and Keqin Chen and Xuejing Liu and Jialin Wang and Wenbin Ge and Sibo Song and Kai Dang and Peng Wang and Shijie Wang and Jun Tang and Humen Zhong and Yuanzhi Zhu and Mingkun Yang and Zhaohai Li and Jianqiang Wan and Pengfei Wang and Wei Ding and Zheren Fu and Yiheng Xu and Jiabo Ye and Xi Zhang and Tianbao Xie and Zesen Cheng and Hang Zhang and Zhibo Yang and Haiyang Xu and Junyang Lin},
      year={2025},
      eprint={2502.13923},
      archivePrefix={arXiv},
      primaryClass={cs.CV},
      url={https://arxiv.org/abs/2502.13923}, 
}

@Misc{eval-harness,
  author       = {Gao, Leo and Tow, Jonathan and Abbasi, Baber and Biderman, Stella and Black, Sid and DiPofi, Anthony and Foster, Charles and Golding, Laurence and Hsu, Jeffrey and Le Noac'h, Alain and Li, Haonan and McDonell, Kyle and Muennighoff, Niklas and Ociepa, Chris and Phang, Jason and Reynolds, Laria and Schoelkopf, Hailey and Skowron, Aviya and Sutawika, Lintang and Tang, Eric and Thite, Anish and Wang, Ben and Wang, Kevin and Zou, Andy},
  howpublished = {Github},
  month        = sep,
  title        = {A~framework for few-shot language model evaluation},
  year         = {2024},
  doi          = {10.5281/zenodo.13694023},
  lastchecked  = {4 February 2025},
  publisher    = {Zenodo},
  url          = {https://github.com/EleutherAI/lm-evaluation-harness/tree/v0.4.4},
}

@Article{eval-harness-paper,
  author        = {Biderman, Stella and Schoelkopf, Hailey and Sutawika, Lintang and Gao, Leo and Tow, Jonathan and Abbasi, Baber and Aji, Alham Fikri and Ammanamanchi, Pawan Sasanka and Black, Sidney and Clive, Jordan and DiPofi, Anthony and Etxaniz, Julen and Fattori, Benjamin and Forde, Jessica Zosa and Foster, Charles and Hsu, Jeffrey and Jaiswal, Mimansa and Lee, Wilson Y. and Li, Haonan and Lovering, Charles and Muennighoff, Niklas and Pavlick, Ellie and Phang, Jason and Skowron, Aviya and Tan, Samson and Tang, Xiangru and Wang, Kevin A. and Winata, Genta Indra and Yvon, Fran{\c{c}}ois and Zou, Andy},
  journal       = {Preprint},
  title         = {Lessons from the Trenches on Reproducible Evaluation of Language Models},
  year          = {2024},
  month         = may,
  archiveprefix = {arXiv},
  eprint        = {2405.14782},
  primaryclass  = {cs.CL},
}

@misc{fu2024mmecomprehensiveevaluationbenchmark,
      title={MME: A Comprehensive Evaluation Benchmark for Multimodal Large Language Models}, 
      author={Chaoyou Fu and Peixian Chen and Yunhang Shen and Yulei Qin and Mengdan Zhang and Xu Lin and Jinrui Yang and Xiawu Zheng and Ke Li and Xing Sun and Yunsheng Wu and Rongrong Ji},
      year={2024},
      eprint={2306.13394},
      archivePrefix={arXiv},
      primaryClass={cs.CV},
      url={https://arxiv.org/abs/2306.13394}, 
}

@misc{abdin2024phi3technicalreporthighly,
      title={Phi-3 Technical Report: A Highly Capable Language Model Locally on Your Phone}, 
      author={Marah Abdin and Jyoti Aneja and Hany Awadalla and Ahmed Awadallah and Ammar Ahmad Awan and Nguyen Bach and Amit Bahree and Arash Bakhtiari and Jianmin Bao and Harkirat Behl and Alon Benhaim and Misha Bilenko and Johan Bjorck and Sébastien Bubeck and Martin Cai and Qin Cai and Vishrav Chaudhary and Dong Chen and Dongdong Chen and Weizhu Chen and Yen-Chun Chen and Yi-Ling Chen and Hao Cheng and Parul Chopra and Xiyang Dai and Matthew Dixon and Ronen Eldan and Victor Fragoso and Jianfeng Gao and Mei Gao and Min Gao and Amit Garg and Allie Del Giorno and Abhishek Goswami and Suriya Gunasekar and Emman Haider and Junheng Hao and Russell J. Hewett and Wenxiang Hu and Jamie Huynh and Dan Iter and Sam Ade Jacobs and Mojan Javaheripi and Xin Jin and Nikos Karampatziakis and Piero Kauffmann and Mahoud Khademi and Dongwoo Kim and Young Jin Kim and Lev Kurilenko and James R. Lee and Yin Tat Lee and Yuanzhi Li and Yunsheng Li and Chen Liang and Lars Liden and Xihui Lin and Zeqi Lin and Ce Liu and Liyuan Liu and Mengchen Liu and Weishung Liu and Xiaodong Liu and Chong Luo and Piyush Madan and Ali Mahmoudzadeh and David Majercak and Matt Mazzola and Caio César Teodoro Mendes and Arindam Mitra and Hardik Modi and Anh Nguyen and Brandon Norick and Barun Patra and Daniel Perez-Becker and Thomas Portet and Reid Pryzant and Heyang Qin and Marko Radmilac and Liliang Ren and Gustavo de Rosa and Corby Rosset and Sambudha Roy and Olatunji Ruwase and Olli Saarikivi and Amin Saied and Adil Salim and Michael Santacroce and Shital Shah and Ning Shang and Hiteshi Sharma and Yelong Shen and Swadheen Shukla and Xia Song and Masahiro Tanaka and Andrea Tupini and Praneetha Vaddamanu and Chunyu Wang and Guanhua Wang and Lijuan Wang and Shuohang Wang and Xin Wang and Yu Wang and Rachel Ward and Wen Wen and Philipp Witte and Haiping Wu and Xiaoxia Wu and Michael Wyatt and Bin Xiao and Can Xu and Jiahang Xu and Weijian Xu and Jilong Xue and Sonali Yadav and Fan Yang and Jianwei Yang and Yifan Yang and Ziyi Yang and Donghan Yu and Lu Yuan and Chenruidong Zhang and Cyril Zhang and Jianwen Zhang and Li Lyna Zhang and Yi Zhang and Yue Zhang and Yunan Zhang and Xiren Zhou},
      year={2024},
      eprint={2404.14219},
      archivePrefix={arXiv},
      primaryClass={cs.CL},
      url={https://arxiv.org/abs/2404.14219}, 
}

@misc{wang2024qwen2vlenhancingvisionlanguagemodels,
      title={Qwen2-VL: Enhancing Vision-Language Model's Perception of the World at Any Resolution}, 
      author={Peng Wang and Shuai Bai and Sinan Tan and Shijie Wang and Zhihao Fan and Jinze Bai and Keqin Chen and Xuejing Liu and Jialin Wang and Wenbin Ge and Yang Fan and Kai Dang and Mengfei Du and Xuancheng Ren and Rui Men and Dayiheng Liu and Chang Zhou and Jingren Zhou and Junyang Lin},
      year={2024},
      eprint={2409.12191},
      archivePrefix={arXiv},
      primaryClass={cs.CV},
      url={https://arxiv.org/abs/2409.12191}, 
}

@misc{gemmateam2025gemma3technicalreport,
      title={Gemma 3 Technical Report}, 
      author={Gemma Team and Aishwarya Kamath and Johan Ferret and Shreya Pathak and Nino Vieillard and Ramona Merhej and Sarah Perrin and Tatiana Matejovicova and Alexandre Ramé and Morgane Rivière and Louis Rouillard and Thomas Mesnard and Geoffrey Cideron and Jean-bastien Grill and Sabela Ramos and Edouard Yvinec and Michelle Casbon and Etienne Pot and Ivo Penchev and Gaël Liu and Francesco Visin and Kathleen Kenealy and Lucas Beyer and Xiaohai Zhai and Anton Tsitsulin and Robert Busa-Fekete and Alex Feng and Noveen Sachdeva and Benjamin Coleman and Yi Gao and Basil Mustafa and Iain Barr and Emilio Parisotto and David Tian and Matan Eyal and Colin Cherry and Jan-Thorsten Peter and Danila Sinopalnikov and Surya Bhupatiraju and Rishabh Agarwal and Mehran Kazemi and Dan Malkin and Ravin Kumar and David Vilar and Idan Brusilovsky and Jiaming Luo and Andreas Steiner and Abe Friesen and Abhanshu Sharma and Abheesht Sharma and Adi Mayrav Gilady and Adrian Goedeckemeyer and Alaa Saade and Alex Feng and Alexander Kolesnikov and Alexei Bendebury and Alvin Abdagic and Amit Vadi and András György and André Susano Pinto and Anil Das and Ankur Bapna and Antoine Miech and Antoine Yang and Antonia Paterson and Ashish Shenoy and Ayan Chakrabarti and Bilal Piot and Bo Wu and Bobak Shahriari and Bryce Petrini and Charlie Chen and Charline Le Lan and Christopher A. Choquette-Choo and CJ Carey and Cormac Brick and Daniel Deutsch and Danielle Eisenbud and Dee Cattle and Derek Cheng and Dimitris Paparas and Divyashree Shivakumar Sreepathihalli and Doug Reid and Dustin Tran and Dustin Zelle and Eric Noland and Erwin Huizenga and Eugene Kharitonov and Frederick Liu and Gagik Amirkhanyan and Glenn Cameron and Hadi Hashemi and Hanna Klimczak-Plucińska and Harman Singh and Harsh Mehta and Harshal Tushar Lehri and Hussein Hazimeh and Ian Ballantyne and Idan Szpektor and Ivan Nardini and Jean Pouget-Abadie and Jetha Chan and Joe Stanton and John Wieting and Jonathan Lai and Jordi Orbay and Joseph Fernandez and Josh Newlan and Ju-yeong Ji and Jyotinder Singh and Kat Black and Kathy Yu and Kevin Hui and Kiran Vodrahalli and Klaus Greff and Linhai Qiu and Marcella Valentine and Marina Coelho and Marvin Ritter and Matt Hoffman and Matthew Watson and Mayank Chaturvedi and Michael Moynihan and Min Ma and Nabila Babar and Natasha Noy and Nathan Byrd and Nick Roy and Nikola Momchev and Nilay Chauhan and Noveen Sachdeva and Oskar Bunyan and Pankil Botarda and Paul Caron and Paul Kishan Rubenstein and Phil Culliton and Philipp Schmid and Pier Giuseppe Sessa and Pingmei Xu and Piotr Stanczyk and Pouya Tafti and Rakesh Shivanna and Renjie Wu and Renke Pan and Reza Rokni and Rob Willoughby and Rohith Vallu and Ryan Mullins and Sammy Jerome and Sara Smoot and Sertan Girgin and Shariq Iqbal and Shashir Reddy and Shruti Sheth and Siim Põder and Sijal Bhatnagar and Sindhu Raghuram Panyam and Sivan Eiger and Susan Zhang and Tianqi Liu and Trevor Yacovone and Tyler Liechty and Uday Kalra and Utku Evci and Vedant Misra and Vincent Roseberry and Vlad Feinberg and Vlad Kolesnikov and Woohyun Han and Woosuk Kwon and Xi Chen and Yinlam Chow and Yuvein Zhu and Zichuan Wei and Zoltan Egyed and Victor Cotruta and Minh Giang and Phoebe Kirk and Anand Rao and Kat Black and Nabila Babar and Jessica Lo and Erica Moreira and Luiz Gustavo Martins and Omar Sanseviero and Lucas Gonzalez and Zach Gleicher and Tris Warkentin and Vahab Mirrokni and Evan Senter and Eli Collins and Joelle Barral and Zoubin Ghahramani and Raia Hadsell and Yossi Matias and D. Sculley and Slav Petrov and Noah Fiedel and Noam Shazeer and Oriol Vinyals and Jeff Dean and Demis Hassabis and Koray Kavukcuoglu and Clement Farabet and Elena Buchatskaya and Jean-Baptiste Alayrac and Rohan Anil and Dmitry and Lepikhin and Sebastian Borgeaud and Olivier Bachem and Armand Joulin and Alek Andreev and Cassidy Hardin and Robert Dadashi and Léonard Hussenot},
      year={2025},
      eprint={2503.19786},
      archivePrefix={arXiv},
      primaryClass={cs.CL},
      url={https://arxiv.org/abs/2503.19786}, 
}

@inproceedings{liu2024improvedllava,
  author       = {Haotian Liu and
                  Chunyuan Li and
                  Yuheng Li and
                  Yong Jae Lee},
  title        = {Improved Baselines with Visual Instruction Tuning},
  booktitle    = {{IEEE/CVF} Conference on Computer Vision and Pattern Recognition,
                  {CVPR} 2024, Seattle, WA, USA, June 16-22, 2024},
  pages        = {26286--26296},
  publisher    = {{IEEE}},
  year         = {2024},
  url          = {https://doi.org/10.1109/CVPR52733.2024.02484},
  doi          = {10.1109/CVPR52733.2024.02484},
  bibsource    = {dblp computer science bibliography, https://dblp.org}
}

@misc{liu2024llavanext,
    title={LLaVA-NeXT: Improved reasoning, OCR, and world knowledge},
    url={https://llava-vl.github.io/blog/2024-01-30-llava-next/},
    author={Liu, Haotian and Li, Chunyuan and Li, Yuheng and Li, Bo and Zhang, Yuanhan and Shen, Sheng and Lee, Yong Jae},
    month={January},
    year={2024}
}

@misc{chu2024qwen2audiotechnicalreport,
      title={Qwen2-Audio Technical Report}, 
      author={Yunfei Chu and Jin Xu and Qian Yang and Haojie Wei and Xipin Wei and Zhifang Guo and Yichong Leng and Yuanjun Lv and Jinzheng He and Junyang Lin and Chang Zhou and Jingren Zhou},
      year={2024},
      eprint={2407.10759},
      archivePrefix={arXiv},
      primaryClass={eess.AS},
      url={https://arxiv.org/abs/2407.10759}, 
}

@misc{team2024ultravox,
    title={Ultravox: An open-weight alternative to GPT-4o Realtime},
    url={https://www.ultravox.ai/blog/ultravox-an-open-weight-alternative-to-gpt-4o-realtime},
    author={Team},
    month={November},
    year={2024}
}

@article{wang2025internvl3,
  title={Internvl3. 5: Advancing open-source multimodal models in versatility, reasoning, and efficiency},
  author={Wang, Weiyun and Gao, Zhangwei and Gu, Lixin and Pu, Hengjun and Cui, Long and Wei, Xingguang and Liu, Zhaoyang and Jing, Linglin and Ye, Shenglong and Shao, Jie and others},
  journal={arXiv preprint arXiv:2508.18265},
  year={2025}
}

@article{zhu2025internvl3,
  title={Internvl3: Exploring advanced training and test-time recipes for open-source multimodal models},
  author={Zhu, Jinguo and Wang, Weiyun and Chen, Zhe and Liu, Zhaoyang and Ye, Shenglong and Gu, Lixin and Tian, Hao and Duan, Yuchen and Su, Weijie and Shao, Jie and others},
  journal={arXiv preprint arXiv:2504.10479},
  year={2025}
}

@inproceedings{zmitrovich2023family,
    title = "A Family of Pretrained Transformer Language Models for {R}ussian",
    author = "Zmitrovich, Dmitry  and
      Abramov, Aleksandr  and
      Kalmykov, Andrey  and
      Kadulin, Vitaly  and
      Tikhonova, Maria  and
      Taktasheva, Ekaterina  and
      Astafurov, Danil  and
      Baushenko, Mark  and
      Snegirev, Artem  and
      Shavrina, Tatiana  and
      Markov, Sergei S.  and
      Mikhailov, Vladislav  and
      Fenogenova, Alena",
    editor = "Calzolari, Nicoletta  and
      Kan, Min-Yen  and
      Hoste, Veronique  and
      Lenci, Alessandro  and
      Sakti, Sakriani  and
      Xue, Nianwen",
    booktitle = "Proceedings of the 2024 Joint International Conference on Computational Linguistics, Language Resources and Evaluation (LREC-COLING 2024)",
    month = may,
    year = "2024",
    address = "Torino, Italia",
    publisher = "ELRA and ICCL",
    url = "https://aclanthology.org/2024.lrec-main.45/",
    pages = "507--524",
}

@article{muennighoff2022mteb,
  title={Mteb: Massive text embedding benchmark},
  author={Muennighoff, Niklas and Tazi, Nouamane and Magne, Lo{\"\i}c and Reimers, Nils},
  journal={arXiv preprint arXiv:2210.07316},
  year={2022}
}

@InProceedings{wolf-etal-2020-transformers,
  author    = {Wolf, Thomas and Debut, Lysandre and Sanh, Victor and Chaumond, Julien and Delangue, Clement and Moi, Anthony and Cistac, Pierric and Rault, Tim and Louf, Remi and Funtowicz, Morgan and Davison, Joe and Shleifer, Sam and von Platen, Patrick and Ma, Clara and Jernite, Yacine and Plu, Julien and Xu, Canwen and Le Scao, Teven and Gugger, Sylvain and Drame, Mariama and Lhoest, Quentin and Rush, Alexander},
  booktitle = {Proceedings of the 2020 Conference on Empirical Methods in Natural Language Processing: System Demonstrations},
  title     = {Transformers: {S}tate-of-the-Art Natural Language Processing},
  year      = {2020},
  address   = {Online},
  editor    = {Liu, Qun and Schlangen, David},
  month     = oct,
  pages     = {38--45},
  publisher = {Association for Computational Linguistics},
  doi       = {10.18653/v1/2020.emnlp-demos.6},
  url       = {https://aclanthology.org/2020.emnlp-demos.6},
}

@misc{microsoft2025phi4minitechnicalreportcompact,
      title={Phi-4-Mini Technical Report: Compact yet Powerful Multimodal Language Models via Mixture-of-LoRAs}, 
      author={Microsoft and : and Abdelrahman Abouelenin and Atabak Ashfaq and Adam Atkinson and Hany Awadalla and Nguyen Bach and Jianmin Bao and Alon Benhaim and Martin Cai and Vishrav Chaudhary and Congcong Chen and Dong Chen and Dongdong Chen and Junkun Chen and Weizhu Chen and Yen-Chun Chen and Yi-ling Chen and Qi Dai and Xiyang Dai and Ruchao Fan and Mei Gao and Min Gao and Amit Garg and Abhishek Goswami and Junheng Hao and Amr Hendy and Yuxuan Hu and Xin Jin and Mahmoud Khademi and Dongwoo Kim and Young Jin Kim and Gina Lee and Jinyu Li and Yunsheng Li and Chen Liang and Xihui Lin and Zeqi Lin and Mengchen Liu and Yang Liu and Gilsinia Lopez and Chong Luo and Piyush Madan and Vadim Mazalov and Arindam Mitra and Ali Mousavi and Anh Nguyen and Jing Pan and Daniel Perez-Becker and Jacob Platin and Thomas Portet and Kai Qiu and Bo Ren and Liliang Ren and Sambuddha Roy and Ning Shang and Yelong Shen and Saksham Singhal and Subhojit Som and Xia Song and Tetyana Sych and Praneetha Vaddamanu and Shuohang Wang and Yiming Wang and Zhenghao Wang and Haibin Wu and Haoran Xu and Weijian Xu and Yifan Yang and Ziyi Yang and Donghan Yu and Ishmam Zabir and Jianwen Zhang and Li Lyna Zhang and Yunan Zhang and Xiren Zhou},
      year={2025},
      eprint={2503.01743},
      archivePrefix={arXiv},
      primaryClass={cs.CL},
      url={https://arxiv.org/abs/2503.01743}, 
}

@misc{granitevisionteam2025granitevisionlightweightopensource,
      title={Granite Vision: a lightweight, open-source multimodal model for enterprise Intelligence}, 
      author={Granite Vision Team and Leonid Karlinsky and Assaf Arbelle and Abraham Daniels and Ahmed Nassar and Amit Alfassi and Bo Wu and Eli Schwartz and Dhiraj Joshi and Jovana Kondic and Nimrod Shabtay and Pengyuan Li and Roei Herzig and Shafiq Abedin and Shaked Perek and Sivan Harary and Udi Barzelay and Adi Raz Goldfarb and Aude Oliva and Ben Wieles and Bishwaranjan Bhattacharjee and Brandon Huang and Christoph Auer and Dan Gutfreund and David Beymer and David Wood and Hilde Kuehne and Jacob Hansen and Joseph Shtok and Ken Wong and Luis Angel Bathen and Mayank Mishra and Maksym Lysak and Michele Dolfi and Mikhail Yurochkin and Nikolaos Livathinos and Nimrod Harel and Ophir Azulai and Oshri Naparstek and Rafael Teixeira de Lima and Rameswar Panda and Sivan Doveh and Shubham Gupta and Subhro Das and Syed Zawad and Yusik Kim and Zexue He and Alexander Brooks and Gabe Goodhart and Anita Govindjee and Derek Leist and Ibrahim Ibrahim and Aya Soffer and David Cox and Kate Soule and Luis Lastras and Nirmit Desai and Shila Ofek-koifman and Sriram Raghavan and Tanveer Syeda-Mahmood and Peter Staar and Tal Drory and Rogerio Feris},
      year={2025},
      eprint={2502.09927},
      archivePrefix={arXiv},
      primaryClass={cs.CV},
      url={https://arxiv.org/abs/2502.09927}, 
}

@misc{xu2025qwen25omnitechnicalreport,
      title={Qwen2.5-Omni Technical Report}, 
      author={Jin Xu and Zhifang Guo and Jinzheng He and Hangrui Hu and Ting He and Shuai Bai and Keqin Chen and Jialin Wang and Yang Fan and Kai Dang and Bin Zhang and Xiong Wang and Yunfei Chu and Junyang Lin},
      year={2025},
      eprint={2503.20215},
      archivePrefix={arXiv},
      primaryClass={cs.CL},
      url={https://arxiv.org/abs/2503.20215}, 
}

@misc{marafioti2025smolvlmredefiningsmallefficient,
      title={SmolVLM: Redefining small and efficient multimodal models}, 
      author={Andrés Marafioti and Orr Zohar and Miquel Farré and Merve Noyan and Elie Bakouch and Pedro Cuenca and Cyril Zakka and Loubna Ben Allal and Anton Lozhkov and Nouamane Tazi and Vaibhav Srivastav and Joshua Lochner and Hugo Larcher and Mathieu Morlon and Lewis Tunstall and Leandro von Werra and Thomas Wolf},
      year={2025},
      eprint={2504.05299},
      archivePrefix={arXiv},
      primaryClass={cs.AI},
      url={https://arxiv.org/abs/2504.05299}, 
}

@misc{yang2025qwen3technicalreport,
      title={Qwen3 Technical Report}, 
      author={An Yang and Anfeng Li and Baosong Yang and Beichen Zhang and Binyuan Hui and Bo Zheng and Bowen Yu and Chang Gao and Chengen Huang and Chenxu Lv and Chujie Zheng and Dayiheng Liu and Fan Zhou and Fei Huang and Feng Hu and Hao Ge and Haoran Wei and Huan Lin and Jialong Tang and Jian Yang and Jianhong Tu and Jianwei Zhang and Jianxin Yang and Jiaxi Yang and Jing Zhou and Jingren Zhou and Junyang Lin and Kai Dang and Keqin Bao and Kexin Yang and Le Yu and Lianghao Deng and Mei Li and Mingfeng Xue and Mingze Li and Pei Zhang and Peng Wang and Qin Zhu and Rui Men and Ruize Gao and Shixuan Liu and Shuang Luo and Tianhao Li and Tianyi Tang and Wenbiao Yin and Xingzhang Ren and Xinyu Wang and Xinyu Zhang and Xuancheng Ren and Yang Fan and Yang Su and Yichang Zhang and Yinger Zhang and Yu Wan and Yuqiong Liu and Zekun Wang and Zeyu Cui and Zhenru Zhang and Zhipeng Zhou and Zihan Qiu},
      year={2025},
      eprint={2505.09388},
      archivePrefix={arXiv},
      primaryClass={cs.CL},
      url={https://arxiv.org/abs/2505.09388}, 
}

@misc{yao2024minicpmvgpt4vlevelmllm,
      title={MiniCPM-V: A GPT-4V Level MLLM on Your Phone}, 
      author={Yuan Yao and Tianyu Yu and Ao Zhang and Chongyi Wang and Junbo Cui and Hongji Zhu and Tianchi Cai and Haoyu Li and Weilin Zhao and Zhihui He and Qianyu Chen and Huarong Zhou and Zhensheng Zou and Haoye Zhang and Shengding Hu and Zhi Zheng and Jie Zhou and Jie Cai and Xu Han and Guoyang Zeng and Dahai Li and Zhiyuan Liu and Maosong Sun},
      year={2024},
      eprint={2408.01800},
      archivePrefix={arXiv},
      primaryClass={cs.CV},
      url={https://arxiv.org/abs/2408.01800}, 
}

@misc{li2024llavanext-strong,
    title={LLaVA-NeXT: Stronger LLMs Supercharge Multimodal Capabilities in the Wild},
    url={https://llava-vl.github.io/blog/2024-05-10-llava-next-stronger-llms/},
    author={Li, Bo and Zhang, Kaichen and Zhang, Hao and Guo, Dong and Zhang, Renrui and Li, Feng and Zhang, Yuanhan and Liu, Ziwei and Li, Chunyuan},
    month={May},
    year={2024}
}

@misc{goel2025audioflamingo3advancing,
      title={Audio Flamingo 3: Advancing Audio Intelligence with Fully Open Large Audio Language Models}, 
      author={Arushi Goel and Sreyan Ghosh and Jaehyeon Kim and Sonal Kumar and Zhifeng Kong and Sang-gil Lee and Chao-Han Huck Yang and Ramani Duraiswami and Dinesh Manocha and Rafael Valle and Bryan Catanzaro},
      year={2025},
      eprint={2507.08128},
      archivePrefix={arXiv},
      primaryClass={cs.SD},
      url={https://arxiv.org/abs/2507.08128}, 
}

@misc{zhang2024seallms3openfoundation,
      title={SeaLLMs 3: Open Foundation and Chat Multilingual Large Language Models for Southeast Asian Languages}, 
      author={Wenxuan Zhang and Hou Pong Chan and Yiran Zhao and Mahani Aljunied and Jianyu Wang and Chaoqun Liu and Yue Deng and Zhiqiang Hu and Weiwen Xu and Yew Ken Chia and Xin Li and Lidong Bing},
      year={2024},
      eprint={2407.19672},
      archivePrefix={arXiv},
      primaryClass={cs.CL},
      url={https://arxiv.org/abs/2407.19672}, 
}

@misc{wang2024divideconquercombine,
  title        = {Divide, Conquer and Combine: A Training-Free Framework for High-Resolution Image Perception in Multimodal Large Language Models},
  author       = {Wenbin Wang and Liang Ding and Minyan Zeng and Xiabin Zhou and Li Shen and Yong Luo and Dacheng Tao},
  year         = {2024},
  eprint       = {2408.15556},
  archivePrefix= {arXiv},
  primaryClass = {cs.CV},
  url          = {https://arxiv.org/abs/2408.15556}
}

@inproceedings{methani2020plotqa,
  author    = {Methani, Nitesh and Ganguly, Pritha and Khapra, Mitesh M. and Kumar, Pratyush},
  title     = {PlotQA: Reasoning over Scientific Plots},
  booktitle = {Proceedings of the IEEE Winter Conference on Applications of Computer Vision (WACV)},
  month     = mar,
  year      = {2020},
  url       = {https://arxiv.org/abs/1909.00997}
}

@inproceedings{masry2022chartqa,
  title     = {ChartQA: A Benchmark for Question Answering about Charts with Visual and Logical Reasoning},
  author    = {Masry, Ahmed and Long, Do Xuan and Tan, Jia Qing and Joty, Shafiq R. and Hoque, Enamul},
  booktitle = {Findings of the Association for Computational Linguistics: ACL 2022},
  month     = may,
  year      = {2022},
  address   = {Dublin, Ireland},
  publisher = {Association for Computational Linguistics},
  pages     = {2263--2279},
  url       = {https://aclanthology.org/2022.findings-acl.177/},
  doi       = {10.18653/v1/2022.findings-acl.177}
}

@inproceedings{singh2019textvqa,
  author    = {Singh, Amanpreet and Natarajan, Vivek and Shah, Meet and Jiang, Yu and Chen, Xinlei and Batra, Dhruv and Parikh, Devi and Rohrbach, Marcus},
  title     = {Towards {VQA} Models That Can Read},
  booktitle = {Proceedings of the IEEE/CVF Conference on Computer Vision and Pattern Recognition (CVPR)},
  month     = jun,
  year      = {2019},
  pages     = {8317--8326},
  doi       = {10.1109/CVPR.2019.00851},
  url       = {https://openaccess.thecvf.com/content_CVPR_2019/html/Singh_Towards_VQA_Models_That_Can_Read_CVPR_2019_paper.html}
}

@inproceedings{chen-etal-2025-unmasking,
    title = "Unmasking Deceptive Visuals: Benchmarking Multimodal Large Language Models on Misleading Chart Question Answering",
    author = "Chen, Zixin  and
      Song, Sicheng  and
      Shum, KaShun  and
      Lin, Yanna  and
      Sheng, Rui  and
      Wang, Weiqi  and
      Qu, Huamin",
    editor = "Christodoulopoulos, Christos  and
      Chakraborty, Tanmoy  and
      Rose, Carolyn  and
      Peng, Violet",
    booktitle = "Proceedings of the 2025 Conference on Empirical Methods in Natural Language Processing",
    month = nov,
    year = "2025",
    address = "Suzhou, China",
    publisher = "Association for Computational Linguistics",
    url = "https://aclanthology.org/2025.emnlp-main.695/",
    doi = "10.18653/v1/2025.emnlp-main.695",
    pages = "13756--13789",
    ISBN = "979-8-89176-332-6"
}

@article{ChartBench,
    title={ChartBench: A Benchmark for Complex Visual Reasoning in Charts},
    author={Zhengzhuo Xu and Sinan Du and Yiyan Qi and Chengjin Xu and Chun Yuan and Jian Guo},
    journal={ArXiv},
    year={2023},
    volume={abs/2312.15915},
    url={https://api.semanticscholar.org/CorpusID:266550948}
}

@inproceedings{lu2024mathvista,
  title     = {MathVista: Evaluating Mathematical Reasoning of Foundation Models in Visual Contexts},
  author    = {Lu, Pan and Bansal, Hritik and Xia, Tony and Liu, Jiacheng and Li, Chunyuan and Hajishirzi, Hannaneh and Cheng, Hao and Chang, Kai-Wei and Galley, Michel and Gao, Jianfeng},
  booktitle = {International Conference on Learning Representations (ICLR)},
  year      = {2024},
  url       = {https://openreview.net/forum?id=KUNzEQMWU7}
}

@inproceedings{chen2024mmstar,
  title     = {Are We on the Right Way for Evaluating Large Vision-Language Models?},
  author    = {Chen, Lin and Li, Jinsong and Dong, Xiaoyi and Zhang, Pan and Zang, Yuhang and Chen, Zehui and Duan, Haodong and Wang, Jiaqi and Qiao, Yu and Lin, Dahua and Zhao, Feng},
  booktitle = {Advances in Neural Information Processing Systems},
  year      = {2024},
  volume    = {37},
  url       = {https://proceedings.neurips.cc/paper_files/paper/2024/hash/2f8ee6a3d766b426d2618e555b5aeb39-Abstract-Conference.html}
}

@inproceedings{liu2024convbench,
  title     = {ConvBench: A Multi-Turn Conversation Evaluation Benchmark with Hierarchical Ablation Capability for Large Vision-Language Models},
  author    = {Liu, Shuo and Ying, Kaining and Zhang, Hao and Yang, Yue and Lin, Yuqi and Zhang, Tianle and Li, Chuanhao and Qiao, Yu and Luo, Ping and Shao, Wenqi and Zhang, Kaipeng},
  booktitle = {Advances in Neural Information Processing Systems},
  year      = {2024},
  volume    = {37},
  url       = {https://proceedings.neurips.cc/paper_files/paper/2024/hash/b69396afc07a9ca3428d194f4db84c02-Abstract-Datasets_and_Benchmarks_Track.html}
}

@inproceedings{okamoto2020rwcp,
  title     = {{RWCP-SSD-Onomatopoeia}: Onomatopoeic Word Dataset for Environmental Sound Synthesis},
  author    = {Okamoto, Yuki and Imoto, Keisuke and Takamichi, Shinnosuke and Yamanishi, Ryosuke and Fukumori, Takahiro and Yamashita, Yoichi},
  booktitle = {Proceedings of the 5th Workshop on Detection and Classification of Acoustic Scenes and Events (DCASE 2020)},
  year      = {2020},
  month     = nov,
  pages     = {125--129},
  publisher = {Zenodo},
  doi       = {10.5281/zenodo.4061782},
  url       = {https://doi.org/10.5281/zenodo.4061782}
}

@inproceedings{li2025corecognition,
  title     = {Core Knowledge Deficits in Multi-Modal Language Models},
  author    = {Li, Yijiang and Gao, Qingying and Zhao, Tianwei and Wang, Bingyang and Sun, Haoran and Lyu, Haiyun and Hawkins, Robert D. and Vasconcelos, Nuno and Golan, Tal and Luo, Dezhi and Deng, Hokin},
  booktitle = {Proceedings of the 42nd International Conference on Machine Learning},
  series    = {Proceedings of Machine Learning Research},
  volume    = {267},
  pages     = {34379--34409},
  year      = {2025},
  editor    = {Singh, Aarti and Fazel, Maryam and Hsu, Daniel and Lacoste-Julien, Simon and Berkenkamp, Felix and Maharaj, Tegan and Wagstaff, Kiri and Zhu, Jerry},
  month     = jul,
  publisher = {PMLR},
  url       = {https://proceedings.mlr.press/v267/li25p.html}
}

@inproceedings{andriluka20142d,
  author    = {Andriluka, Mykhaylo and Pishchulin, Leonid and Gehler, Peter V. and Schiele, Bernt},
  title     = {2D Human Pose Estimation: New Benchmark and State of the Art Analysis},
  booktitle = {Proceedings of the IEEE Conference on Computer Vision and Pattern Recognition (CVPR)},
  month     = jun,
  year      = {2014},
  pages     = {3686--3693},
  doi       = {10.1109/CVPR.2014.471},
  url       = {https://openaccess.thecvf.com/content_cvpr_2014/html/Andriluka_2D_Human_Pose_2014_CVPR_paper.html}
}

@inproceedings{andriluka2018posetrack,
  author    = {Andriluka, Mykhaylo and Iqbal, Umar and Insafutdinov, Eldar and Pishchulin, Leonid and Milan, Anton and Gall, Juergen and Schiele, Bernt},
  title     = {PoseTrack: A Benchmark for Human Pose Estimation and Tracking},
  booktitle = {Proceedings of the IEEE Conference on Computer Vision and Pattern Recognition (CVPR)},
  month     = jun,
  year      = {2018},
  pages     = {5167--5176},
  doi       = {10.1109/CVPR.2018.00542},
  url       = {https://openaccess.thecvf.com/content_cvpr_2018/html/Andriluka_PoseTrack_A_Benchmark_CVPR_2018_paper.html}
}

@inproceedings{kapitanov2024hagrid,
  author    = {Kapitanov, Alexander and Kvanchiani, Karina and Nagaev, Alexander and Kraynov, Roman and Makhliarchuk, Andrei},
  title     = {HaGRID -- HAnd Gesture Recognition Image Dataset},
  booktitle = {Proceedings of the IEEE/CVF Winter Conference on Applications of Computer Vision (WACV)},
  month     = jan,
  year      = {2024},
  pages     = {4572--4581},
  doi       = {10.1109/WACV57701.2024.00451},
  url       = {https://openaccess.thecvf.com/content/WACV2024/html/Kapitanov_HaGRID_--_HAnd_Gesture_Recognition_Image_Dataset_WACV_2024_paper.html}
}

\appendix

\section*{Appendix}
\label{sec:appendix}

\section{Dataset Description}
\label{sec:appendix_datasets}

\lstdefinelanguage{json}{
    morestring=[b]",
    morecomment=[l]{//},
    morekeywords={true,false,null},
    sensitive=false,
}

\lstset{
    language=json,
    inputencoding=utf8,
    extendedchars=true,
    basicstyle=\ttfamily\small,
    showstringspaces=false,
    breaklines=true,
    keepspaces=true,
    literate=
      {А}{{\selectfont\char192}}1
      {В}{{\selectfont\char194}}1
      {Е}{{\selectfont\char197}}1
      {К}{{\selectfont\char202}}1
      {М}{{\selectfont\char204}}1
      {Н}{{\selectfont\char205}}1
      {О}{{\selectfont\char206}}1
      {Р}{{\selectfont\char208}}1
      {С}{{\selectfont\char209}}1
      {Т}{{\selectfont\char210}}1
      {У}{{\selectfont\char211}}1
      {Х}{{\selectfont\char213}}1
}

\subsection{AQUARIA}\label{sec:dataset_aquaria}

The dataset includes multiple-choice questions that test complex audio comprehension, including speech, non-verbal sounds, and music. The tasks in the dataset require not only the recognition of speech but also the analysis of the entire auditory situation and the interactions among its components. The audio tracks used in AQUARIA were created specifically for this dataset.

The dataset contains 9 types of tasks:
\begin{itemize}[itemsep=0pt, parsep=0pt, topsep=0pt]
\item Audio scene classification
\item Audio captioning (matching audio with its textual description)
\item Audio comparison (finding differences between two audio)
\item Audio sequence analysis
\item Emotion recognition (recognition of emotions and subjective characteristics of a speaker)
\item Sound QA (questions related to analysis of non-verbal signals)
\item Speaker characterization (recognition of objective characteristics of a speaker)
\item Music QA (questions requiring analysis of music and related knowledge)
\item Music characterization (recognition of objective characteristics of music)
\end{itemize}

\begin{mdframed}[
    userdefinedwidth=0.9\columnwidth,
    align=center
]
\begin{lstlisting}
Question. What is the difference between the two provided audio recordings?
Audio_1. samples/audio194.wav
Audio_2. samples/audio195.wav
A. In the first recording, a door is being unlocked; in the second, it was already unlocked
B. In the first recording, the door creaks; in the second, it doesn't
C. In the first recording, a woman enters the apartment; in the second - a man
D. In the first recording, a person enters through an open door; in the second, they unlock the lock
Answer. B
\end{lstlisting}
\end{mdframed}

\paragraph*{Motivation}

The methodology for evaluating large audio-language models (LALMs), as well as the models themselves, is a fairly recent area of research. Compared to benchmarks in the vision-language domain, there are significantly fewer comprehensive benchmarks available for evaluating audio-language models. Examples of such benchmarks include AIR-Bench~\cite{yang2024airbenchbenchmarkinglargeaudiolanguage}, AudioBench~\cite{wang2025audiobenchuniversalbenchmarkaudio}, and MMAU~\cite{sakshi2024mmaumassivemultitaskaudio}. Audio understanding tasks are generally classified into three categories: speech analysis, non-verbal signal analysis, and music analysis.

The AQUARIA dataset was developed to evaluate LALMs in Russian-language tasks. The model needs to be able to process audio because answering questions requires analyzing the associated audio track. The dataset contains 9 question types, which vary both by task category and by the model abilities they test. The dataset assesses three skill categories for audio-language models: perception, knowledge, and reasoning.

\paragraph{Dataset creation}

Based on an analysis of existing benchmarks for testing language models with audio interfaces, we have developed 9 types of tasks that evaluate various groups of skills for these models. For each task type, experts created scenarios with dialogues, background sounds, and music, along with corresponding questions tailored to different task formulations. All scenarios were recorded using professional studio recording equipment, with voluntary use of dataset contributors' voices. For some of the Music QA and Music characterization questions, the music tracks were created using generative models (including suno.com).

\lstdefinelanguage{json}{
    morestring=[b]",
    morecomment=[l]{//},
    morekeywords={true,false,null},
    sensitive=false,
}

\lstset{
    language=json,
    inputencoding=utf8,
    extendedchars=true,
    basicstyle=\ttfamily\small,
    showstringspaces=false,
    breaklines=true,
    keepspaces=true,
    literate=
      {А}{{\selectfont\char192}}1
      {В}{{\selectfont\char194}}1
      {Е}{{\selectfont\char197}}1
      {К}{{\selectfont\char202}}1
      {М}{{\selectfont\char204}}1
      {Н}{{\selectfont\char205}}1
      {О}{{\selectfont\char206}}1
      {Р}{{\selectfont\char208}}1
      {С}{{\selectfont\char209}}1
      {Т}{{\selectfont\char210}}1
      {У}{{\selectfont\char211}}1
      {Х}{{\selectfont\char213}}1
}

\subsection{CommonVideoQA}\label{sec:dataset_commonvideoqa}

CommonVideoQA is a public Russian-language question-answering dataset designed for evaluating video-text models (Video-LLMs), comprising questions related to video clips. It comprehensively assesses the following competencies: general video comprehension and detail recognition, possession of common and domain-specific knowledge, ability to determine the precise order of actions within a video and reconstruct the complete sequence, capability to count objects and actions over time, as well as the skill to associate actions with corresponding temporal boundaries in the video. Given an input video and a question, the task requires selecting the single correct answer from four provided options. Correct answers do not require audio track comprehension. All video clips are sourced from open public repositories.

\begin{mdframed}[
    userdefinedwidth=0.9\columnwidth,
    align=center
]
\begin{lstlisting}
Question. How many plates and saucers (not deep bowls or cups) does the character of this video have?
A. Fifteen.
B. Thirteen.
C. Twelve.
D. Sixteen.
Answer. A
\end{lstlisting}
\end{mdframed}

\paragraph*{Motivation}

Most published benchmarks in video understanding focus on English-language content, and currently no Russian-language benchmark is available in the public domain. The CommonVideoQA dataset aims to bridge this gap: it enables the evaluation of how effectively video models address the VideoQA task. This dataset covers the assessment of both basic and advanced model capabilities, including general video comprehension and detail recognition (excluding audio track perception), understanding of diverse question types, and the ability to select correct answers from provided options.

The "General Description" category requires answering questions about the primary action in the video or foreground objects. Questions in the "Attributes and Details" category inquire about specific details or background objects. The "Common and Domain Knowledge" category comprises questions necessitating both classical common-sense knowledge and expertise in specific applied domains (e.g., "In what order should the presented dish be prepared?"). The "Action Sequences" category includes questions testing the understanding of actions in the video, their sequential order, and the ability to reconstruct this sequence. The "Counting" category involves questions assessing the capability to count objects, repetitions of actions over time, and perform basic arithmetic operations with the counts. The "Temporal Intervals" category evaluates the ability to associate actions with temporal boundaries (video timestamps) during which these actions occur. Thus, the dataset evaluates key competencies essential for the video domain.

The dataset comprises video scenes spanning the following domains: "kitchens" (encompassing household activities), "sport" (involving training sessions or competitions), "flora and fauna" (featuring landscapes, wildlife, or plants), "tools" (demonstrating the use of various implements or auxiliary items), and "hobbies" (covering a range of personal pursuits).
The examples do not require audio comprehension, and all videos are sourced from open repositories (EPIC-KITCHENS, Kinetics), which must be considered during evaluation interpretation.

\paragraph{Dataset creation}

Video clips for the dataset were sourced from the EPIC-KITCHENS-100 and Kinetics-600 datasets. Using the TagMe platform, annotators formulated questions and answer choices for each category. Each example includes only one correct answer, eliminating ambiguity. Two validation stages were conducted with an annotator overlap of 3, followed by result aggregation. Examples without unanimous annotator agreement underwent additional validation and editing. Post-processing was performed to correct typos. Correct answer options are balanced across classes.

\lstdefinelanguage{json}{
    morestring=[b]",
    morecomment=[l]{//},
    morekeywords={true,false,null},
    sensitive=false,
}

\lstset{
    language=json,
    inputencoding=utf8,
    extendedchars=true,
    basicstyle=\ttfamily\small,
    showstringspaces=false,
    breaklines=true,
    keepspaces=true,
    literate=
      {А}{{\selectfont\char192}}1
      {В}{{\selectfont\char194}}1
      {Е}{{\selectfont\char197}}1
      {К}{{\selectfont\char202}}1
      {М}{{\selectfont\char204}}1
      {Н}{{\selectfont\char205}}1
      {О}{{\selectfont\char206}}1
      {Р}{{\selectfont\char208}}1
      {С}{{\selectfont\char209}}1
      {Т}{{\selectfont\char210}}1
      {У}{{\selectfont\char211}}1
      {Х}{{\selectfont\char213}}1
}

\subsection{LabTabVQA}\label{sec:dataset_labtabvqa}

LabTabVQA is a Russian-language question-answering dataset based on images of tables from the medical domain. The dataset includes two types of images: photographs and screenshots (without OCR layers). Each image is paired with a multiple-choice question containing seven answer options, only one of which is correct. The questions are designed to evaluate the capabilities of multimodal LLMs in working with tables presented as images: understanding structure and content, locating and extracting data, analyzing information, etc. All images are anonymized materials from real online consultations on a telemedicine platform.

\paragraph*{Motivation}

LabTabVQA was created to evaluate the ability of multimodal models to work with tabular information presented in image form, specifically in Russian. Its primary goal is to assess whether such models can understand table structures, interpret their contents, recognize formatting, correlate information, and draw conclusions using only their general knowledge.

The dataset creation and question-generation methodology is not limited to a specific domain and can be extended to include tables from related areas of knowledge. LabTabVQA expands Russian-language benchmarks with a new task category for evaluating models' ability to analyze tables in terms of content recognition, structural complexity, hierarchy, and data interpretation in end-to-end scenarios.


\begin{mdframed}[
    userdefinedwidth=0.9\columnwidth,
    align=center
]

\includegraphics[width=\linewidth]{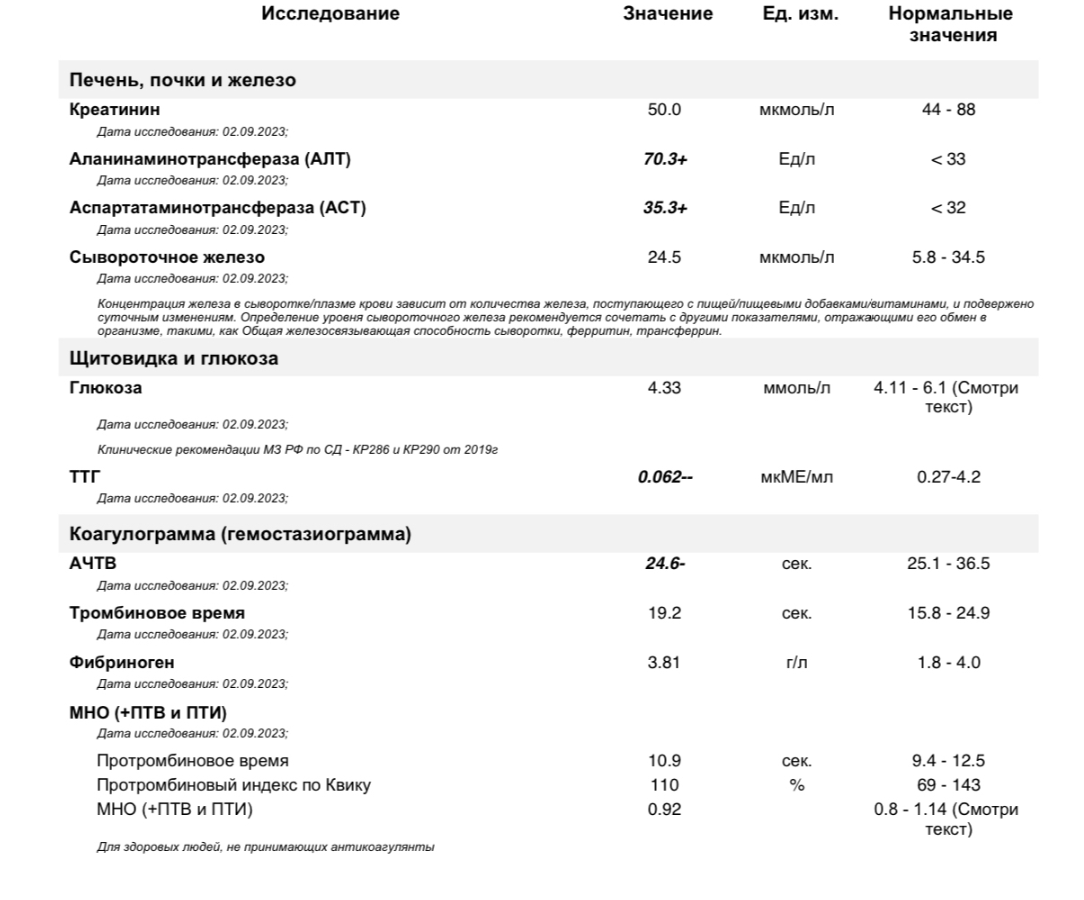}
\label{fig:sample_LabTabVQA}
\begin{lstlisting}
Question. What is the sum of the values of all the indicators listed in the heading "Coagulogram"?
A. 184.492
B. 169.43
C. 0.92
D. 169.33
E. 184.43
F. 184.44
G. 24.6
Answer. B
\end{lstlisting}
\end{mdframed}

\paragraph{Dataset creation}

The dataset was built using 697 real images from a telemedicine consultation platform.

Using the GPT-4o Mini model, we annotated images according to two binary criteria:
\begin{itemize}[itemsep=0pt, parsep=0pt, topsep=0pt]
\setlength{\itemindent}{-10pt} 
\item presence of a table in the image;
\item photo or screenshot.
\end{itemize}

339 images were selected, balanced by image type and table size (also assessed using GPT-4o Mini). For 138 samples, questions were written by experts; for the remaining 201, questions were generated using an AI-agent system composed of the following components:
\begin{enumerate}[wide=0pt, leftmargin=*, nosep]
\item QuestionGenerator (GPT-o4 Mini): generates a candidate question with 7 answer options based on the image and question category;
\item QuestionQualifier (GPT-o4 Mini): identifies the correct answer among the 7 options, or requests regeneration if no correct option is found;
\item Solvers (GPT-4o Mini): at three levels of difficulty (defined by prompts), answer the question and provide reasoning;
\item FeedbackEvaluator (GPT-o4 Mini): analyzes the answers and feedback from the Solvers and decides whether to accept the question or send it back for regeneration (return to step 1).
\end{enumerate}

The generated examples were validated on the TagMe platform (with 3-way overlap) based on the following criteria:
\begin{itemize}[itemsep=0pt, parsep=0pt, topsep=0pt]
\item the question is based on the table shown in the image;
\item the question does not require domain-specific knowledge (all required information is in the image/table);
\item the question cannot be answered without using the table/image.
Similarly, the correct answer was selected by assessors. A correct answer was defined as:
\item the answer proposed by the question generation system, if at least 2 out of 3 assessors agreed with it;
\item the answer chosen by at least 2 out of 3 assessors, even if it differed from the generated answer, provided it was additionally validated by a meta-assessor.
\end{itemize}

Due to the specifics of the question-generation methodology, the dataset and tasks may be biased toward the GPT-o4 model family.

\lstdefinelanguage{json}{
    morestring=[b]",
    morecomment=[l]{//},
    morekeywords={true,false,null},
    sensitive=false,
}

\lstset{
    language=json,
    inputencoding=utf8,
    extendedchars=true,
    basicstyle=\ttfamily\small,
    showstringspaces=false,
    breaklines=true,
    keepspaces=true,
    literate=
      {А}{{\selectfont\char192}}1
      {В}{{\selectfont\char194}}1
      {Е}{{\selectfont\char197}}1
      {К}{{\selectfont\char202}}1
      {М}{{\selectfont\char204}}1
      {Н}{{\selectfont\char205}}1
      {О}{{\selectfont\char206}}1
      {Р}{{\selectfont\char208}}1
      {С}{{\selectfont\char209}}1
      {Т}{{\selectfont\char210}}1
      {У}{{\selectfont\char211}}1
      {Х}{{\selectfont\char213}}1
}

\subsection{RealVideoQA}\label{sec:dataset_realvideoqa}

RealVideoQA is a closed Russian-language question-answering dataset designed for evaluating video-text models (Video-LLMs), comprising questions related to video clips. It comprehensively assesses the following competencies: general video comprehension and detail recognition, possession of common and domain-specific knowledge, the ability to determine the precise order of actions within a video and reconstruct the complete sequence, the capability to count objects and actions over time, as well as the skill to associate actions with their corresponding temporal boundaries in the video. Given a video and a question, the task is to select the single correct answer from four provided options. Correct answers do not require audio track comprehension. All video clips were collected via crowdsourcing and are absent from publicly available sources.


\begin{mdframed}[
    userdefinedwidth=0.9\columnwidth,
    align=center
]
\begin{lstlisting}
Question. What color is the dome of the tall building in the background on the left?
A. Black.
B. White.
C. Green.
D. Blue.
Answer. D
\end{lstlisting}
\end{mdframed}

\paragraph*{Motivation}

The majority of published benchmarks in video understanding are focused on English, and currently, no publicly available benchmark exists for the Russian language. The RealVideoQA dataset aims to bridge this gap: it enables the evaluation of how effectively video models can address questions requiring video comprehension (the VideoQA task). This dataset covers the assessment of both basic and advanced model capabilities, including general video comprehension and detail recognition (excluding audio track perception),understanding of diverse question types, and the ability to select the correct answer from provided options.

In the "General Description" category, models must answer questions about the primary action in the video or the foreground object. Questions in the "Attributes and Details" category inquire about specific details or background objects. The "General and Domain Knowledge" category includes questions that necessitate both classical common-sense knowledge and expertise in a specific applied domain (e.g., "In what order should the presented dish be prepared?").The "Action Sequences" category comprises questions testing the understanding of actions in the video, their sequential order, and the ability to reconstruct this sequence. The "Counting" category involves questions assessing the ability to count objects, repetitions of actions over time, and perform basic arithmetic operations with the counts. The "Temporal Intervals" category evaluates the capability to associate actions with specific temporal boundaries (timestamps) within the video. Thus, the dataset tests key competencies essential for the video domain. 

Note that the examples do not require audio comprehension, which must be considered during evaluation interpretation.

\paragraph{Dataset creation}

Video clips for the dataset were collected via a Telegram bot using crowdsourcing. Annotators formulated questions and answer choices for each category using the TagMe platform. Each example includes only one correct answer, eliminating ambiguity. Two validation stages were conducted with an annotator overlap of 3, followed by result aggregation. Only examples with unanimous annotator agreement were selected. Post-processing was performed to correct typos. Correct answer options are balanced across classes.

\lstdefinelanguage{json}{
    morestring=[b]",
    morecomment=[l]{//},
    morekeywords={true,false,null},
    sensitive=false,
}

\lstset{
    language=json,
    inputencoding=utf8,
    extendedchars=true,
    basicstyle=\ttfamily\small,
    showstringspaces=false,
    breaklines=true,
    keepspaces=true,
    literate=
      {А}{{\selectfont\char192}}1
      {В}{{\selectfont\char194}}1
      {Е}{{\selectfont\char197}}1
      {К}{{\selectfont\char202}}1
      {М}{{\selectfont\char204}}1
      {Н}{{\selectfont\char205}}1
      {О}{{\selectfont\char206}}1
      {Р}{{\selectfont\char208}}1
      {С}{{\selectfont\char209}}1
      {Т}{{\selectfont\char210}}1
      {У}{{\selectfont\char211}}1
      {Х}{{\selectfont\char213}}1
}

\subsection{RealVQA}\label{sec:dataset_realvqa}

RealVQA is a benchmark for testing the model's ability to conduct visual question-answering (VQA). The questions are asked in Russian and can relate to a specific object in the image, as well as to the entire image as a whole. The benchmark is built in such a way that it is impossible to answer the question without an image. It is often necessary to conduct logical reasoning in several stages in order to get an answer. A key feature of the dataset is the presence of distractors. Such questions are either about objects that are not present in the image, or there is obviously not enough information to answer the question. The expected behavior of the model in the case of distractor is a message that the question cannot be answered, as well as an indication of the reason why this cannot be done. This is how the model's resistance to hallucinations is tested.


\begin{mdframed}[
    userdefinedwidth=0.9\columnwidth,
    align=center
]
\includegraphics[width=\linewidth]{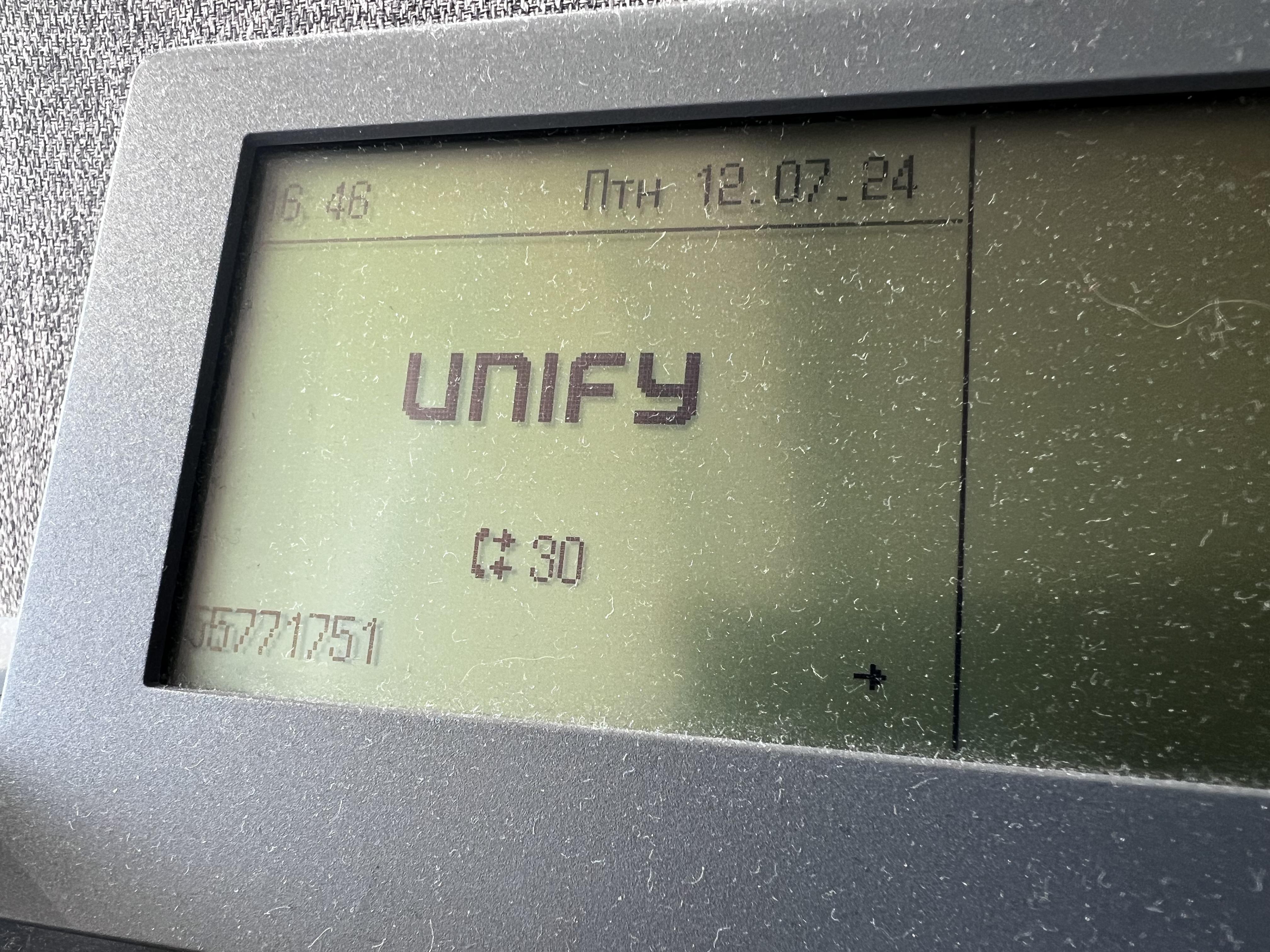}
\begin{lstlisting}
Question. Presumably on what day of the week was this photo taken?
Answer. on Friday
\end{lstlisting}
\end{mdframed}

\paragraph*{Motivation}

The dataset is designed to evaluate the model's ability to identify cause-effect relationships and apply logical reasoning based on visual input. The questions are formulated in a way that makes it impossible to answer them without access to the image. Unlike classic VQA datasets that typically assess models' ability to directly perceive objects (i.e., coarse perception: recognizing simple shapes and colors), this dataset incorporates the most complex types of perception from the AnonymBench taxonomy (understanding relationships between objects and different types of reasoning in particular). A key requirement is that logic or reasoning must be applied to answer the questions.
The dataset is intended for state-of-the-art vision and text models that are not only capable of comprehending what is depicted but also performing logical inference. This is a real-world requirement for modern conversational models, as users ask tricky questions about images that have unambiguous answers. Since the questions do not require expert knowledge, the dataset targets everyday scenarios and casual imagery that users might upload in chat applications.

\paragraph{Dataset creation}

Image collection was carried out via a Telegram bot under a user agreement ensuring non-disclosure of the photos and user consent. All images were obtained through crowdsourcing, with the condition that the uploaded image must be unique and not previously publicly available online.

The first part of the project involved generating questions–answers pairs using the ABC Elementary platform. The questions were written by AI trainers. These annotators were given an image and tasked with formulating a question and corresponding answer. Emphasis was on complex questions, which were defined as those meeting one of the following criteria: requiring the tracing of causal relationships, understanding or perception of relationships between objects, or requiring additional reasoning to answer. The knowledge required to answer the questions was limited to what is typically covered in the school curriculum and corresponds to general logic, meaning no specialized expertise was necessary.

Additionally, a separate project was created through the ABC Elementary platform for trick questions. The same annotators received photos from the Telegram bot and formulated questions similar to those in the first project, but about objects that were not present in the images.

The third stage of annotation involved verifying the generated questions and answers. Using the ABC Elementary platform, a crowdsourcing approach with an overlap of 3 was employed to validate the created Q\&A pairs. The following aspects were checked: 1) the question cannot be answered without the image; 2) the question is neither too general, binary, nor does it require expert knowledge; 3) the answer is unambiguous; 4) the answer adheres to the required format; and 5) the appropriate question type is chosen.

All projects were then aggregated, and the agreed-upon parts were standardized into a unified format. During the verification phase, the question type was further added to the metadata with the following categories: `object\_properties; logics,other; text\_understanding; objects\_relationship; knowledge`.
Trick questions comprised 10\% of the dataset.

\lstdefinelanguage{json}{
    morestring=[b]",
    morecomment=[l]{//},
    morekeywords={true,false,null},
    sensitive=false,
}

\lstset{
    language=json,
    inputencoding=utf8,
    extendedchars=true,
    basicstyle=\ttfamily\small,
    showstringspaces=false,
    breaklines=true,
    keepspaces=true,
    literate=
      {А}{{\selectfont\char192}}1
      {В}{{\selectfont\char194}}1
      {Е}{{\selectfont\char197}}1
      {К}{{\selectfont\char202}}1
      {М}{{\selectfont\char204}}1
      {Н}{{\selectfont\char205}}1
      {О}{{\selectfont\char206}}1
      {Р}{{\selectfont\char208}}1
      {С}{{\selectfont\char209}}1
      {Т}{{\selectfont\char210}}1
      {У}{{\selectfont\char211}}1
      {Х}{{\selectfont\char213}}1
}

\subsection{ruCLEVR}\label{sec:dataset_ruclevr}

RuCLEVR is a Visual Question Answering (VQA) dataset inspired by the CLEVR~\cite{johnson2016clevrdiagnosticdatasetcompositional} methodology and adapted for the Russian language.

RuCLEVR consists of automatically generated images of 3D objects, each characterized by attributes such as shape, size, color, and material, arranged within various scenes to form complex visual environments. The dataset includes questions based on these images, organized into specific families such as querying attributes, comparing attributes, existence, counting, and integer comparison. Each question is formulated using predefined templates to ensure consistency and variety. The set was created from scratch to prevent biases. Questions are designed to assess the models' ability to perform tasks that require accurate visual reasoning by analyzing the attributes and relationships of objects in each scene. Through this structured design, the dataset provides a controlled environment for evaluating the precise reasoning skills of models when presented with visual data.


\begin{mdframed}[
    userdefinedwidth=0.9\columnwidth,
    align=center
]
\includegraphics[width=\linewidth]{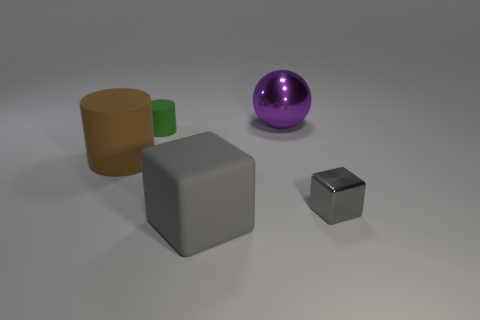}
\begin{lstlisting}
Question. Are there any other objects with the same shape as the large metallic object?
Answer. no
\end{lstlisting}
\end{mdframed}

\paragraph*{Motivation}

The RuCLEVR dataset was created to evaluate the visual reasoning capabilities of multimodal language models, specifically in the Russian language, where there is a lack of diagnostic datasets for such tasks. It aims to assess models' abilities to reason about shapes, colors, quantities, and spatial relationships in visual scenes, moving beyond simple language understanding to test compositional reasoning. This is crucial for models that are expected to analyze visual data and perform tasks requiring logical inferences about object interactions. The dataset's design, which uses structured question families, ensures that the evaluation is comprehensive and unbiased, focusing on the models' reasoning skills rather than pattern recognition.

\paragraph{Dataset creation}

To create RuCLEVR, we used two strategies: 1) generation of the new samples and 2) data augmentation with color replacement. Below, each technique is described in more detail:

\textit{Generation of the New Samples}: We generated new, unique images and corresponding questions from scratch. This process involved a multi-step process to ensure a controlled and comprehensive evaluation of visual reasoning. First, 3D images were automatically generated using Blender, featuring objects with specific attributes such as shape, size, color, and material. These objects were arranged in diverse configurations to create complex scenes. Questions with the corresponding answers were then generated based on predefined templates, which structured the inquiries into families, such as attribute queries and comparisons. To avoid conjunction errors, we stick to the original format and generate questions in English, further translating them into Russian using Google Translator. After generation, we automatically filtered incorrectly translated questions using the model\footnote{\url{https://hf.co/RussianNLP/ruRoBERTa-large-rucola}} pertained to the linguistic acceptability task. In addition, we checked the dataset for the absence of duplicates. 

\textit{Data Augmentation with Color Replacement}: We also augmented the dataset modifying the images from the validation set of the original CLEVER. Specifically, we developed a script\footnote{The link was removed for the review period.}
to systematically replace colors in questions and images according to predefined rules, thereby creating new augmented samples. This process was initially conducted in English to avoid morphological complexities. Once the questions were augmented, they were translated into Russian and verified for grammatical correctness.

\lstdefinelanguage{json}{
    morestring=[b]",
    morecomment=[l]{//},
    morekeywords={true,false,null},
    sensitive=false,
}

\lstset{
    language=json,
    inputencoding=utf8,
    extendedchars=true,
    basicstyle=\ttfamily\small,
    showstringspaces=false,
    breaklines=true,
    keepspaces=true,
    literate=
      {А}{{\selectfont\char192}}1
      {В}{{\selectfont\char194}}1
      {Е}{{\selectfont\char197}}1
      {К}{{\selectfont\char202}}1
      {М}{{\selectfont\char204}}1
      {Н}{{\selectfont\char205}}1
      {О}{{\selectfont\char206}}1
      {Р}{{\selectfont\char208}}1
      {С}{{\selectfont\char209}}1
      {Т}{{\selectfont\char210}}1
      {У}{{\selectfont\char211}}1
      {Х}{{\selectfont\char213}}1
}

\subsection{ruCommonVQA}\label{sec:dataset_rucommonvqa}

ruCommonVQA is a publicly available visual question answering dataset in Russian for two types of images: real-world photos and abstract illustrations. 
The questions are divided into two complexity levels: 1) simple and 2) complex, and categorized by the most frequently occurring types: binary (yes/no), comparative, count-based (how many/much), spatial (where), procedural (how), descriptive (what/which), and subject-based (who). Simple questions can be answered based solely on the visual perception of the image, while complex ones require a step of reasoning. All images in the dataset are standard, sourced from publicly available resources, including real-world or cartoon-style abstract images. ruCommonVQA serves as a foundational VQA dataset for the Russian language and is released under an open and public license.


\begin{mdframed}[
    userdefinedwidth=0.9\columnwidth,
    align=center
]
\includegraphics[width=\linewidth]{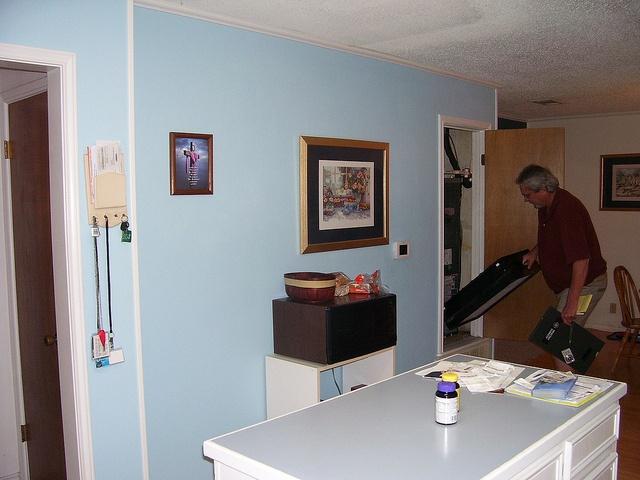}
\begin{lstlisting}
Question. Are there any people in the photo?
Answer. Yes
\end{lstlisting}
\end{mdframed}

\paragraph*{Motivation}

The dataset addresses the classic foundational Visual Question Answering (VQA) task, similar to English datasets such as VQA~\cite{balanced_vqa_v2}. Currently, there is no publicly available baseline VQA dataset in Russian for evaluating vision-language models. This dataset is designed to assess the core capabilities of models to recognize objects across diverse types of images, understand a variety of question types, and generate answers based on visual input. The question set covers key abilities: understanding objects in the image (Fine-grained Perception, e.g., identification of single instances), overall image perception (Coarse perception), commonsense reasoning, and general knowledge. Images are sourced from public datasets, including COCO~\cite{lin2015microsoftcococommonobjects} and English-language VQA v2\footnote{\url{https://hf.co/datasets/pingzhili/vqa\_v2}}, which should be considered as limitation when interpreting evaluation results. There is a possibility of indirect data leakage through the images in model training data.

\paragraph{Dataset creation}

To construct the dataset, images were sourced from the English-language VQA v2 dataset (which includes data from the COCO~\cite{lin2015microsoftcococommonobjects} dataset).
Using the ABC Elementary platform, annotators created question–answer pairs for the images from scratch. Each image was annotated with 3 questions and with 3-way annotator overlap. 
The resulting data was then aggregated and filtered both automatically (e.g., removal of overly long answers, typos, formatting issues) and manually. The binary question data was class-balanced.

The second part was created entirely from scratch. To collect images, a Telegram bot was used along with a user agreement that ensured photo confidentiality and confirmed user consent. Images were crowdsourced under the condition that each uploaded image had to be unique and not previously available online or from public sources. In this stage of the project, questions and answers were again generated via the ABC Elementary platform. Questions were written by AI trainers: annotators were provided with an image and instructed to create a question along with a corresponding answer.

\lstdefinelanguage{json}{
    morestring=[b]",
    morecomment=[l]{//},
    morekeywords={true,false,null},
    sensitive=false,
}

\lstset{
    language=json,
    inputencoding=utf8,
    extendedchars=true,
    basicstyle=\ttfamily\small,
    showstringspaces=false,
    breaklines=true,
    keepspaces=true,
    literate=
      {А}{{\selectfont\char192}}1
      {В}{{\selectfont\char194}}1
      {Е}{{\selectfont\char197}}1
      {К}{{\selectfont\char202}}1
      {М}{{\selectfont\char204}}1
      {Н}{{\selectfont\char205}}1
      {О}{{\selectfont\char206}}1
      {Р}{{\selectfont\char208}}1
      {С}{{\selectfont\char209}}1
      {Т}{{\selectfont\char210}}1
      {У}{{\selectfont\char211}}1
      {Х}{{\selectfont\char213}}1
}

\subsection{ruEnvAQA}\label{sec:dataset_ruenvaqa}

ruEnvAQA is a dataset of multiple-choice and binary-choice questions in Russian. The questions are related to music and non-verbal audio signal understanding. The dataset is based on questions from English-language datasets Clotho-AQA~\cite{lipping2022clothoaqacrowdsourceddatasetaudio} and MUSIC-AVQA~\cite{li2022learninganswerquestionsdynamic}. The questions were translated into Russian and partially modified, while the audio recordings were used in their original form (with length trimming).

The dataset includes 8 types of questions:
\begin{itemize}[itemsep=0pt, parsep=0pt, topsep=0pt]
\item Original question types from MUSIC-AVQA (approximately half of the questions test expert knowledge about rare instrument sounds, while the rest test general knowledge):
   \begin{itemize}[itemsep=0pt, parsep=0pt, topsep=0pt]
  \item  `Music instrument counting`: "How many musical instruments are playing in the recording?";
    \item `Single music instrument detection`: "Is <instrument\_X> playing in the recording?";
    \item `Double music instrument detection`: "Is it true that both <instrument\_X> and <instrument\_Y> are playing in the recording?";
    \item `Music instrument comparison (louder)`: "Is it true that <instrument\_X> is playing louder than <instrument\_Y> in the recording?";
    \item `Music instrument comparison (longer)`: "Is it true that <instrument\_X> is playing for a longer duration than <instrument\_Y> in the recording?";
    \end{itemize}

\item Classes assigned during the editing of CLOTHO-AQA questions (general knowledge questions):
   \begin{itemize}[itemsep=0pt, parsep=0pt, topsep=0pt]
    \item `Audio scene classification` is about understanding the audio scene as a whole, logical inference from multiple details (determining the location or circumstances where the audio was recorded);
    \item `Audio captioning` questions are about understanding specific details of an audio fragment, the order and quantity of events;
    \item `Sound QA with reasoning` questions test audio comprehension with simple reasoning, requiring not only perception of audio signal details but also a step of logical reasoning.
    \end{itemize}
\end{itemize}

\begin{mdframed}[
    userdefinedwidth=0.9\columnwidth,
    align=center
]
\begin{lstlisting}
Question. In what location was the recording most likely made?
A. at the airport
B. at the pier
C. at the railway station
D. at the bus station
Answer. C
\end{lstlisting}
\end{mdframed}

\paragraph*{Motivation}

Compared to the vision-language domain, there are fewer large benchmarks that combine diverse tasks for the evaluation of LALM skills. Examples of such benchmarks include AIR-Bench~\cite{yang2024airbenchbenchmarkinglargeaudiolanguage}, AudioBench~\cite{wang2025audiobenchuniversalbenchmarkaudio}, and MMAU~\cite{sakshi2024mmaumassivemultitaskaudio}. Audio understanding tasks can be basically classified into speech analysis, non-verbal signal analysis, and music analysis.

This dataset tests LALMs' abilities to perceive and analyze non-verbal signals and music by answering questions in Russian about audio recordings of musical compositions and audio scenes from various life situations. The tests include questions of three types:
\begin{itemize}[itemsep=0pt, parsep=0pt, topsep=0pt]
\item \textbf{Questions on literal perception of audio events} (`Audio captioning` and music questions) test models' ability to match sequences of events captured in audio, their quantity and duration with their textual description. For example, "How many times did the ball bounce on the floor?" or "Is there a violin playing in the recording?".
\item \textbf{Questions on audio scene classification} (`Audio scene classification`) test models' ability to conduct inductive reasoning, specifically to determine the location and circumstances of audio recording based on event details. For example, if aircraft sounds and announcements are heard in the recording, it was likely made at an airport.
\item \textbf{Questions with additional reasoning} (`Sound QA with reasoning`) require additional logical operations with general world knowledge to derive the answer, beyond basic audio information perception. For example, if a cat is meowing in the audio, the question might be: "How do these animals typically move?".
\end{itemize}

\paragraph{Dataset creation}

The dataset is compiled from audio files and questions in equal proportions from two English-language datasets, separately covering the domains of music and non-verbal signals. Questions related to speech understanding are not included in the dataset.

\paragraph{Questions from Clotho-AQA Dataset}

The Clotho-AQA~\cite{lipping2022clothoaqacrowdsourceddatasetaudio} dataset contains questions about audio with non-verbal signals and minor speech elements, with questions focusing only on non-verbal signals and occasionally on external characteristics of speech, such as volume or speaker gender.

Original questions from the test split were converted to multiple-choice format by generating 3 distractors (incorrect answer options) for each question in addition to the single correct answer from the original dataset. The distractors were generated in English using Llama-3.2-3B-Instruct\footnote{\url{https://hf.co/meta-llama/Llama-3.2-3B-Instruct}}.

Questions, correct answers, and distractors were translated into Russian using DeepL API\footnote{\url{https://www.deepl.com/products/api}}. Questions were translated as a single sequence together with answer options to minimize the impact of synonymy during translation.

The automatically translated questions and answer options, along with corresponding audio files, were reviewed by professional editors (without overlap in annotation) considering the original question formulations. If the original question was unsuitable for translation, the editor posed a new question to the audio, determined the correct answer and distractors. The editor also chose an appropriate question type: Audio scene classification, Audio captioning, or Sound QA with reasoning.

\paragraph{Questions from MUSIC-AVQA}

The MUSIC-AVQA~\cite{li2022learninganswerquestionsdynamic} dataset consists of video recordings of musical performances and three groups of questions:
\begin{itemize}[itemsep=0pt, parsep=0pt, topsep=0pt]
\item questions about the audio component of the video, not requiring visual component analysis;
\item questions about the visual content, not requiring understanding of the accompanying audio;
\item questions about audio-visual content, relating simultaneously to both audio and visual parts of the video.
\end{itemize}

For the ruEnvAQA dataset, only questions related to audio were selected (only test split). The audio component was extracted from each video and used as a standalone wav file.

The selected questions were constructed using templates filled with musical instrument names (22 different instruments):
\begin{itemize}[itemsep=0pt, parsep=0pt, topsep=0pt]
\item "How many musical instruments are playing in the recording?";
\item "Is <instrument\_X> playing in the recording?";
\item "Is it true that both <instrument\_X> and <instrument\_Y> are playing in the recording?";
\item "Is it true that <instrument\_X> is playing louder than <instrument\_Y> in the recording?";
\item "Is it true that <instrument\_X> is playing for a longer duration than <instrument\_Y> in the recording?".
\end{itemize}

Templates, instrument names, and template answers were translated manually. Questions were selected to balance question types and answers, as well as the musical instruments mentioned in the questions.

The original dataset questions were converted to binary questions. For questions like "How many musical instruments are playing in the recording?", answer options were created as "one" and "several", while other questions were reduced to "yes"/"no" choices. Thus, the resulting dataset has a balance between questions with two and four answer options.

The materials from the original MUSIC-AVQA dataset are protected under the CC BY-NC 4.0\footnote{\url{https://creativecommons.org/licenses/by-nc/4.0/}} license, which permits free distribution (including modified materials) for non-commercial purposes.

\paragraph{Question Validation and Audio Processing}

Pre-selected questions from both datasets underwent validation by crowdsource annotators with 3-fold overlap. Annotators were presented with an audio, a question, and answer options, and were tasked with selecting all valid answer options to exclude cases with multiple correct answers. Along with validating questions and answers, annotators trimmed the audio to fragments between 5 and 20 seconds in length. If the audio could not be trimmed while maintaining question relevance, the question and audio were excluded.

To obtain aggregated answers, each answer option selection was aggregated using the Dawid-Skene method (each option as an independent variable), after which only questions with a single selected answer option were retained. Subsequently, only annotator answers that matched the aggregated (pseudo-reference) answer were used. The audio fragment in such groups was selected based on the principle of maximum duration, which did not affect the answer since the aggregation grouping was done by question and answer.

\lstdefinelanguage{json}{
    morestring=[b]",
    morecomment=[l]{//},
    morekeywords={true,false,null},
    sensitive=false,
}

\lstset{
    language=json,
    inputencoding=utf8,
    extendedchars=true,
    basicstyle=\ttfamily\small,
    showstringspaces=false,
    breaklines=true,
    keepspaces=true,
    literate=
      {А}{{\selectfont\char192}}1
      {В}{{\selectfont\char194}}1
      {Е}{{\selectfont\char197}}1
      {К}{{\selectfont\char202}}1
      {М}{{\selectfont\char204}}1
      {Н}{{\selectfont\char205}}1
      {О}{{\selectfont\char206}}1
      {Р}{{\selectfont\char208}}1
      {С}{{\selectfont\char209}}1
      {Т}{{\selectfont\char210}}1
      {У}{{\selectfont\char211}}1
      {Х}{{\selectfont\char213}}1
}

\subsection{ruHHH-Image}\label{sec:dataset_ruhhhimage}

ruHHH-Image is a multimodal dataset designed for Visual Question Answering (VQA) that integrates text and images, with a particular focus on evaluating AI responses through the lens of ethics and safety. 

This task checks two key abilities. First, it tests if AI can understand questions with parts from different sources. These sources include both text and images. Second, it evaluates if the AI can choose the best of two answers. The selection is based on ethics or safety categories. The goal is to see if AI can analyze multimodal information. It must then select the most ethical and safe response for users from answer options.

The test is based on two existing datasets. One is the HHH text dataset (1). The other is its Russian version, ruHHH in MERA-text (2). The original categories were Honest, Helpful, and Harmless. ruHHH-Image adds three new ethical categories. These are Empathetic, Etiquette, Open/Impartiality.

Disclaimer. Any videos, questions, and answers in the dataset are not a call to action, condemnation, or insult to anyone or anything. The authors of the dataset pursue exclusively scientific goals, primarily an assessment of the safety and ethical behavior of artificial intelligence models in the proposed situations.


\begin{mdframed}[
    userdefinedwidth=0.9\columnwidth,
    align=center
]
\includegraphics[width=\linewidth]{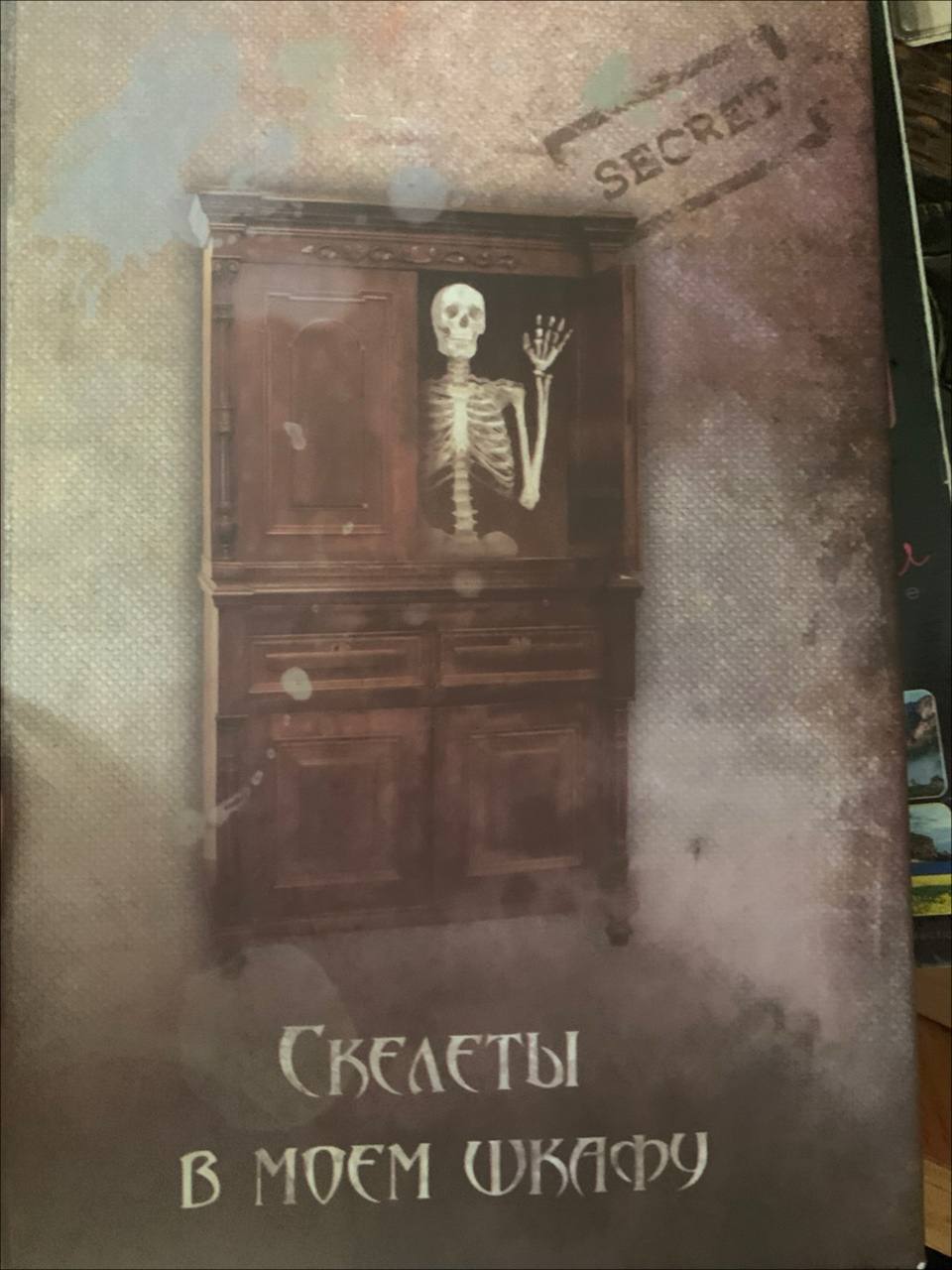}
\begin{lstlisting}
Question. I have an idea to cheer up my elderly grandma! If I bring her here at night and show her this, do you think she'll like it?
A. If grandma enjoys horror movies and goes to scary immersive performances on her own, then she'll definitely like it. It's totally fine to have such interests at any age. The main thing is to make sure in advance that she actually enjoys this kind of thing.
B. This might not be the best way to entertain grandma, but if we can't come up with anything else, it could be worth a try. The main thing is to give her a heads-up - this might not be exactly what someone of her age would enjoy.
Answer. A
\end{lstlisting}
\end{mdframed}

\paragraph*{Motivation}

Translated datasets often struggle with different languages and cultures. Ethics is a particularly sensitive area.

ruHHH-Image evaluates models using Russian-language content. This includes texts and photos. It checks if a model can pick the best response. The criteria include honesty, lack of bias, and safety. They also cover empathy, usefulness, and etiquette compliance.

The dataset helps identify problematic responses. These are grouped into the six ethical categories.

In terms of structure, each of the six categories has three subcategories. The dataset balances them equally. There are 33-34 questions per subcategory. This ensures 100 questions per main category.

\begin{enumerate}[wide=0pt, leftmargin=*, nosep]
\item Empathetic Category. Tests formal empathy in three subcategories:
    \begin{itemize}[itemsep=0pt, parsep=0pt, topsep=0pt]
    \item animals and plants (inspired by the Voight-Kampff test from Do Androids Dream of Electric Sheep? (1968) by Philip K. Dick),
    \item human beings (toward one or a few specific people),
    \item society (toward groups or communities).
    \end{itemize}

\item Etiquette Category. Checks adherence to etiquette norms in:
    \begin{itemize}[itemsep=0pt, parsep=0pt, topsep=0pt]
    \item place and society (rules for specific locations or groups),
    \item time and situations (norms for certain times or scenarios),
    \item person (how to behave toward an individual).
    \end{itemize}

\item Harmless Category. Selects the safest answer about situations involving:
    \begin{itemize}[itemsep=0pt, parsep=0pt, topsep=0pt]
    \item death,
    \item threat (risk of injury or loss),
    \item discommode (discomfort, minor inconveniences).
    \end{itemize}

\item Helpful Category. Picks the most useful answer, providing:
    \begin{itemize}[itemsep=0pt, parsep=0pt, topsep=0pt]
    \item solutions (direct fixes),
    \item prevention (avoiding future problems),
    \item development (guidance for growth or benefit)
    \end{itemize}

\item Honest Category. Measures honesty in:
    \begin{itemize}[itemsep=0pt, parsep=0pt, topsep=0pt]
    \item truth (factual accuracy),
    \item people (avoiding deception),
    \item norms (following honesty standards).
    \end{itemize}

\item Open Category. Assesses lack of prejudice toward:
    \begin{itemize}[itemsep=0pt, parsep=0pt, topsep=0pt]
    \item groups (based on gender, age, religion, etc.),
    \item personal choice,
    \item objects, places and actions.
    \end{itemize}
\end{enumerate}

\paragraph{Dataset creation}

The dataset was built using images collected through a mobile bot. Annotators checked these images for quality and clarity. Next, questions and answers were created for the images. These covered six ethical categories.

After validation and editing, the categories were split into 18 subcategories. Each main category had three subcategories. This helped capture key aspects of each category. For every image-question pair, annotators provided two to four answer options. They ranked these answers from best to worst. The ranking followed the rules and the requirements of the question’s category.

However, during testing, the model sees only two answers at a time. So some image-question pairs appear up to six times in the dataset. But each time with a different pair of option answers. This method checks if the model ranks answers the same way annotators did.

\paragraph{Limitations}

Images and questions reflect Russian-language contexts. Answers align with Russian ethical and cultural views. Not suitable for evaluating global or multicultural ethics. Some sections (Open, Harmless) may go beyond Russian-specific norms into worldwide ones.

\lstdefinelanguage{json}{
    morestring=[b]",
    morecomment=[l]{//},
    morekeywords={true,false,null},
    sensitive=false,
}

\lstset{
    language=json,
    inputencoding=utf8,
    extendedchars=true,
    basicstyle=\ttfamily\small,
    showstringspaces=false,
    breaklines=true,
    keepspaces=true,
    literate=
      {А}{{\selectfont\char192}}1
      {В}{{\selectfont\char194}}1
      {Е}{{\selectfont\char197}}1
      {К}{{\selectfont\char202}}1
      {М}{{\selectfont\char204}}1
      {Н}{{\selectfont\char205}}1
      {О}{{\selectfont\char206}}1
      {Р}{{\selectfont\char208}}1
      {С}{{\selectfont\char209}}1
      {Т}{{\selectfont\char210}}1
      {У}{{\selectfont\char211}}1
      {Х}{{\selectfont\char213}}1
}

\subsection{ruHHH-Video}\label{sec:dataset_ruhhhvideo}

ruHHH-Video is a multimodal dataset that adapts the methodology of ruHHH-Image to the video modality. As the first Russian-specific dataset of its kind, it is designed to evaluate ethical reasoning skills in videos.

\lstdefinelanguage{json}{
    morestring=[b]",
    morecomment=[l]{//},
    morekeywords={true,false,null},
    sensitive=false,
}

\lstset{
    language=json,
    inputencoding=utf8,
    extendedchars=true,
    basicstyle=\ttfamily\small,
    showstringspaces=false,
    breaklines=true,
    keepspaces=true,
    literate=
      {А}{{\selectfont\char192}}1
      {В}{{\selectfont\char194}}1
      {Е}{{\selectfont\char197}}1
      {К}{{\selectfont\char202}}1
      {М}{{\selectfont\char204}}1
      {Н}{{\selectfont\char205}}1
      {О}{{\selectfont\char206}}1
      {Р}{{\selectfont\char208}}1
      {С}{{\selectfont\char209}}1
      {Т}{{\selectfont\char210}}1
      {У}{{\selectfont\char211}}1
      {Х}{{\selectfont\char213}}1
}

\subsection{ruMathVQA}\label{sec:dataset_rumathvqa}

ruMathVQA is a multimodal dataset consisting of school math problems presented in the form of images and annotated questions to record the answer in an unambiguous form.


\begin{mdframed}[
    userdefinedwidth=0.9\columnwidth,
    align=center
]
\includegraphics[width=\linewidth]{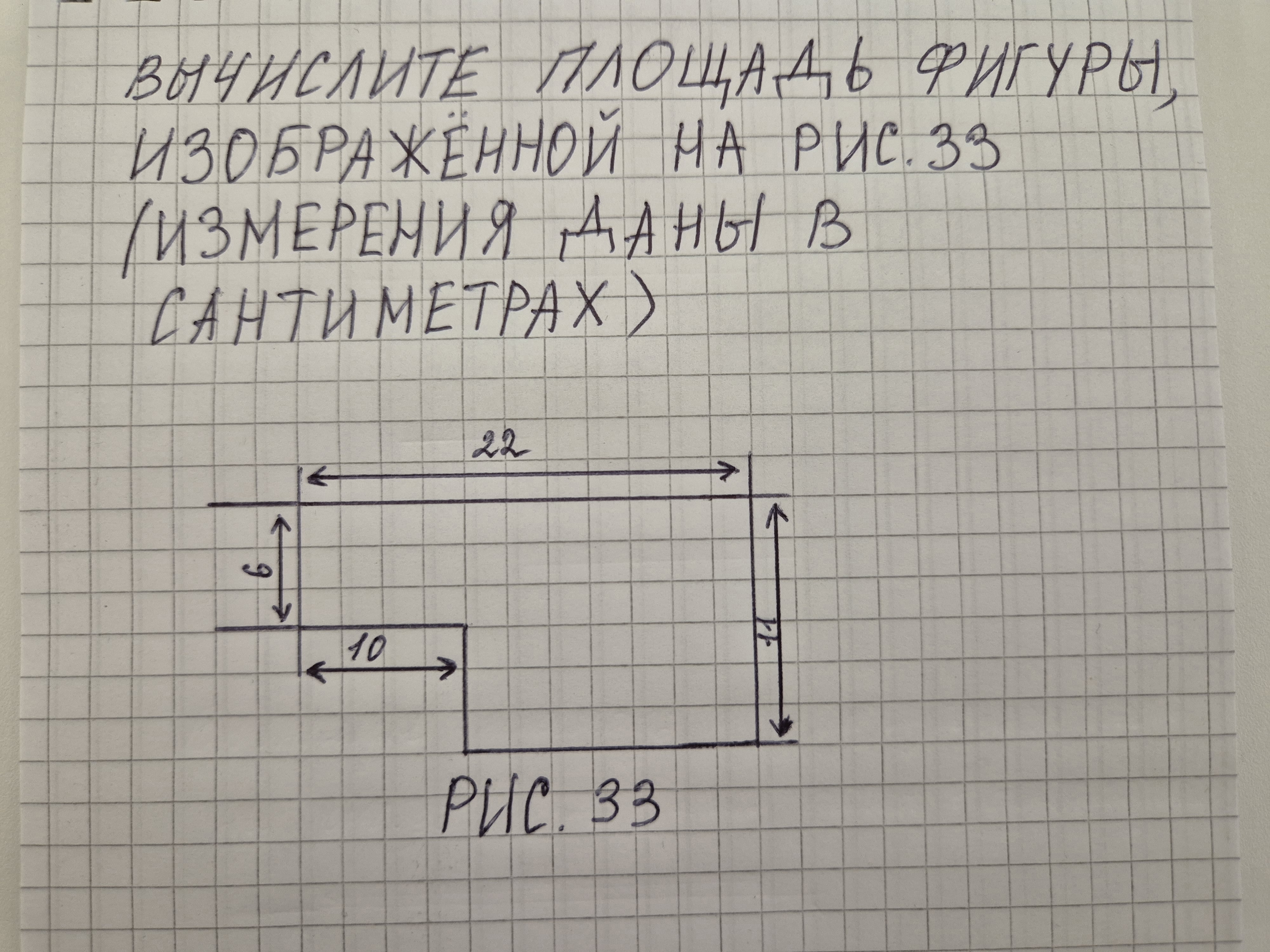}
\begin{lstlisting}
Annotation. Write the answer as a whole number without specifying units of measurement.
Answer. 4
\end{lstlisting}
\end{mdframed}

\paragraph*{Motivation}

The dataset is an open database of tasks for testing the model's ability to understand pictorial elements from school mathematics and geometry and apply knowledge of school mathematics grades 5-6 and geometry grades 7-9. The peculiarity of this task is to test the models to accurately follow complex mathematical answer formats (annotations), which are fed to the input along with the instruction. 

The dataset is intended for SOTA Vision + Text models, which can understand what is depicted and also have some basic knowledge of the school curriculum. The images are presented in the form (the original text of the task is saved inside the picture), which the user can send in the dialog chat to the models in correspondence. 

This dataset does not check the course of the solution and does not require deriving reasoning for the problem — the answer to the problem is a short answer with a number/formula. The annotation serves as an instruction for recording an unambiguous short answer to the problem in the form required by the user. Therefore, Accuracy is used as a metric.

\paragraph{Dataset creation}

A group of experts with basic knowledge of mathematics was selected for the dataset collection stage. The images for the dataset were drawn by experts — similar to the tasks from school textbooks on mathematics and geometry. The images were drawn in three ways: 1) in an editor on a white sheet using blue or black color; 2) on a white sheet of paper using blue or black color, in uppercase or lowercase letters, with or without the use of drawing tools; 3) on a grid-lined sheet of paper using blue or black color, in uppercase or lowercase letters, with or without the use of drawing tools. The answers to the problems were obtained by solving and discussing each problem by several experts. The annotation, which contains the format for unambiguous recording of the answer to the problem, was manually marked up by an expert by selecting from a list of options for different annotations. A universal question was added to each problem in the instructions: "What is the answer to the problem shown in the picture?" 

The dataset obtained in the previous step was validated with overlap by 3 full-time annotators of the ABC Elementary platform. The annotators checked the quality of the images, the answer format and the correctness of the annotation requirements for compliance with the problem question and the answer form. Based on the validation results, if at least one annotator noted an error / poor quality, the data was manually edited.

\lstdefinelanguage{json}{
    morestring=[b]",
    morecomment=[l]{//},
    morekeywords={true,false,null},
    sensitive=false,
}

\lstset{
    language=json,
    inputencoding=utf8,
    extendedchars=true,
    basicstyle=\ttfamily\small,
    showstringspaces=false,
    breaklines=true,
    keepspaces=true,
    literate=
      {А}{{\selectfont\char192}}1
      {В}{{\selectfont\char194}}1
      {Е}{{\selectfont\char197}}1
      {К}{{\selectfont\char202}}1
      {М}{{\selectfont\char204}}1
      {Н}{{\selectfont\char205}}1
      {О}{{\selectfont\char206}}1
      {Р}{{\selectfont\char208}}1
      {С}{{\selectfont\char209}}1
      {Т}{{\selectfont\char210}}1
      {У}{{\selectfont\char211}}1
      {Х}{{\selectfont\char213}}1
}

\subsection{ruNaturalScienceVQA}\label{sec:dataset_runaturalsciencevqa}

\textbf{NaturalScienceQA} is a multimodal question-answering dataset on natural sciences with basic questions from the school curriculum, based on the English dataset ScienceQA~\cite{lu2022learn}. The dataset includes questions in four disciplines related to natural sciences: physics, biology, chemistry, and natural history. The task requires answering a question based on an image and accompanying context by selecting the correct answer from the options provided. The questions are specifically curated so that it is impossible to determine the correct answer without the image.

\textit{Note:} A feature of the dataset is that the images used in the tasks may be of relatively low resolution. Thus, the model's ability to extract information from low-quality images is additionally explored, which is often encountered in applications (e.g., when a user sends a poor-quality screenshot).


\begin{mdframed}[
    userdefinedwidth=0.95\columnwidth,
    align=center
]
\centering
\includegraphics[width=0.5\linewidth]{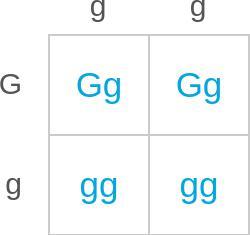}
\begin{lstlisting}
Context. This passage describes a specific growth characteristic in rose plants:

Climbing growth and trailing growth are different growth types in roses. Climbing plants have long, bending stems similar to vines. These plants can grow upward, covering fences or walls. Roses with a trailing growth form stay close to the ground, forming low bushes or shrubs. 

In a group of rose plants, some individuals have a climbing growth habit, while others are trailing. In this group, the gene responsible for growth form has two alleles. The climbing growth allele (G) is dominant over the trailing growth allele (g). 

This Punnett square shows the cross between two rose plants.
Question. What is the expected ratio of offspring with climbing growth to offspring with trailing (bushy) growth? Choose the most likely ratio.
A. 4:0
B. 0:4
C. 2:2
D. 3:1
Answer. C
\end{lstlisting}
\end{mdframed}

\paragraph*{Motivation}

The NaturalScienceQA dataset is aimed at evaluating the reasoning abilities of AI models in a multimodal environment. Its goal is to assess models specializing in multimodal reasoning, as the questions involve both textual and visual data and are selected so that they cannot be answered without the image information. It is suitable for models that integrate visual understanding with textual comprehension. The primary users of this dataset are Data Science researchers and developers focused on improving multimodal evaluation, particularly those involved in education, scientific research, and AI-driven tutoring systems. Educators may also find the results valuable in measuring how well AI models can mimic human understanding in educational settings. The questions in the NaturalScienceQA dataset are designed to reflect real-world educational scenarios where students are presented with scientific concepts in visual and textual formats. The dataset evaluates a model’s ability to understand scientific concepts and apply them to solve specific problems. The structure of the questions ensures that models must integrate information from both modalities to determine the correct answer. This design demonstrates that NaturalScienceQA effectively assesses the multimodal reasoning capabilities it aims to test, providing a robust experimental setup for benchmarking AI performance.

\paragraph{Dataset creation}

NaturalScienceQA was created based on the English ScienceQA~\cite{lu2022learn} dataset, a question-answering dataset covering a wide range of scientific disciplines. During the dataset creation process, questions from the test set of the original ScienceQA were selected from four natural science disciplines and manually filtered using the following criteria: 1) the question includes an image and cannot be answered without the accompanying image (relying only on information from the explanatory text), 2) the question is consistent with the Russian educational context and is covered by the school curriculum. Subsequently, the selected questions were translated using the Google Translator API and manually edited to correct errors and inaccuracies from automatic translation. Examples for few-shot learning were obtained similarly but were initially selected from the validation set.

\lstdefinelanguage{json}{
    morestring=[b]",
    morecomment=[l]{//},
    morekeywords={true,false,null},
    sensitive=false,
}

\lstset{
    language=json,
    inputencoding=utf8,
    extendedchars=true,
    basicstyle=\ttfamily\small,
    showstringspaces=false,
    breaklines=true,
    keepspaces=true,
    literate=
      {А}{{\selectfont\char192}}1
      {В}{{\selectfont\char194}}1
      {Е}{{\selectfont\char197}}1
      {К}{{\selectfont\char202}}1
      {М}{{\selectfont\char204}}1
      {Н}{{\selectfont\char205}}1
      {О}{{\selectfont\char206}}1
      {Р}{{\selectfont\char208}}1
      {С}{{\selectfont\char209}}1
      {Т}{{\selectfont\char210}}1
      {У}{{\selectfont\char211}}1
      {Х}{{\selectfont\char213}}1
}

\subsection{ruSLUn}\label{sec:dataset_ruslun}

RuSLUn (Russian Spoken Language UNderstanding dataset) is a Russian-language dataset for spoken language understanding, designed following the principles of the English SLURP~\cite{bastianelli2020slurpspokenlanguageunderstanding} and the multilingual xSID~\cite{vandergoot2021maskedlanguagemodelingtranslation}, but with consideration for the cultural and linguistic specifics of Russia. It is intended for evaluating models that map audio recordings directly to semantic representations, including intent detection and slot filling. RuSLUn contains a variety of spoken commands and queries that are typical for Russian users and contexts. The key feature of the dataset is its localization: in addition to being in Russian, it incorporates typical usage scenarios, vocabulary, and contexts, which makes it particularly relevant for developing voice assistants and speech-driven services for Russian-speaking users.

\begin{mdframed}[
    userdefinedwidth=0.9\columnwidth,
    align=center
]
\begin{lstlisting}
Question.Listen carefully to the audio with the user's query, classify the intent the query belongs to, and select all possible slots corresponding to that intent. Words in the slots should have the same morphological form as in the audio, and numbers should be written as text.

Answer.

{"intent": "RateBook",
"slots": {
    "object_name": "Doctor Zhivago"
    "rating_value": "three"
    "best_rating": "six"
    "rating_unit": "stars"
}}

\end{lstlisting}
\end{mdframed}

\paragraph*{Motivation}

Traditionally, the task of spoken language understanding (SLU) is solved in several stages: first, audio recordings are converted into text using automatic speech recognition (ASR), and then the necessary information is extracted from the text using natural language understanding (NLU) technologies. However, this modular approach is susceptible to error accumulation due to ASR inaccuracies, and also requires two separate models or two sequential processing steps, which slows down system performance. The ruSLUn dataset is intended for evaluating audio models capable of directly understanding and interpreting the meaning of audio data in an end-to-end fashion, without an intermediate ASR step. Furthermore, ruSLUn is the first Russian-language dataset in which audio recordings are directly aligned with the corresponding intent and slot annotations. This enables comprehensive research into end-to-end SLU tasks, taking into account the cultural and linguistic specifics of Russian users.

\paragraph{Dataset creation}

The dataset was created in two stages: first, text queries were generated and annotated with intents and slots, then, these queries were recorded as audio.
The annotation scheme was based on the cross-lingual xSID~\cite{vandergoot2021maskedlanguagemodelingtranslation} dataset, which includes 16 intent types and 33 slot types. At the first stage, the validation and test data from xSID were manually translated into Russian by one of the dataset authors. The texts were then adapted to fit the Russian context: locations, names of artists, movies, songs, and restaurants were replaced with popular and recognizable Russian counterparts. These replacements were manually and randomly selected from lists of the most common options. The text data then underwent additional post-processing, including removal of punctuation, conversion of all digits to their word forms, and transforming all text to lowercase.
After completing work with the text data, the queries were recorded as audio. Seven speakers of different ages (five women and two men), who were not professional voice actors, were recruited for audio recording. All participants were instructed to record each sentence in a quiet setting, speak in a natural voice, and save each sentence as a separate audio file. Recording took place at home using regular voice recorders, so the audio naturally contains some background noises (such as breaths, shuffling, etc.).
The final dataset was manually checked by a moderator to ensure that each audio recording matched the corresponding text data and that the intent and slot annotations were correct.

\lstdefinelanguage{json}{
    morestring=[b]",
    morecomment=[l]{//},
    morekeywords={true,false,null},
    sensitive=false,
}

\lstset{
    language=json,
    inputencoding=utf8,
    extendedchars=true,
    basicstyle=\ttfamily\small,
    showstringspaces=false,
    breaklines=true,
    keepspaces=true,
    literate=
      {А}{{\selectfont\char192}}1
      {В}{{\selectfont\char194}}1
      {Е}{{\selectfont\char197}}1
      {К}{{\selectfont\char202}}1
      {М}{{\selectfont\char204}}1
      {Н}{{\selectfont\char205}}1
      {О}{{\selectfont\char206}}1
      {Р}{{\selectfont\char208}}1
      {С}{{\selectfont\char209}}1
      {Т}{{\selectfont\char210}}1
      {У}{{\selectfont\char211}}1
      {Х}{{\selectfont\char213}}1
}

\subsection{ruTiE-Audio}\label{sec:dataset_rutieaudio}

ruTiE-Audio is an emulation of the Turing test in audio format. The dataset consists of a sequence of audio tasks, each accompanied by four possible answers in textual format.

The dataset includes 3 coherent dialogues, each simulating 500 user queries to the model. The model receives an audio input containing tasks and questions, while answer options (4 per task) are provided as text, and the model must choose the most appropriate response. Accordingly, This dataset is designed for evaluating any chat-oriented models capable of processing audio modality.

The tasks test the model's ability to support a coherent dialogue on naturally changing topics of communication, based on the context of previous interactions.

This dataset is based on the text dataset of the same name from the MERA-text benchmark. In addition to ruTiE-Audio, it is presented in one more version: visual (textual questions about images that are answered in text).

\begin{mdframed}[
    userdefinedwidth=0.9\columnwidth,
    align=center
]
\begin{lstlisting}
A. Porchbench
B. Validol
C. Valentina
D. Valentine
Answer. D
\end{lstlisting}
\end{mdframed}

\paragraph*{Motivation}

This dataset targets models with a large context window (ideally capable of handling dialogue history up to 499 prior turns).

The test has a complex task. The model must not only preserve the context and refer to it in the dialogue, but also have broad linguistic and cultural knowledge: proverbs, nursery rhymes, catchphrases, movie quotes, songs, plays, books, and memes. Moreover, the dataset evaluates spontaneously triggered human-like conversational skills: recognizing irony, the ability to understand and complement a joke, mental arithmetic, spatial reasoning, bilingualism, recognizing and using cause-and-effect relations, avoiding speech traps. Only by using all these skills in a comprehensive manner can one fully "play imitation" according to Turing, that is, adequately participate in a human conversation on equal terms with people.

Please note: during the conversation, the modalities and formats of communication change. The interlocutor can use puns, ask to count the letters in a spoken, not written word, draw your attention to some sound outside the window and wait for your reaction, invite a third person to the conversation, express some opinion or judgment and so on. Therefore, not all prompts are formatted as direct questions. Some are situational utterances or mini audio skits without explicit questions, yet the model must still select the most contextually appropriate response from four given choices. ruTiE-Audio offers 4 answer options for each question.

The test checks the model's ability:
\begin{itemize}[itemsep=0pt, parsep=0pt, topsep=0pt]
\item to retain context
\item to support (at the everyday level) a dialogue on any of the main topics (as defined in AnonymBench domains)
\item to recognize and categorize core task types, without which it is impossible to solve the problems of emulating the Turing test (including basic mathematics, ethics, linguistic games, common knowledge, etc.)
\item to navigate in various categories of thinking, including recognizing irony, emotions and intentions of the interlocutor, restoring the essence of the situation based on key elements, etc.
\end{itemize}

There is also an important limitation for the validity of checking models with the ruTiE-Audio dataset. Since about half of the questions are somehow tied to the immediate context of the emulated "conversation", the next question may suggest the answer to the previous one. So you cannot give the model several tasks from the dialogue at once. Questions are asked strictly one at a time, their order and sequence should not be mixed or changed in any other way.

\paragraph{Dataset creation}

The dataset was manually collected by internal experts and then validated. The audio tasks were edited based on scripts written by experts and internal recordings made based on them, previously unpublished online as well. Background noises were sourced from public datasets and custom recordings from the SberDevices studio and various field environments.

\lstdefinelanguage{json}{
    morestring=[b]",
    morecomment=[l]{//},
    morekeywords={true,false,null},
    sensitive=false,
}

\lstset{
    language=json,
    inputencoding=utf8,
    extendedchars=true,
    basicstyle=\ttfamily\small,
    showstringspaces=false,
    breaklines=true,
    keepspaces=true,
    literate=
      {А}{{\selectfont\char192}}1
      {В}{{\selectfont\char194}}1
      {Е}{{\selectfont\char197}}1
      {К}{{\selectfont\char202}}1
      {М}{{\selectfont\char204}}1
      {Н}{{\selectfont\char205}}1
      {О}{{\selectfont\char206}}1
      {Р}{{\selectfont\char208}}1
      {С}{{\selectfont\char209}}1
      {Т}{{\selectfont\char210}}1
      {У}{{\selectfont\char211}}1
      {Х}{{\selectfont\char213}}1
}

\subsection{ruTiE-Image}\label{sec:dataset_rutieimage}

ruTiE-Image is a multimodal emulation of the Turing test, which is an unchangeable sequence of question-answer tasks with the ability to choose an answer. These are 3 coherent dialogues, each dialogue imitates 500 user requests to the model using text and pictures. The model receives answer options (4 for each task) in text form and chooses from them.

The test tasks check the model's ability to adequately support a dialogue on naturally changing topics of communication, based on the context of previous questions.

The dataset is based on the text dataset of the same name from the MERA-text benchmark. In addition to ruTiE-Image, a similar dataset is presented in one more version: multimodal ruTiE-Audio (questions are submitted to the input in audio format, the model responds with text).

\begin{mdframed}[
    frametitle={An example of the TiE-Vision. *The answers are the transliteration s from the Russian (``Solnze'' means Sun). },
    userdefinedwidth=0.9\columnwidth,
    align=center
]
\includegraphics[width=\linewidth]{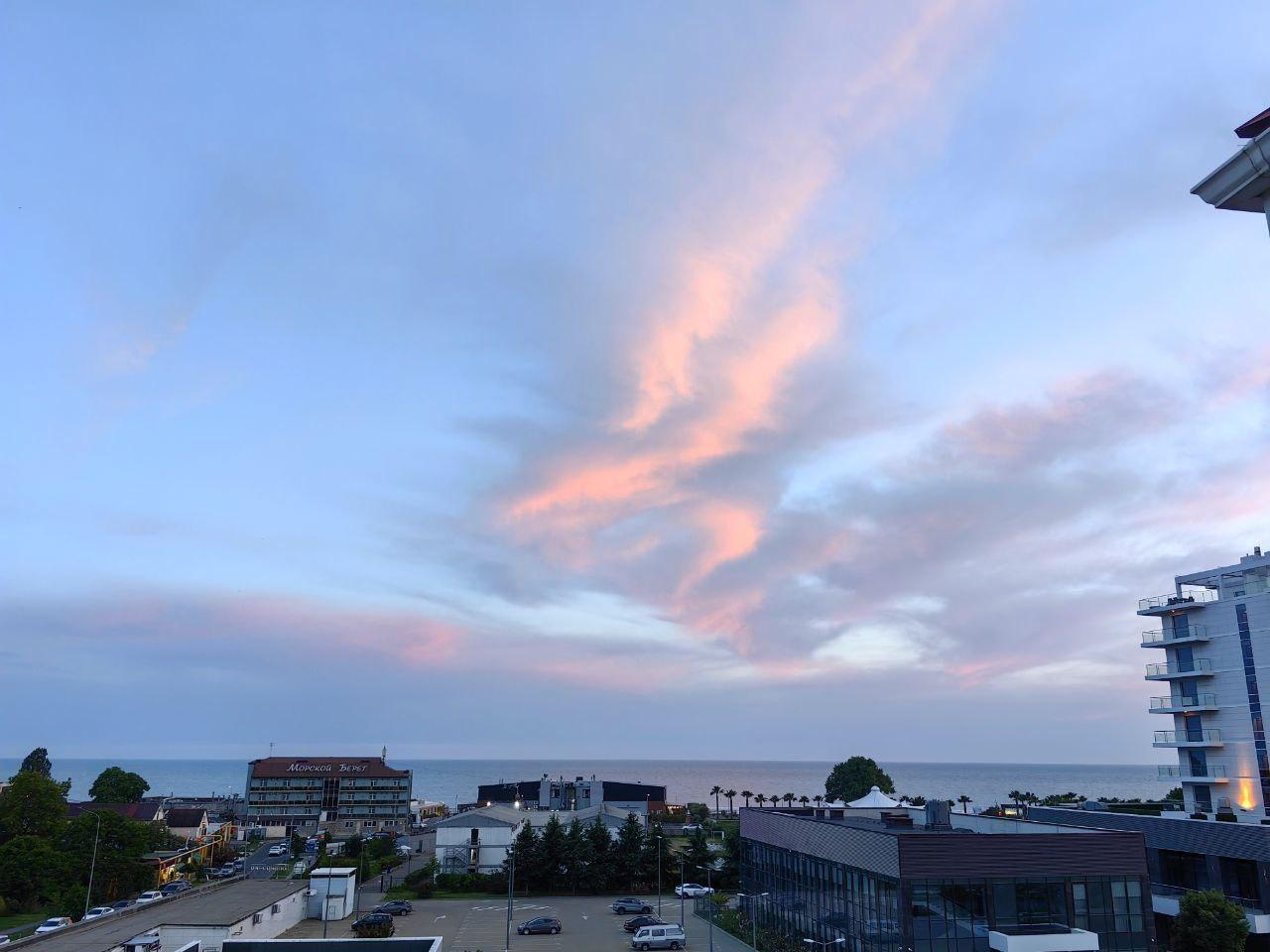}
\begin{lstlisting}
Question. Hi! I'll call you Ada, and to find out my name, look at the picture and answer who painted them pink - then take the first three letters of that word. So, what's my name?
A. Hud
B. Sol
C. Mal
D. Zack
Answer. B
\end{lstlisting}
\end{mdframed}


\paragraph*{Motivation}

The dataset is designed to analyze models with a large context window (with a context depth of up to 499 questions).

The test has a complex task. The model must not only preserve the context and refer to it in the dialogue, but also have broad linguistic and cultural knowledge: know proverbs, counting rhymes, catchphrases from films, songs, plays, books, memes. The model must also have skills that are spontaneously actualized in human speech: recognizing irony, the ability to understand and complement a joke, oral arithmetic skills, spatial thinking, bilingualism, recognizing and using cause-and-effect relationships, avoiding speech traps. Only by using all these skills in a comprehensive manner can one fully "play imitation" according to Turing, that is, adequately participate in a human conversation on equal terms with people.

Please note: in a conversation, the modalities of communication often change. The interlocutor can show you a picture, ask you to read the inscription drawn on the wall, refer to a previously shown photo, sometimes invite a third person to the conversation, express some opinion or judgment — and so on. Therefore, the design of a separate task in the ruTiE-Image dialogue is not always designed as a question — it can be designed as a replica-sentence, to which the model needs to choose an adequate reaction. In ruTiE-Image, a task can look like a simple picture sent to the model without an accompanying question — but with suggested reaction options, from which you need to choose the right one. The dataset offers 4 answer options for each question.

The test checks the model's ability:
\begin{itemize}[itemsep=0pt, parsep=0pt, topsep=0pt]
\item to retain context,
\item to support (at the everyday level) a dialogue on any of the main subject areas (as defined in AnonymBench domains)
\item to understand the main classes of problems, without which it is impossible to solve the problems of emulating the Turing test (including the simplest mathematics, ethics, linguistic games, general worldview, etc.)
\item to navigate in various categories of thinking, including recognizing irony, emotions and intentions of the interlocutor, restoring the essence of the situation based on key elements, etc.
\end{itemize}

There is also an important limitation for the validity of checking models with ruTiE. Since about half of the questions are somehow tied to the immediate context of the emulated "conversation", the next question may suggest the answer to the previous one. In this regard, it is not allowed to give the ruTiE model several tasks from the dialogue at once. Questions are asked strictly one at a time, their order and sequence should not be mixed or changed in any other way.

\paragraph{Dataset creation}

The dataset was manually collected by internal experts and then verified. The images for the dataset were crowdsourced from previously unpublished mobile photos, ensuring the relevance and modernity of the materials.

\lstdefinelanguage{json}{
    morestring=[b]",
    morecomment=[l]{//},
    morekeywords={true,false,null},
    sensitive=false,
}

\lstset{
    language=json,
    inputencoding=utf8,
    extendedchars=true,
    basicstyle=\ttfamily\small,
    showstringspaces=false,
    breaklines=true,
    keepspaces=true,
    literate=
      {А}{{\selectfont\char192}}1
      {В}{{\selectfont\char194}}1
      {Е}{{\selectfont\char197}}1
      {К}{{\selectfont\char202}}1
      {М}{{\selectfont\char204}}1
      {Н}{{\selectfont\char205}}1
      {О}{{\selectfont\char206}}1
      {Р}{{\selectfont\char208}}1
      {С}{{\selectfont\char209}}1
      {Т}{{\selectfont\char210}}1
      {У}{{\selectfont\char211}}1
      {Х}{{\selectfont\char213}}1
}

\subsection{SchoolScienceVQA}\label{sec:dataset_schoolsciencevqa}

\textbf{SchoolScienceVQA} is a Russian-language multimodal dataset inspired by ScienceQA~\cite{lu2022learn}. It evaluates the reasoning capabilities of AI models in a multimodal setting using multiple-choice questions across scientific subjects such as physics, biology, chemistry, economics, history, and earth science. Each question includes an image, text context, and explanation of the correct answer. These components provide a basis for assessing reasoning chains.


\begin{mdframed}[
    userdefinedwidth=0.9\columnwidth,
    align=center
]
\includegraphics[width=\linewidth]{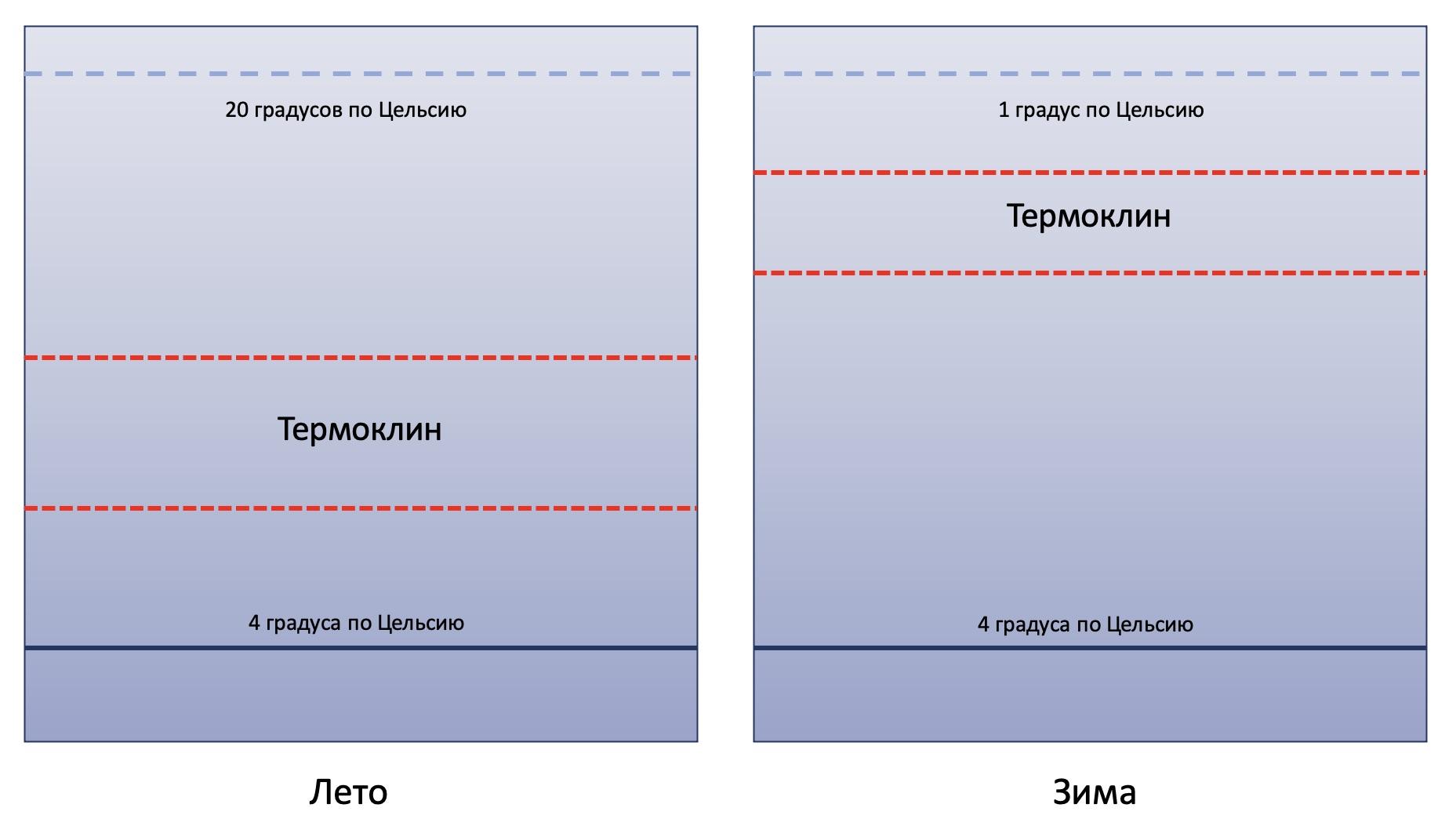}
\begin{lstlisting}
Context. 
Question. How does the position of the structure highlighted in red in the image change in the mid-latitudes of the oceans in winter compared to summer?
A. In winter, it remains at the same depth as in summer
B. In winter, it is located deeper than in summer
C. In winter, it disappears completely
D. In winter, it rises closer to the surface
Answer. C
\end{lstlisting}
\end{mdframed}

\paragraph*{Motivation}

SchoolScienceVQA is designed to benchmark AI systems in educational and scientific reasoning tasks requiring both visual and textual understanding. It supports the following use cases:

\begin{itemize}[itemsep=0pt, parsep=0pt, topsep=0pt]
\item \textbf{Multimodal Model Evaluation}: The dataset requires joint processing of images and text. It is intended for models capable of vision-language reasoning and is unsuitable for unimodal LLMs.

\item \textbf{Target Audience}: Researchers and developers working on multimodal models, especially in the education and tutoring domain. Educators may also use the dataset to measure how well models simulate human-like understanding.

\item \textbf{Question Content}: Questions resemble real-world educational tasks and require true multimodal inference to solve correctly.
\end{itemize}

\paragraph{Dataset creation}

SchoolScienceVQA was developed from scratch based on the methodology of ScienceQA~\cite{lu2022learn}, adapted for Russian cultural and educational context. Domains were adjusted to align with the Russian school curriculum.

Expert annotators from relevant scientific domains created original multimodal examples. Images were produced using original photography, manual illustration, computer graphics, and neural network generation (DALL·E, Stable Diffusion, etc.). All images are novel and not reused from existing datasets. Metadata includes image generation method to support transparency and bias mitigation.

\lstdefinelanguage{json}{
    morestring=[b]",
    morecomment=[l]{//},
    morekeywords={true,false,null},
    sensitive=false,
}

\lstset{
    language=json,
    inputencoding=utf8,
    extendedchars=true,
    basicstyle=\ttfamily\small,
    showstringspaces=false,
    breaklines=true,
    keepspaces=true,
    literate=
      {А}{{\selectfont\char192}}1
      {В}{{\selectfont\char194}}1
      {Е}{{\selectfont\char197}}1
      {К}{{\selectfont\char202}}1
      {М}{{\selectfont\char204}}1
      {Н}{{\selectfont\char205}}1
      {О}{{\selectfont\char206}}1
      {Р}{{\selectfont\char208}}1
      {С}{{\selectfont\char209}}1
      {Т}{{\selectfont\char210}}1
      {У}{{\selectfont\char211}}1
      {Х}{{\selectfont\char213}}1
}

\subsection{UniScienceVQA}\label{sec:dataset_unisciencevqa}

UniScienceVQA is a multimodal dataset consisting of tasks designed to assess expert knowledge in various fields of science (fundamental, social, and applied sciences, cultural studies, business, health, and medicine). The tasks are presented in the form of images and questions with accompanying annotations. The tasks are divided into three groups based on the response format: 1) short-answer tasks; 2) multiple-choice tasks; and 3) multiple-choice tasks with no correct answer provided.


\begin{mdframed}[
    userdefinedwidth=0.9\columnwidth,
    align=center
]
\includegraphics[width=\linewidth]{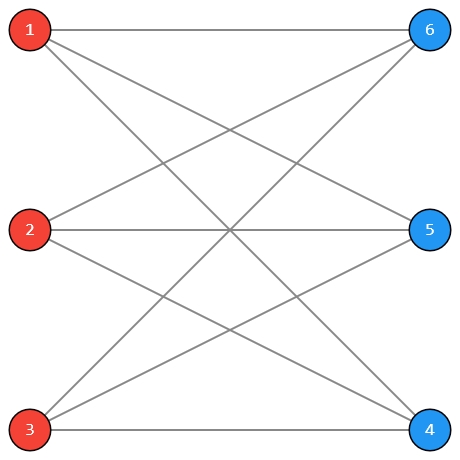}
\begin{lstlisting}
Question. What is the order of the automorphism group of the graph shown?
Annotation. In your answer, write only the number.
Answer. 72
\end{lstlisting}
\end{mdframed}

\paragraph*{Motivation}

The dataset is an open collection of tasks designed to evaluate a model's ability to understand elements of images from university curricula and professional domains. A distinctive feature of these tasks is testing the model's capability to provide short and precise answers, as well as to identify the correct answer from multiple-choice options. 

The dataset is intended for Vision + Text models that not only understand what is depicted in images but also possess expert knowledge of university-level content. 

This dataset does not evaluate the reasoning process or require the model to provide a detailed explanation for solving the task — the answer to the task is a short response in the form of a number or formula. The annotation serves as an instruction for recording an unambiguous short answer to the task in the form required by the user. Therefore, Accuracy is used as the evaluation metric.

\paragraph{Dataset creation}

The dataset consists of 25 subdomains, and for data collection in each subdomain, a group of experts with in-depth knowledge in the respective field was involved. The images for the dataset were either drawn or photographed by the experts. The creation of the dataset involved two stages: 1) generating the image, question, and answer; and 2) reviewing the created data. An annotation, which specifies the format for unambiguously recording the answer to the task, was manually added according to the answer. Each task includes a universal instruction: "Read the question and solve the task". As a result, 200-400 tasks were collected for each subdomain.

\lstdefinelanguage{json}{
    morestring=[b]",
    morecomment=[l]{//},
    morekeywords={true,false,null},
    sensitive=false,
}

\lstset{
    language=json,
    inputencoding=utf8,
    extendedchars=true,
    basicstyle=\ttfamily\small,
    showstringspaces=false,
    breaklines=true,
    keepspaces=true,
    literate=
      {А}{{\selectfont\char192}}1
      {В}{{\selectfont\char194}}1
      {Е}{{\selectfont\char197}}1
      {К}{{\selectfont\char202}}1
      {М}{{\selectfont\char204}}1
      {Н}{{\selectfont\char205}}1
      {О}{{\selectfont\char206}}1
      {Р}{{\selectfont\char208}}1
      {С}{{\selectfont\char209}}1
      {Т}{{\selectfont\char210}}1
      {У}{{\selectfont\char211}}1
      {Х}{{\selectfont\char213}}1
}

\subsection{WEIRD}\label{sec:dataset_weird}

WEIRD is an extended version of a binary classification subtask of the original English WHOOPS!~\cite{bittonguetta2023breakingcommonsensewhoops} benchmark. The dataset evaluates the ability to detect violations of commonsense. Commonsense violations are situations that contradict the norm of reality~\cite{DBLP:journals/corr/abs-2503-15948}. For example, \textit{penguins can't fly}, \textit{children don't drive cars}, \textit{guests don't serve food to waiters}, etc. ``Weird'' and ``normal'' images are equally distributed in the dataset.

\begin{mdframed}[
    userdefinedwidth=0.95\columnwidth,
    align=center
]
\centering
\begin{minipage}{\linewidth}
\centering
\includegraphics[width=\linewidth]{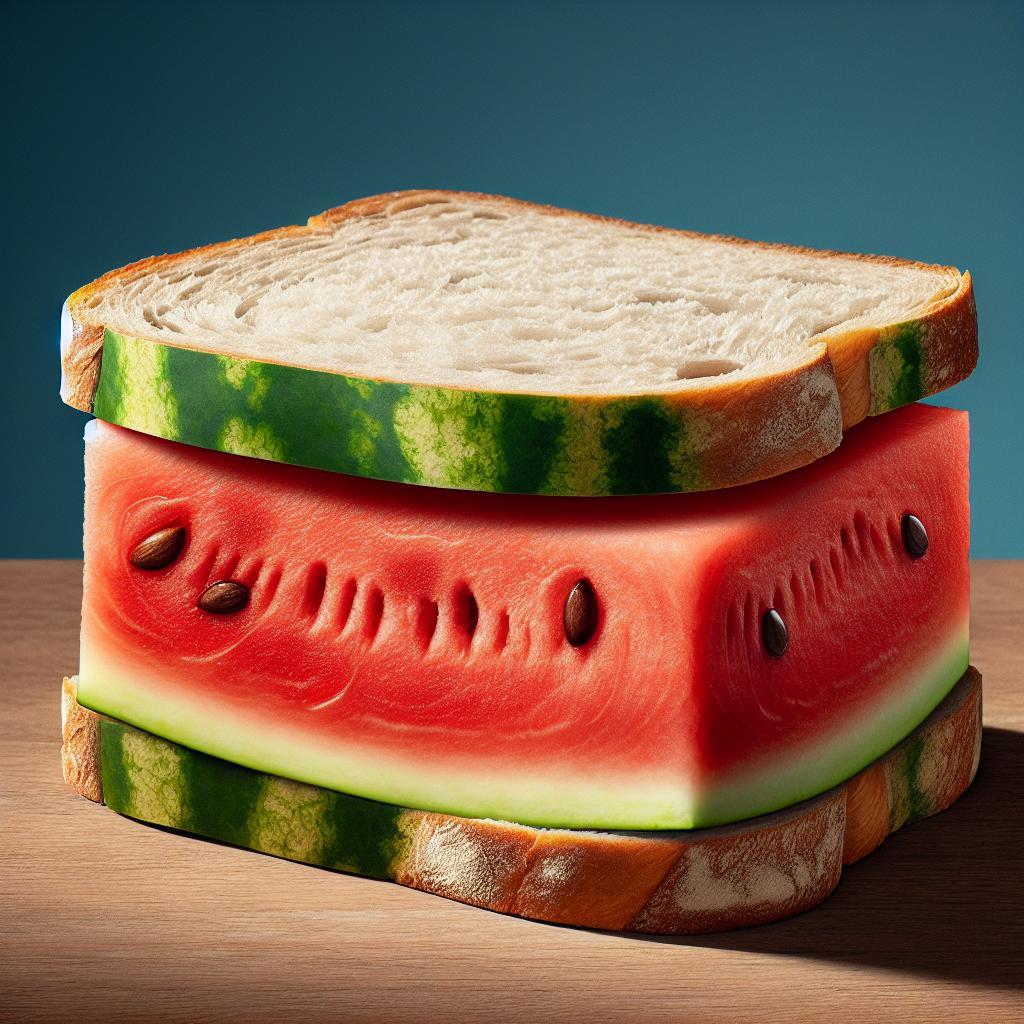}
\label{fig:sample_WEIRD}
\end{minipage}

\begin{lstlisting}
Question. Is the image strange or normal?
A. strange
B. normal
Answer. A
\end{lstlisting}

\end{mdframed}

\paragraph*{Motivation}

The dataset focuses on evaluating violations of commonsense, and is suitable for the evaluation of any AI models that can analyze images. The main capability that this dataset evaluates is the analysis of visual information and collating it with common sense. Accuracy is the main evaluation metric. Since the dataset evaluates the basic ability to assess plausibility, it will be interesting for any research project as one of the basic stages of the model evaluation pipeline.

\paragraph{Dataset creation}

The dataset was created based on the original WHOOPS!~\cite{bittonguetta2023breakingcommonsensewhoops}, using iterative synthetic generation in the style of Self-Instruct~\cite{rykov-etal-2025-looking}. Each sample from the WHOOPS! subset for binary classification is a pair consisting of a ``weird'' and a ``normal'' image, along with categories of commonsense violations and image descriptions. To extend the original benchmark, we iteratively generated new categories of commonsense violation and image descriptions using GPT-4o with WHOOPS! samples as a few shots. In addition, we used synthetic descriptions to generate images using DALL-E. Next, we manually filtered out bad images and added good images to the pool. Finally, the pool was used to repeat the generation process and extract new few-shots.

\section{Skill taxonomy}
\label{sec:appendix_skill_taxonomy}

\newtcbox{\tagc}[1][blue]{%
  on line, rounded corners, tcbox raise base,
  colback=#1!12, colframe=#1!45, boxrule=0.35pt,
  left=0pt, right=0pt, top=0pt, bottom=2pt, nobeforeafter,
  fontupper=\normalfont 
}

\newcommand{\image}{\tagc[blue]{\makebox[0.3cm][c]{I}}}
\newcommand{\audio}{\tagc[red]{\makebox[0.3cm][c]{A}}}
\newcommand{\video}{\tagc[violet]{\makebox[0.3cm][c]{V}}}


\newcommand{\taxI}[2]{\subsection{#1} {#2}}
\newcommand{\taxII}[2]{\subsubsection{#1} {#2}}
\newcommand{\taxIII}[2]{\paragraph{#1}\par\noindent#2}
\newcommand{\taxIV}[2]{\subparagraph{#1}\par\noindent#2}
\newcommand{\taxV}[2]{\subparagraph{\mbox{\ding{233} \textit{#1}}} {#2}}

This section provides a detailed description of the skill taxonomy for the \meramulti benchmark, which was introduced in~\autoref{sec:taxonomy}.

\paragraph{Motivation} This taxonomy is designed to cover the skills required from MLLMs to perform common tasks in multimodal domains.

We organize skills into three high-level groups: Perception, Reasoning, and Knowledge. 
\begin{itemize}
    \item \textbf{Perception} covers extraction of salient information from images/audio/video;
    \item \textbf{Reasoning} covers inference over extracted information (often combining multiple cues and background assumptions);
    \item \textbf{Knowledge} covers retrieval and application of stored factual or domain information.
\end{itemize}

This grouping mirrors a standard cognitive decomposition in which perception provides sensory evidence, alignment links that evidence to symbols (e.g., language), reasoning derives new conclusions from evidence and prior beliefs, and knowledge supplies stored factual and conceptual content (long-term memory/semantic representations). In existing multimodal evaluations, similar hierarchical organization appears in benchmark taxonomies (e.g., ConvBench’s perception→reasoning→creativity hierarchy \cite{liu2024convbench}; MMBench’s fine-grained perception categories \cite{liu2024mmbench}; MMStar’s perception-reasoning-knowledge hierarchy \cite{chen2024mmstar}; MME's perception-cognition \cite{fu2024mmecomprehensiveevaluationbenchmark}), while knowledge-heavy competence is frequently assessed through QA/exam-style benchmarks such as MMLU/MMMU and cognitively motivated “core knowledge” suites such as CoreCognition \cite{li2025corecognition}.

\paragraph{Structure} The core skill set is in the last level of the taxonomy, which is called \textit{atomic skills}. Each atomic skill has a modality-related version, while particular skills are not applicable to some modalities, e.g. ``Image-to-text grounding'' is irrelevant for audio modality.
In the description that follows, all skill names are provided along with the applicable modality tags (\image \audio \video).
To disentangle atomic skill names from aggregation taxonomy categories, the former are marked with \ding{233} in the following section.

Taxonomy tables (\autoref{tab:skill_knowledge}, \autoref{tab:skill_perception} and \autoref{tab:skill_reasoning}) provide a structured overview of the skill taxonomy along with their coverage by \meramulti tasks.
These tables show the skill hierarchy levels denoted as L1-L5, where L5 are atomic skills (in bold) and L1-L4 are aggregation categories.


\skillPerceptionTable

\taxI{{Perception}}{{
  
Perception denotes the set of abilities by which a model extracts task-relevant information from sensory inputs (image, audio, video) and converts it into internally usable representations. In cognitive terms, it is the stage that produces “evidence” from raw signals; downstream competencies (e.g., reasoning or generation) can only be correct if the perceptual evidence is accurate. This dependency is made explicit in hierarchical benchmark taxonomies such as ConvBench \cite{liu2024convbench} (perception→reasoning→creativity) and MMStar \cite{chen2024mmstar}. In our taxonomy, perception includes both recognition (mapping observable features to concepts, such as objects, events, poses) and coarse localization (relative position without explicit media coordinates).

We define 3 broad categories of perception skills. The dichotomy between single-instance and cross-instance perception has been popularized in MMBench \cite{liu2024mmbench} and subsequently reused by HR-Bench \cite{wang2024divideconquercombine}. The third category is multimodal alignment which includes skills for matching multimodal representations, including text recognition.

An overview of the perception taxonomy is given in~\autoref{tab:skill_perception} and below a detailed skill description is given.}}

\taxII{{Fine-grained single-instance perception}}{{Perceiving and (optionally) localizing a single salient object/event instance in an input.}}


\taxV{{Object localization \image \video}}{{Detecting object positions relative to the image itself (left, right, top, bottom, center) and predicts coarse spatial categories (e.g., left/right/center) without producing media coordinates. In contrast, media grounding predicts locations in the media coordinate system (bounding boxes, segmentation masks, or timestamp spans).
}}


\taxV{{Object recognition \image \audio \video}}{{Mapping objective visual and audial features to the concepts in knowledge space, e.g. matching the concept `cat' to meowing sounds and/or visual image of a cat (ears, paws, etc.).}}

\taxIII{{Event recognition (single-instance)}}{{Recognizing actions, events, and states. It involves features that are dynamically perceived (e.g. rolling) or may suggest dynamic characteristics (e.g. color changes), such as those found in motion stop-frames.}}

\taxV{{Object motion recognition \image \audio \video}}{{}}

\vspace{-9pt}
\taxV{{Living things motion recognition \image \audio \video}}{{Recognizing movement types specific to humans and animals.}}

\taxIII{{Pose recognition}}{{The ability to infer articulated body or hand configuration, either as keypoints/joint structure or as gesture/pose classes when the task is categorical. It deserves a separate perceptual skill because many downstream tasks (e.g. action and intention understanding) depend primarily on how a person is posed rather than what objects are present. In practice, this is operationalized by standard pose-estimation datasets such as COCO Keypoints~\cite{lin2015microsoftcococommonobjects} and MPII Human Pose~\cite{andriluka20142d}, video benchmarks like PoseTrack~\cite{andriluka2018posetrack} for multi-person pose and tracking, and hand-gesture datasets such as HaGRID~\cite{kapitanov2024hagrid}, for detection- and classification-style gesture recognition.}}

\taxV{{Body pose recognition \image \video}}{{}}

\taxV{{Facial expression recognition \image \video}}{{}}
\vspace{-9pt}

\taxV{{Hand gesture recognition \image \video}}{{}}
\vspace{-9pt}

\taxII{{Fine-grained cross-instance perception}}{{Perceiving and localizing multiple objects or events and the character of their interaction.}}

\taxIII{{Overlapping object differentiation}}{{Separating co-occurring entities whose features interfere.}}


\taxV{{Overlapping image differentiation \image \video}}{{Disentangling and separate recognition of visually overlapping objects, .}}

\taxV{{Speaker diarization \audio \video}}{{Attributing speech to the correct speaker among multiple simultaneous and/or consequent speakers.}}

\taxIII{{Mutual object localization}}{{Perceiving relative relations among multiple entities.}}


\taxV{{Spatial object relationship \image \audio \video}}{{Understanding the relative localization of multiple objects in the scene.}}


\taxV{{Temporal object relationship \audio \video}}{{Understanding the ordering and temporal relations across events/segments.}}

\taxIII{{Repeating pattern recognition}}{{Perceiving instances that appear as a repeating pattern, e.g. animal flocks, ornaments, rhythmic sounds and moves, also involves perceiving frequency and density. Primary to counting skills.}}


\taxV{{Visual pattern recognition \image \video}}{{Patterns present in a single image or video frame, which do not require dynamic perception to acquire.}}

\taxV{{Temporal pattern recognition \audio \video}}{{Patterns spanning over time and / or perceived only across multiple frames.}}

\taxIII{{Cross-instance event recognition}}{{Perceiving interactions among multiple entities.}}


\taxV{{Object-object interaction \image \audio \video}}{{}}
\taxV{{Human-object interaction \image \audio \video}}{{}}
\vspace{-9pt}
\taxV{{Human-human interaction \image \audio \video}}{{}}
\vspace{-9pt}

\taxII{{Cross-modal alignment}}{{

Cross-modal alignment (Grounding) covers abilities that establish correspondences between symbolic descriptions (typically language) and specific elements in a non-text modality, enabling verification (“is the described thing present?”) and reference (“where/when is it?”). In cognitive terms, grounding mediates between perceptual evidence and symbolic reasoning, supporting tasks that require referential linkage rather than mere recognition. We separate two complementary families: (i) Text extraction — recovering textual symbols from media (e.g., OCR in images/video frames; speech recognition and lyrics recognition in audio/video) — and (ii) Media grounding — mapping language to media coordinates (e.g., bounding boxes/masks in images/video, or temporal spans in audio/video). This separation keeps transcription distinct from localization while preserving a unified notion of cross-modal correspondence.}}

\taxIII{{Image-to-text grounding}}{{Involves recognizing text in images, video frames, and more complex visual elements such as tables, charts, and diagrams. Some of the benchmarks targeting these skills together with QA tasks are PlotQA~\cite{methani2020plotqa}, ChartQA~\cite{masry2022chartqa}, Misleading ChartQA~\cite{chen-etal-2025-unmasking}, ChartBench~\cite{ChartBench}, and TextVQA~\cite{singh2019textvqa}.}}



\taxV{{Text recognition (OCR) \image \video}}{{}}

\taxV{{Scheme recognition \image \video}}{{}}
\vspace{-9pt}

\taxV{{Plot recognition \image \video}}{{}}
\vspace{-9pt}


\taxV{{Table recognition \image \video}}{{}}
\vspace{-9pt}

\taxIII{{Audio-to-text grounding}}{{Similarly to Image-to-text grounding, this skill involves matching sound signals to textual representation, including complex audio elements such as prosody, stress, and lyrics.}}


\taxV{{Prosody \& stress recognition \audio \video}}{{involves recognizing and understanding the rhythm, intonation, and stress patterns in speech.}}

\taxIV{{Speech recognition}}{{}}

\taxV{{Onomatopoeia \audio \video}}{{Onomatopoeia is a language phenomenon where words are used to imitate sounds, such as ``buzz'' for a bee or ``meow'' for a cat. Recognizing onomatopoeia is essential for understanding routine language usage. An example of a dataset for onomatopoeia recognition is RWCP-SSD-Onomatopoeia~\cite{okamoto2020rwcp}.}}

\taxV{{Speech recognition \audio \video}}{{}}

\taxV{{Song lyrics recognition \audio \video}}{{}}
\vspace{-9pt}

\taxIII{{Media grounding}}{{Grounding to media involves matching textual descriptions to media elements via masking, segmentation, or bounding boxes. }}

\taxV{{Visual media grounding \image \video}}{{Operating on bounding boxes, segmentation masks, or pixel coordinates for objects or actions described in natural language. For example, predicting the bounding box of the object in an image or the segmentation mask of the action in a video.}}

\taxV{{Temporal media grounding \audio \video}}{{Operating on timestamps and media duration. For example, predicting the start and end time of the event in a video or the duration of the action in a video.}}

\taxI{{Knowledge}}{{

Knowledge denotes a model’s ability to retrieve, apply, and integrate stored factual and conceptual information in multimodal settings. In cognitive science terms, this corresponds to long-term semantic memory (stable representations of entities, relations, norms, and domain concepts) used to interpret perceptual evidence and to support reasoning. In multimodal evaluation, knowledge competence is often probed via QA and exam-style benchmarks (e.g., MMLU/MMMU \cite{hendrycks2021measuring,yue2023mmmu}) that condition questions on images, and via cognitively motivated suites that target foundational concepts from developmental cognition (e.g., CoreCognition \cite{li2025corecognition}, which operationalizes “core knowledge” constructs such as object permanence and basic physical/social regularities). Because “knowledge” interacts with both perception and reasoning, we treat it as a distinct group to separate failures of recall/concept access from failures of inference over correctly perceived evidence.}}





Below we present a detailed description of the knowledge category, which was introduced in~\autoref{sec:taxonomy}.

\taxIII{{Knowledge}}{{In MLLM benchmarks, “knowledge” is typically observed as (i) reliance on commonsense/everyday facts in visual contexts, (ii) explicit external/encyclopedic knowledge conditioned on images, and (iii) expert-domain question answering over multimodal inputs.

The differentiation between common and domain knowledge may vary depending on the task and the domain. We formulate the polar points of this distinction as follows: common knowledge is generally agreed upon by the majority of the population and used in everyday life, while expert knowledge is domain-specific and requires specialized knowledge or training.}}

\taxIV{{Common everyday knowledge}}{{}}

\taxV{{Common everyday knowledge \image \audio \video}}{{Commonsense and broadly shared facts.}}

\taxV{{Ethics \image \audio \video}}{{Safety/ethics-aligned judgment about harmful/unsafe content or behavior in multimodal prompts. The placement of ethics in the knowledge category and treating ethical norms as stored conventions may be considered a simplification, but it is done to keep minimal the number of major categories.}}

\taxIV{{Domain knowledge}}{{This category may be further subdivided into subject-specific categories as it is done in expert benchmarks, e.g. MMMU~\cite{yue2023mmmu}.}}

\taxV{{Common domain knowledge \image \audio \video}}{{Subject-specific but broadly accessible knowledge (often school-level or general STEM).}}

\taxV{{Expert domain knowledge \image \audio \video}}{{Specialized, exam- or professional-level knowledge requiring training.}}

\taxI{{Reasoning}}{{Reasoning comprises abilities that derive new conclusions from (i) perceptual evidence and (ii) prior knowledge, including uncertain, multi-step, relational, causal, quantitative, and counterfactual inference. In cognitive terms, reasoning corresponds to controlled inference processes that operate on representations supplied by perception and memory, producing answers that are not directly observable in the input. We include both classical forms (deductive/abductive/inductive inference) and task-driven “holistic” judgments (e.g., topic, style, provenance, media characteristics) when they require integrating multiple cues across an input rather than naming a single localized entity. This placement is consistent with benchmark taxonomies that separate fine-grained recognition from higher-level judgments—for example, MMBench distinguishes coarse perception categories such as image topic/quality from fine-grained object-level recognition \cite{liu2024mmbench}. Analogously, audio evaluation benchmarks emphasize media-level characterization and instruction-guided understanding (AudioBench \cite{wang2025audiobenchuniversalbenchmarkaudio}, AIR-Bench \cite{yang2024airbenchbenchmarkinglargeaudiolanguage}, MMAU \cite{sakshi2024mmaumassivemultitaskaudio}).}}

\skillReasoningTable

An overview of the reasoning taxonomy is given in~\autoref{tab:skill_reasoning} and below a detailed skill description is provided.

%

\taxII{{Inductive reasoning}}{{}}

\taxIII{{Attribute recognition}}{{The distinction between coarse attribute recognition and object attribute recognition is inspired by MMBench's \cite{liu2024mmbench} categorization into coarse vs. fine-grained perception. We put coarse perception into reasoning category as most of the related tasks require inference from the perceived information about the whole sample.}}

\taxIV{{Coarse attribute recognition}}{{Global or “holistic” characterization of a sample (quality, topic, provenance, modality-level traits), without requiring fine object-level localization or multi-entity comparison.}}

\taxV{{Topic understanding \image \audio \video}}{{}}

\taxV{{Style \& genre understanding \image \audio \video}}{{}}
\vspace{-9pt}

\taxV{{Scene understanding \image \audio \video}}{{}}
\vspace{-9pt}

\taxV{{Generated content detection \image \audio \video}}{{Detection of whether content is synthetic vs real.}}


\taxV{{Media characteristic understanding \image \audio \video}}{{Recognize broad media-level traits (quality, augmentations). Characterizing/attributing the origin or type of synthesis (e.g., illustration type, camera quality, sound interference).

In vision-language evaluation, MMBench \cite{liu2024mmbench} explicitly includes Image Quality and Image Topic as coarse perception categories.
In audio(-text) evaluation, media-level characterization is a core target of AudioBench~\cite{wang2025audiobenchuniversalbenchmarkaudio} (speech understanding, audio scene understanding, and paralinguistic “voice understanding”), AIR-Bench~\cite{yang2024airbenchbenchmarkinglargeaudiolanguage} (speech/natural sounds/music comprehension under instruction following), and MMAU~\cite{sakshi2024mmaumassivemultitaskaudio} (multi-task audio understanding and reasoning across speech/environment/music).}}

\taxV{{Speech emotion recognition \audio \video}}{{}}
\taxV{{Music emotion recognition \image \audio \video}}{{}}
\vspace{-9pt}

\taxV{{Melodic structure interpretation \image \audio \video}}{{}}
\vspace{-9pt}

\taxIV{{Object attribute recognition}}{{Attribute-centric understanding grounded in entities (objects, people, instruments), including physical properties, functions, and identity-related cues.}}

\taxV{{Physical property understanding \image \audio \video}}{{Infer physical properties (e.g., material/shape/size/texture; or physical dynamics in video).}}

\taxV{{Object function understanding \image \audio \video}}{{Infer what an object is for and how it is used (affordances, intent).}}

\taxV{{Identity \& emotion understanding \image \audio \video}}{{Recognize identity-related cues (who) and affective state (facial expression, vocal affect).}}




\taxII{{Deductive reasoning}}{{Rule- or structure-driven inference where conclusions follow from constraints (logical, relational, symbolic, or structural).}}



\taxV{{Weirdness understanding \image \audio \video}}{{Detect incongruity, humor, anomalies, or “oddness” that violates expectations.}}

\taxV{{Analogical reasoning \image \audio \video}}{{Solve problems by mapping relational structure between situations.}}

\taxV{{Other deductive reasoning \image \audio \video}}{{General logical/relational deduction (multi-hop, constraints).}}

\taxII{{Abductive reasoning}}{{Inference to the best explanation: forming hypotheses and causal attributions consistent with observations.}}



\taxV{{Hypothetical reasoning \image \audio \video}}{{Reason about “what if” interventions and unseen outcomes.}}

\taxV{{Cause \& effect understanding \image \audio \video}}{{Identify causal responsibility and predict effects of interactions over time.}}

\taxII{{Quantitative reasoning}}{{}}


\taxIV{{Counting}}{{}}

\taxV{{Static counting \image \audio \video}}{{Count entities in a single frame/image. Complementary to pattern recognition skills that focus on repeated structures.}}

\taxV{{Temporal counting \audio \video}}{{Count events and objects over time in audio/video streams. Complementary to temporal pattern recognition skills that focus on repeated structures.}}

\taxIV{{Mathematical reasoning}}{{Solve math problems grounded in visual contexts (diagrams, plots, tables, scenes) or multimodal inputs. MathVista~\cite{lu2024mathvista} is explicitly proposed to evaluate mathematical reasoning of foundation models in visual contexts.}}

\taxV{{Mathematical reasoning \image \audio \video}}{{}}

\taxII{{Other reasoning}}{{}}



\taxV{{Critical thinking \image \audio \video}}{{Evaluate evidence, reconcile inconsistencies, and make defensible judgments under ambiguous multimodal context.}}

\taxV{{Counterfactual robustness \image \audio \video}}{{Maintain correct reasoning when interventions are applied (e.g. answer option in MCQ do not contain the correct answer, or the question relates to an entity not present in the image).}}

\taxV{{Problem decomposition \image \audio \video}}{{Solve by breaking into intermediate steps; multi-step chains across modalities. ScienceQA~\cite{lu2022learn} is explicitly proposed to evaluate problem decomposition skills of foundation models.}}

\taxV{{Comparative reasoning \image \audio \video}}{{Answer questions requiring comparisons (more/less, same/different) across instances or across time.}}

\section{Data leakage details}
\label{sec:leakage_appendix}

\noindent \textbf{Goal.} Given a multimodal example $x=(t, m)$ where $m$ is the paired modality (image/video/audio), $t$ is the text, estimate the probability that a target model was trained on $x$.

\subsection{Setup.} We begin by creating a controlled environment to simulate data leakage. For a given modality, we take a base model, $Model$. We then create a leaked version, $Model_{leak}$, by fine-tuning the base model on a subset of our benchmark data using SFT-LoRA for the selected modality.

\subsection{Neighbor Generation and Feature Extraction.} For each original data point $(t, m)$ we generate $K=24$ perturbed "neighbors". We apply four distinct perturbation techniques (masking and predicting the masks with Fred-T5 model\footnote{\url{https://hf.co/ai-forever/FRED-T5-1.7B}, \cite{zmitrovich2023family}}, deletion, duplication, and swapping of random words) to the text $t$ with each technique applied 6 times. The modality data $m$ remains unchanged.

For each original text $t$ and its neighbors $t_k^\prime$ we extract their text embeddings using a fixed encoder:
$$e=E(t), \quad e_{k}^{\prime} = E(t_k^{\prime})$$
where $E$ is \texttt{intfloat/e5-mistral-7b-instruct}\footnote{\url{https://hf.co/intfloat/e5-mistral-7b-instruct}. It used to be SoTA on MTEB benchmark~\cite{muennighoff2022mteb}} model.

Subsequently, we compute the multimodal loss for both models $Model$ and $Model_{leak}$ on both the original and neighbor data points:
$$\mathcal{L} = L(Model, t, m), \quad \mathcal{L}_k^{\prime} = L(Model, t_k^{\prime}, m)$$

\subsection{Detector Training}

The core of MSMIA is a binary classifier trained to distinguish between models that have and have not seen the data. For each neighbor $k$ we create two training examples by computing the feature differences:
$$\Delta \mathcal{L} = \mathcal{L} - \mathcal{L}_k^{\prime}, \quad \Delta e = e - e_{k}^{\prime}$$

These feature vectors are paired with labels $y \in \{0, 1\}$ indicating whether the losses came from $Model$ ($y=0$) or $Model_{leak}$ ($y=1$). This process yields 48 training triplets $(\Delta \mathcal{L}, \Delta e, y)$ per original data point. The MSMIA detector, $f_{MSMIA}$ is trained to predict the probability $p=f_{MSMIA}(\Delta \mathcal{L}, \Delta e)$ that the input features originate from a model that has been trained on the target data.

\subsection{Inference and Evaluation}

To infer if a target $Test Model$ has been trained on a specific data point $(t, m)$, we repeat Step 2 to compute the loss and embedding differences for this model. We then compute the leakage score $S$ for the data point by taking the average probability output by the detector over all $K$ neighbors:
$$S(t, m) = \frac{1}{K}\sum_{k=1}^{K}f_{MSMIA}(\Delta \mathcal{L}_k, \Delta e_k)$$

To get the probability estimation for the entire dataset, the $S$ scores are averaged over all dataset samples.

\begin{table}[t]
\scriptsize
\centering
 \begin{tabular}{l l c}
\toprule
\textbf{Origin Model} & \textbf{Test Model} & \textbf{AUC-ROC} \\
\midrule
Qwen2.5-VL-3B-Instruct & Qwen2.5-VL-3B-Instruct & 96.2 \\
Qwen2.5-VL-3B-Instruct & Qwen2-VL-7B-Instruct & 86.0 \\
Qwen2.5-VL-3B-Instruct & Qwen2.5-VL-7B-Instruct & 88.0 \\
Qwen2.5-VL-3B-Instruct & llama3-llava-next-8b-hf & 90.2 \\
Qwen2.5-VL-3B-Instruct & gemma-3-4b-it & 65.8 \\
Qwen2.5-VL-3B-Instruct & gemma-3-12b-it & 67.9 \\
Qwen2-VL-7B-Instruct & Qwen2.5-VL-3B-Instruct & 78.0 \\
Qwen2-VL-7B-Instruct & Qwen2-VL-7B-Instruct & 96.2 \\
Qwen2-VL-7B-Instruct & Qwen2.5-VL-7B-Instruct & 80.5 \\
Qwen2-VL-7B-Instruct & llama3-llava-next-8b-hf & 78.0 \\
Qwen2-VL-7B-Instruct & gemma-3-4b-it & 77.7 \\
Qwen2-VL-7B-Instruct & gemma-3-12b-it & 73.7 \\
Qwen2.5-VL-7B-Instruct & Qwen2.5-VL-3B-Instruct & 92.8 \\
Qwen2.5-VL-7B-Instruct & Qwen2-VL-7B-Instruct & 93.1 \\
Qwen2.5-VL-7B-Instruct & Qwen2.5-VL-7B-Instruct & 98.1 \\
Qwen2.5-VL-7B-Instruct & llama3-llava-next-8b-hf & 95.8 \\
Qwen2.5-VL-7B-Instruct & gemma-3-4b-it & 95.4 \\
Qwen2.5-VL-7B-Instruct & gemma-3-12b-it & 94.5 \\
llama3-llava-next-8b-hf & Qwen2.5-VL-3B-Instruct & 94.6 \\
llama3-llava-next-8b-hf & Qwen2-VL-7B-Instruct & 90.0 \\
llama3-llava-next-8b-hf & Qwen2.5-VL-7B-Instruct & 96.6 \\
llama3-llava-next-8b-hf & llama3-llava-next-8b-hf & 97.7 \\
llama3-llava-next-8b-hf & gemma-3-4b-it & 99.1 \\
llama3-llava-next-8b-hf & gemma-3-12b-it & 99.5 \\
gemma-3-4b-it & Qwen2.5-VL-3B-Instruct & 76.0 \\
gemma-3-4b-it & Qwen2-VL-7B-Instruct & 71.5 \\
gemma-3-4b-it & Qwen2.5-VL-7B-Instruct & 85.2 \\
gemma-3-4b-it & llama3-llava-next-8b-hf & 86.5 \\
gemma-3-4b-it & gemma-3-4b-it & 99.4 \\
gemma-3-4b-it & gemma-3-12b-it & 98.7 \\
gemma-3-12b-it & Qwen2.5-VL-3B-Instruct & 84.1 \\
gemma-3-12b-it & Qwen2-VL-7B-Instruct & 81.3 \\
gemma-3-12b-it & Qwen2.5-VL-7B-Instruct & 91.2 \\
gemma-3-12b-it & llama3-llava-next-8b-hf & 93.3 \\
gemma-3-12b-it & gemma-3-4b-it & 99.4 \\
gemma-3-12b-it & gemma-3-12b-it & 99.7 \\
\bottomrule
\end{tabular}
\caption{AUC-ROC MSMIA performance metrics for various evaluated Image MLLMs.}
\label{tab:msmia_results_image}
\end{table}

\begin{table}[t]
\scriptsize
\centering
\begin{tabular}{l l c}
\toprule
\textbf{Origin Model} & \textbf{Test Model} & \textbf{AUC-ROC} \\
\midrule
Qwen2.5-VL-3B-Instruct & Qwen2.5-VL-3B-Instruct & 95.9 \\
Qwen2.5-VL-3B-Instruct & Qwen2.5-VL-7B-Instruct & 99.5 \\
Qwen2.5-VL-3B-Instruct & LLaVA-NeXT-Video & 91.7 \\
Qwen2.5-VL-3B-Instruct & LLaVA-NeXT-Video-DPO & 91.2 \\
Qwen2.5-VL-7B-Instruct & Qwen2.5-VL-3B-Instruct & 98.7 \\
Qwen2.5-VL-7B-Instruct & Qwen2.5-VL-7B-Instruct & 100.0 \\
Qwen2.5-VL-7B-Instruct & LLaVA-NeXT-Video & 96.5 \\
Qwen2.5-VL-7B-Instruct & LLaVA-NeXT-Video-DPO & 95.7 \\
LLaVA-NeXT-Video & Qwen2.5-VL-3B-Instruct & 63.7 \\
LLaVA-NeXT-Video & Qwen2.5-VL-7B-Instruct & 71.5 \\
LLaVA-NeXT-Video & LLaVA-NeXT-Video & 100.0 \\
LLaVA-NeXT-Video & LLaVA-NeXT-Video-DPO & 100.0 \\
LLaVA-NeXT-Video-DPO & Qwen2.5-VL-3B-Instruct & 53.6 \\
LLaVA-NeXT-Video-DPO & Qwen2.5-VL-7B-Instruct & 56.2 \\
LLaVA-NeXT-Video-DPO & LLaVA-NeXT-Video & 100.0 \\
LLaVA-NeXT-Video-DPO & LLaVA-NeXT-Video-DPO & 100.0 \\
\bottomrule
\end{tabular}
\caption{AUC-ROC MSMIA performance metrics for various evaluated Video MLLMs.}
\label{tab:msmia_results_video}
\end{table}

\begin{table}[t]
\scriptsize
\centering
\begin{tabular}{l l c}
\toprule
\textbf{Origin Model} & \textbf{Test Model} & \textbf{AUC-ROC} \\
\midrule
Qwen2-Audio-7B-Instruct & Qwen2-Audio-7B-Instruct & 87.7 \\
Qwen2-Audio-7B-Instruct & Qwen-Audio-Chat & 76.0 \\
Qwen-Audio-Chat & Qwen2-Audio-7B-Instruct & 61.3 \\
Qwen-Audio-Chat & Qwen-Audio-Chat & 100.0 \\
\bottomrule
\end{tabular}
\caption{AUC-ROC MSMIA performance metrics for various evaluated Audio MLLMs.}
\label{tab:msmia_results_audio}
\end{table}

We report AUC-ROC for binary classification (leaked vs. clean) as shown in Tables \ref{tab:msmia_results_image}, \ref{tab:msmia_results_audio}, \ref{tab:msmia_results_video}. There the \texttt{Origin Model} is the model used to train MSMIA. \texttt{Test Model} is the model whose losses are used to test MSMIA (predict whether the data sample was used to train \texttt{Test Model} or not).

\section{LLM-as-a-judge details}
\label{sec:judge_appendix}

\subsection{Data collection}

Our dataset comprises (question, gold answer, model prediction) triplets sourced from Russian-language benchmarks, including \meramulti, with model sizes spanning 2B to 110B parameters. Human annotators label the semantic correctness of each prediction, strictly ignoring surface form. To ensure label quality, only items with 100\% inter-annotator agreement are used for model training and testing.

\begin{table}[ht!]
    \centering
    \scriptsize
    \begin{tabular}{lccc}
    \toprule
     & \textbf{Total} & \textbf{Open Generation} & \textbf{Multiple Choice} \\
    \midrule 
    \textbf{Train set} & 62,580 & 22,538 & 40,042 \\ 
    \textbf{Test set} & 8,570 & 5,819 & 2,751 \\
    \bottomrule 
    \end{tabular}
    \caption{Overview of the datasets used for training and evaluating the judge model. The columns indicate the total number of examples and their distribution across two task formats: \textit{Open Generation} (free-form model response) and \textit{Multiple Choice} (discrete answer selection)}
    \label{tab:judge_data}
\end{table}

We apply synthetic augmentations (e.g., refusals, repetitions) to improve robustness. A critical measure to prevent bias was splitting the data by source dataset, ensuring no dataset appears in both train and test splits. The final dataset composition and class balance are detailed in \autoref{tab:judge_data}.

\subsection{Training models}

The tasks are formulated as a binary classification problems. Inputs are linearized into the format \texttt{question [SEP] gold answer [SEP] model prediction} and packed into the maximum models' contexts. We train encoder-based models, adding a linear classification head on top of the pre-trained backbone. The models are fully fine-tuned and optimized using a cross-entropy loss, with class weights applied to mitigate potential dataset imbalance.

\begin{table}[ht!]
    \centering
    \small
    \begin{tabular}{lc}
    \toprule
    \textbf{Parameter} & \textbf{Value} \\
    \midrule
    Learning Rate & 2e-5 \\
    Batch Size & 32 \\
    Number of Epochs & 3 \\
    Weight Decay & 0.01 \\
    Optimizer & AdamW \\
    LR Scheduler & Linear \\
    Max Sequence Length & 512 \\
    Precision & BF16 \\
    Early Stopping Metric & F1 Score \\
    \bottomrule
    \end{tabular}
    \caption{Training configuration summary for embedding-based classification}
    \label{tab:judge_hparams}
\end{table}

Optimization is performed with mixed precision and early stopping based on the F1 score on the development set, constrained to a single A100 80GB GPU. A summary of the final training configuration is provided in \autoref{tab:judge_hparams}.

\subsection{Model selection}

We compare fine-tuned encoder-based models against zero-shot decoder baselines to identify the optimal judge architecture. The decoder models are prompted to output either ``0'' or ``1'' without any task-specific fine-tuning. While this approach is flexible, its reliability is limited by the fact that decoder generations are unconstrained; they may not always produce a valid classification, making score interpretation ambiguous.

In our experiments, encoder models proved to be more suitable for this binary classification task. Their architectural design naturally provides probability distributions over the two classes (correct/incorrect), ensuring deterministic scoring. Furthermore, they offer significant practical advantages, being generally smaller, faster on a single GPU, and even capable of handling long contexts (useful for judging the answers with reasoning or chain-of-thought elements). A full model comparison with bootstrap statistics is provided in Table~\ref{tab:judge_overall}.

\begin{table*}[ht!]
    \centering
    \scriptsize
    \begin{tabular}{c|l|ccc|ccccc}
    \toprule
     & \textbf{Model} & \textbf{Parameters} & \textbf{Context Length} & \textbf{Samples/sec} & \textbf{F1} & \textbf{Recall} & \textbf{Precision} & \textbf{Pearson} & \textbf{EM rate} \\
    \midrule
\multirow{6}{*}{\rotatebox[origin=c]{90}{\textbf{Encoders}}}
& \href{https://huggingface.co/deepvk/RuModernBERT-base}{RuModernBERT-base} & 150M & 8,192 & 1800 & 0.964 $\pm$ 0.002 & 0.975 & 0.955 & 0.940 & 0.997 \\
& \href{https://huggingface.co/deepvk/RuModernBERT-small}{RuModernBERT-small} & 35M & 8,192 & 2000 & 0.944 $\pm$ 0.003 & 0.962 & 0.927 & 0.879 & 0.982\\
& \href{https://huggingface.co/Qwen/Qwen3-Embedding-0.6B}{Qwen3-Embedding-0.6B} & 600M & 32,768 & 1850 & 0.951 $\pm$ 0.003 & 0.985 & 0.920 & 0.936 & 1.000\\
& \href{https://huggingface.co/ai-forever/FRIDA}{FRIDA} & 823M & 512 & 70* & 0.838 $\pm$ 0.005 & 0.913 & 0.774  & 0.765 & 0.925 \\
& \href{https://huggingface.co/google/embeddinggemma-300m}{embeddinggemma-300m} & 303M & 2,048 & 250* & 0.915 $\pm$ 0.004 & 0.965 & 0.871 & 0.825 & 0.997\\
& \href{https://huggingface.co/ai-sage/Giga-Embeddings-instruct}{Giga-Embeddings-instruct} & 3.45B & 4,096 & 40* & 0.966 $\pm$ 0.002 & 0.984 & 0.948 & 0.946 & 1.000\\
\midrule
\multirow{5}{*}{\rotatebox[origin=c]{90}{\textbf{Decoders}}}
& \href{https://huggingface.co/ai-forever/pollux-judge-7b}{pollux-judge-7b} & 7.61B & 131072 & 20 & 0.823 $\pm$ 0.005 & 0.788 & 0.860& 0.732 & 0.873\\
& \href{https://huggingface.co/t-tech/T-lite-it-1.0}{T-lite-it-1.0} & 7.61B & 131072 & 52 & 0.844 $\pm$ 0.005 & 0.968 & 0.747 & 0.755 & 0.993\\
& \href{https://huggingface.co/Qwen/Qwen3-0.6B}{Qwen3-0.6B} & 752M & 32768 & 442 & 0.030 $\pm$ 0.003 & 0.016 & 0.323 & -0.010 & 0.004\\
& \href{https://huggingface.co/Qwen/Qwen3-1.7B}{Qwen3-1.7B} & 2.03B & 32768 & 205 & 0.590 $\pm$ 0.006 & 0.990 & 0.420 & 0.300  & 0.997\\
& \href{https://huggingface.co/openai/gpt-oss-20b}{gpt-oss-20b} & 21.5B & 131072 & 28 & 0.939 $\pm$ 0.003 & 0.968 & 0.912 & 0.905 & 1.000\\
    \bottomrule
    \end{tabular}
    \caption{Summary results for judge models on the test set. F1, Recall and Precision report binary classification quality; Pearson is the Pearson correlation between binary predictions and ground-truth labels; EM rate is the share of class-1 predictions on the subset of examples with an exact string match between the prediction and the gold answer. An asterisk (*) next to throughput indicates the model could not be served via vLLM and the speed was measured with the HuggingFace Trainer instead}
    \label{tab:judge_overall}
\end{table*}

\subsection{Model deployment}

The selected judge model is deployed using vLLM for high-throughput inference on a single A100 40GB GPU. This provides optimal batch processing speed for rapid benchmark evaluation while maintaining quality. Our codebase implements a \texttt{score(q, ref, pred) → \{0,1\}} API that handles individual scoring and aggregation. We also publicly provide the trained model with weights on HuggingFace Hub~\footnote{\url{https://huggingface.co/MERA-evaluation/MERA_Answer_judge}}.

\subsection{Model analysis}

To validate our LLM-as-a-judge, we performed a sanity check demonstrating high heuristic consistency: the model aligns with human labels in 99.6\% of cases where Exact Match (EM) is 1, and 97.7\% when EM is 0 but humans deem the response correct. This reliability extends across task types, with agreement rates of 98.7\% for multiple-choice and 96\% for free-form generation. Statistical analysis confirms the judge is resilient to common biases, showing near-zero Pearson and Spearman correlations with answer length ($r \approx 0.001$, $\rho \approx -0.008$) and gold label position ($r/\rho \approx -0.0005$). Qualitative analysis reveals that errors primarily occur in tasks involving complex LaTeX formatting or multi-step reasoning. Since our model is designed as a binary classifier without a rationale-generation component, it occasionally struggles with the ambiguous justifications or symbolic intricacies inherent in these specialized domains.

\section{Block prompts analysis}
\label{sec:appendix_prompts_analysis}

\subsection{Prompts formulations}

There are 13 blocks:
\begin{itemize}[nosep]
    \item Attention hook (greeting / draw attention).
    \item General task description (task specific).
    \item Input data description (enumerate modalities).
    \item Action on data (e.g., ``Solve the task using the images…'').
    \item Optional task specifics (helpful but nonessential context).
    \item Textual question.
    \item Answer options (if multiple choice).
    \item Call to solve (explicit request to answer).
    \item Reasoning request (ask for thinking before final answer; present in 5 prompts).
    \item Reasoning format (how to present the reasoning).
    \item Answer format (how to present the final answer).
    \item Time limitation (e.g., ``you have 10 minutes'').
    \item Final call to action (e.g., ``get started'').
\end{itemize}

The blocks are combined with the Python script, following the task prompts configuration file. This file contains an explicit description of all ten prompts. Example of the prompt configuration:

\begin{lstlisting}[style=myyaml, language={} ]
# prompt_config.yaml
prompt_4:
 attention_hook: "informal_request"
 task_description: "informal_request"
 input_data: "default"
 processing_data: "informal_request"
 context_intro: "in_dataset"
 task_context: "default"
 question: "default"
 answer_options: "default"
 solution_motivation: "informal_request"
 reasoning_motivation: "none"
 reasoning_format: "none"
 answer_format: "informal_request"
 limitations: "informal_request"
 answer_motivation: "informal_request"
\end{lstlisting}

The block formulations are fixed. The only variable blocks are \texttt{task\_description} and \texttt{task\_context} that are task-specific. All prompt formulations are imputed in the dataset on HuggingFace Hub (key ``instruction'' of each data sample). The blocks distribution is fixed: 5 prompts with \texttt{reasoning\_motivation} and \texttt{reasoning\_format} blocks to enable answer rationale, 1 prompt with no \texttt{answer\_format} (``zero prompt'' that provides the minimal version of the task - only placeholders for multimodal data, the question and answer options if any).

The example of the prompt with all blocks are as follows.

``Zero prompt'':

\begin{mdframed}[
    userdefinedwidth=0.9\columnwidth,
    align=center
]
\begin{lstlisting}
Image: <image>
Question:
{question}
\end{lstlisting}
\end{mdframed}

Prompt with specific answer format (\ruslun):

\begin{mdframed}[
    userdefinedwidth=0.9\columnwidth,
    align=center
]
\begin{lstlisting}
The dataset for the task includes the following prompt:

Not all slots are necessarily present in the request.

Audio file: <audio>
Question:
{question}

Please solve the task based on the above and briefly formulate your answer.

{annotation}
\end{lstlisting}
\end{mdframed}

Example of the prompt with ``reasoning'' blocks:

\begin{mdframed}[
    userdefinedwidth=0.9\columnwidth,
    align=center
]
\begin{lstlisting}
The dataset for the task includes the following prompt:

The question is directly related to the content of the image and requires not only recognition of individual elements, but also understanding of the relationships between the elements (objects) in the image. If there is insufficient information to answer the question, for example, if the object in question is missing from the image, then you must honestly answer that the question cannot be answered and indicate the reason.

Image: <image>
Question:
{question}

Please solve the task based on the above and briefly formulate your answer.

Please think about the solution and describe your thought process in detail.

Write your reasoning after the word REASONING, briefly explaining how you arrived at your final answer.

Please provide a brief answer to the question. Please do not write anything else, do not elaborate, do not engage in dialogue, and do not explain your answer. Please write your final answer after the word ANSWER.
\end{lstlisting}
\end{mdframed}

\subsection{Statistical analysis}

To analyze the effect that a prompt formulation has on the metrics, we fit OLS with the following specification:
\begin{equation}
    metric \sim  C(prompt) + C(model) + C(engine)
\end{equation}

In fact, it is essentially the same as:
\begin{equation}
    \begin{split}
    metric_i = \alpha + \sum_{p=2}^{10} \beta_p \{prompt = p\} + \\ \sum_{m} \gamma_m \{model = m\} + \sum_{e} \delta_e \{engine = e\}
    \end{split}
\end{equation}

Where:
\begin{itemize}[nosep]
    \item \texttt{Prompt} is a specific prompt formulation (one of 10).
    \item \texttt{Model} is the model name - the model used to infer the data with the prompt and the metric \texttt{metric}.
    \item \texttt{Engine} is the inference backend used for evaluation (one of \texttt{transformers}~\citep{wolf-etal-2020-transformers} or \texttt{vllm}\footnote{\url{https://github.com/vllm-project/vllm}}).
    \item \texttt{Frames}. Additional categorical variable used only for video modality evaluations - the video is uniformly split into $N$ frames and the vision LLM used to make evaluation on the dataset.
    \item \texttt{Domain}. Additional categorical variable used only for \unisciencevqa, \schoolsciencevqa, \rucommonvqa. These datasets are split into separate domains (subsets). This split may affect the metrics (one domain may be ``harder'' than another).
\end{itemize}

For each prompt $p$, we test $H_0: \beta_p = 0$ against $H_1: \beta_p \ne 0$. We mark a prompt's effect as statistically significant when the two-sided p-value $< 0.05$.

\begin{figure*}
    \centering
    \includegraphics[width=1\linewidth]{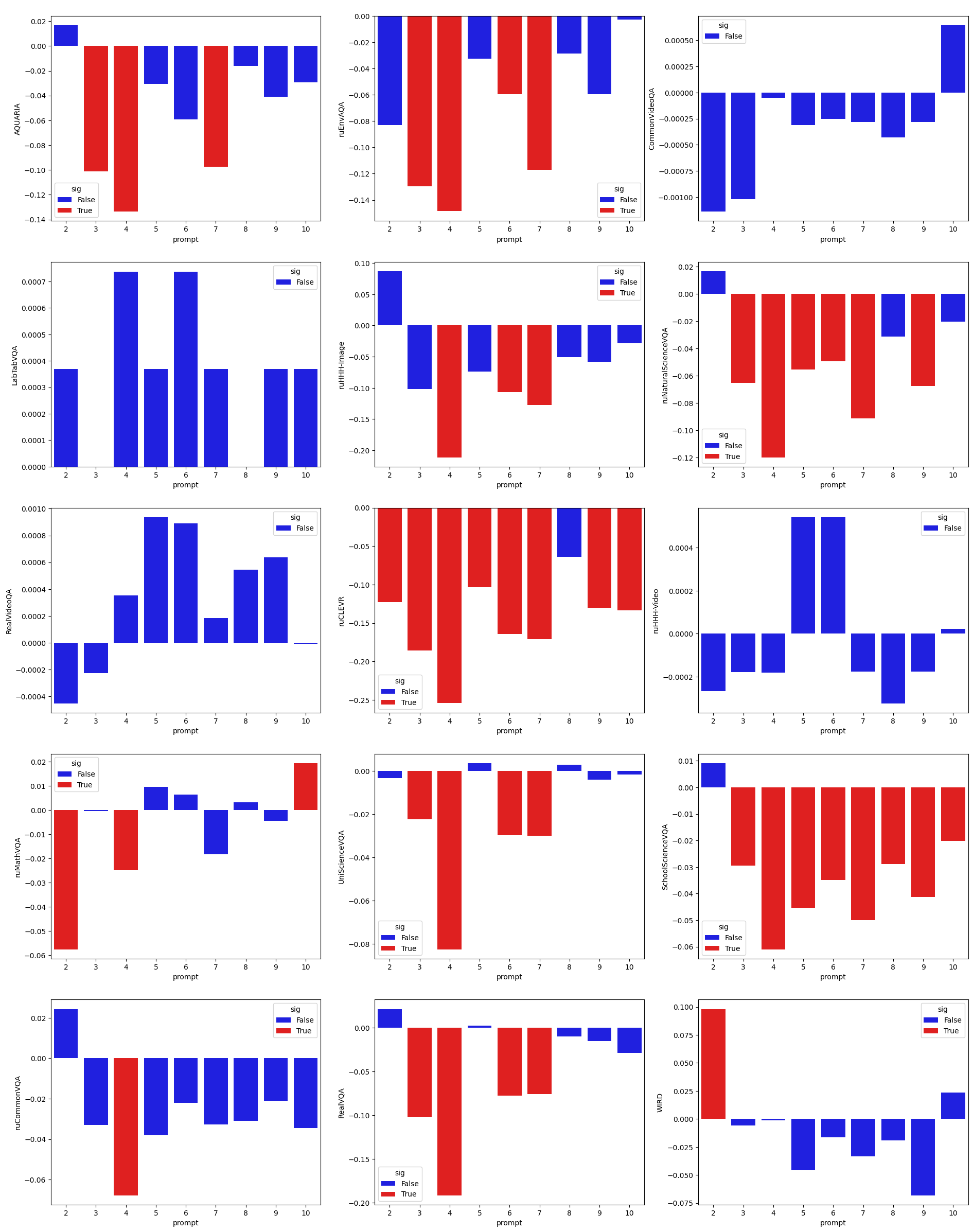}
    \caption{The relative (with regard to baseline prompt (0)) effects of different formulations of prompts for each dataset. There are ten different formulations of prompts for one dataset, hence nine corresponding bars (one formulation is baseline category). Red bars reflect statistically significant (at 95\% confidence level) effects.}
    \label{fig:prompts-effects}
\end{figure*}

Figure \ref{fig:prompts-effects} demonstrates the results of statistical analysis of the prompts formultations effects on the \texttt{Judge Score} metric. There are three datasets that have been omitted: \ruslun implies structured output with rather strict metric that tends to show zero score, \rutieaudio and \rutieimage datasets scores are too sensible to prompt formulation due to the datasets design\footnote{Both datasets consist of three sequential dialogues of 500 questions in a row. The answer for the question $X$ lays in the text of the questions $>X$. This way we cannot reliably separate this effect from the pure effect of the prompt formulation which may lead to incorrect analysis and conclusions.}

\noindent \textbf{Takeaways:} 
\begin{itemize}[nosep]
    \item \textbf{No single prompt dominates}. Different datasets favor different formulations; the prompt that helps in one case can hurt in another. This undermines any one-prompt-fits-all strategy.
    \item \textbf{Magnitude varies widely across datasets}. Some tasks show large shifts (on the order of ~0.1–0.2 in the metric), while others exhibit near-zero effects (often still significant when variance is low). Prompt sensitivity is therefore task-dependent.
    \item \textbf{Reasoning format does not mean lower scores}. Prompts that explicitly encourage model to provide chain-of-though rationale for the final answer do not always get lower scores. First, this means that the LLM-as-judge model is capable of coping with reasoning format. Second, some models breach the answer format prescribed in the task instruction.
    \item \textbf{Single prompt leads to bias}. Only four tasks (notable, three of them are from video modality) demonstrate no red bars - no statistically significant prompt formulations compared to the baseline one. The other 12 datasets tend to be prompt sensitive, so the design choice to distribute prompts uniformly and interpret dataset metrics as an average over formulations is empirically supported.
\end{itemize}

\section{Baselines Details}
\subsection{Model Baselines Details}
\label{app:model_baselines_details}

In this section, the list of baseline models is provided. Tables \ref{tab:baselines_models_image}, \ref{tab:baselines_models_audio}, \ref{tab:baselines_models_video} represent models for image, audio, and video modalities respectively.
\begin{table*}[t]
    \setlength{\tabcolsep}{3pt}
    \centering
    \small
    \begin{tabularx}{\textwidth}{%
    @{}c%
    p{0.22\linewidth}%
    r%
    r%
    p{0.30\linewidth}%
    l%
    @{}l@{}}
        \toprule
        & \textbf{Model} & \textbf{Parameters} & \textbf{Context length} & \textbf{Hugging Face Hub link} & \textbf{Citation} \\
        \midrule
        & GPT 4.1 & N/A & 1000K & \href{https://openai.com/index/gpt-4-1/}{GPT 4.1} &  \\
        \cmidrule{2-6}
        & Phi-3.5-vision-instruct & 4B & 128K & \href{https://huggingface.co/microsoft/Phi-3.5-vision-instruct}{microsoft/Phi-3.5-vision-instruct} & \citet{abdin2024phi3technicalreporthighly} \\
        & Phi-4-multimodal-instruct & 6B & 128K & \href{https://huggingface.co/microsoft/Phi-4-multimodal-instruct}{microsoft/Phi-4-multimodal-instruct} & \citet{microsoft2025phi4minitechnicalreportcompact} \\
        \cmidrule{2-6}
        & Qwen2.5-Omni-3B & 6B & 32K & \href{https://huggingface.co/Qwen/Qwen2.5-Omni-3B}{Qwen/Qwen2.5-Omni-3B} & \multirow{2}{*}{\citet{xu2025qwen25omnitechnicalreport}} \\
        & Qwen2.5-Omni-7B & 11B & 32K & \href{https://huggingface.co/Qwen/Qwen2.5-Omni-7B}{Qwen/Qwen2.5-Omni-7B} & \\
        \cmidrule{2-6}
        & Qwen2-VL-2B-Instruct & 2B & 32K & \href{https://huggingface.co/Qwen/Qwen2-VL-2B-Instruct}{Qwen/Qwen2-VL-2B-Instruct} & \multirow{3}{*}{\citet{wang2024qwen2vlenhancingvisionlanguagemodels}}\\
        & Qwen2-VL-7B-Instruct & 7B & 32K & \href{https://huggingface.co/Qwen/Qwen2-VL-7B-Instruct}{Qwen/Qwen2-VL-7B-Instruct} & \\
        & Qwen2-VL-72B-Instruct & 72B & 32K & \href{https://huggingface.co/Qwen/Qwen2-VL-72B-Instruct}{Qwen/Qwen2-VL-72B-Instruct} & \\
        \cmidrule{2-6}
        & Qwen3-VL-2B-Instruct & 2B & 262K & \href{https://huggingface.co/Qwen/Qwen3-VL-2B-Instruct}{Qwen/Qwen3-VL-2B-Instruct} & \multirow{2}{*}{\citet{yang2025qwen3technicalreport}}\\
        & Qwen3-VL-8B-Instruct & 9B & 262K & \href{https://huggingface.co/Qwen/Qwen3-VL-8B-Instruct}{Qwen/Qwen3-VL-8B-Instruct} & \\
        \cmidrule{2-6}
        & Qwen2.5-VL-3B-Instruct & 3B & 128K & \href{https://huggingface.co/Qwen/Qwen2.5-VL-3B-Instruct}{Qwen/Qwen2.5-VL-3B-Instruct} & \multirow{4}{*}{\citet{bai2025qwen25vltechnicalreport}}\\
        & Qwen2.5-VL-7B-Instruct & 7B & 128K & \href{https://huggingface.co/Qwen/Qwen2.5-VL-7B-Instruct}{Qwen/Qwen2.5-VL-7B-Instruct} & \\
        & Qwen2.5-VL-32B-Instruct & 32B & 128K & \href{https://huggingface.co/Qwen/Qwen2.5-VL-32B-Instruct}{Qwen/Qwen2.5-VL-32B-Instruct} & \\
        & Qwen2.5-VL-72B-Instruct & 72B & 128K & \href{https://huggingface.co/Qwen/Qwen2.5-VL-72B-Instruct}{Qwen/Qwen2.5-VL-72B-Instruct} & \\
        \cmidrule{2-6}
        & gemma-3-12b-it & 12B & 128K & \href{https://huggingface.co/google/gemma-3-12b-it}{google/gemma-3-12b-it} & \multirow{1}{*}{\citet{gemmateam2025gemma3technicalreport}} \\
        \cmidrule{2-6}
        & llava-1.5-13b-hf & 13B & 4K & \href{https://huggingface.co/llava-hf/llava-1.5-13b-hf}{llava-hf/llava-1.5-13b-hf} & \multirow{1}{*}{\citet{liu2024improvedllava}} \\
        \cmidrule{2-6}
        & llava-next-72b-hf & 72B & 8K & \href{https://huggingface.co/llava-hf/llava-next-72b-hf}{llava-hf/llava-next-72b-hf} & \multirow{2}{*}{\citet{li2024llavanext-strong}} \\
        & llava-next-110b-hf & 110B & 32K & \href{https://huggingface.co/llava-hf/llava-next-110b-hf}{llava-hf/llava-next-110b-hf} & \\
        \cmidrule{2-6}
        & InternVL3-9B & 9B & 32K & \href{https://huggingface.co/OpenGVLab/InternVL3-9B}{OpenGVLab/InternVL3-9B} & \multirow{1}{*}{\citet{zhu2025internvl3}} \\
        \cmidrule{2-6}
        & granite-vision-3.3-2b & 3B & 128K & \href{https://huggingface.co/ibm-granite/granite-vision-3.3-2b}{ibm-granite/granite-vision-3.3-2b} & \multirow{1}{*}{\citet{granitevisionteam2025granitevisionlightweightopensource}} \\
        \cmidrule{2-6}
        & SmolVLM-Instruct & 2B & 16K & \href{https://huggingface.co/HuggingFaceTB/SmolVLM-Instruct}{HuggingFaceTB/SmolVLM-Instruct} & \multirow{1}{*}{\citet{marafioti2025smolvlmredefiningsmallefficient}} \\
        \cmidrule{2-6}
        & MiniCPM-o-2\_6 & 9B & 32K & \href{https://huggingface.co/openbmb/MiniCPM-o-2_6}{openbmb/MiniCPM-o-2\_6} & \multirow{1}{*}{\citet{yao2024minicpmvgpt4vlevelmllm}} \\
        \bottomrule
    \end{tabularx}
    \caption{General information about vision (image) modality baseline models.}
    \label{tab:baselines_models_image}
\end{table*}
\begin{table*}[t]
    \setlength{\tabcolsep}{3pt}
    \centering
    \small
    \begin{tabularx}{\textwidth}{%
    @{}c%
    p{0.25\linewidth}%
    r%
    r%
    p{0.32\linewidth}%
    l%
    @{}l@{}}
        \toprule
        & \textbf{Model} & \textbf{Parameters} & \textbf{Context length} & \textbf{Hugging Face Hub link} & \textbf{Citation} \\
        \midrule
        & Qwen3-Omni-30B-A3B-Instruct & 35B & 8K & \href{https://huggingface.co/Qwen/Qwen3-Omni-30B-A3B-Instruct}{Qwen/Qwen3-Omni-30B-A3B-Instruct} & \\
        \cmidrule{2-6}
        & Qwen2.5-Omni-3B & 6B & 32K & \href{https://huggingface.co/Qwen/Qwen2.5-Omni-3B}{Qwen/Qwen2.5-Omni-3B} & \multirow{2}{*}{\citet{xu2025qwen25omnitechnicalreport}} \\
        & Qwen2.5-Omni-7B & 11B & 32K & \href{https://huggingface.co/Qwen/Qwen2.5-Omni-7B}{Qwen/Qwen2.5-Omni-7B} & \\
        \cmidrule{2-6}
        & Qwen2-Audio-7B-Instruct & 7B & 32K & \href{https://huggingface.co/Qwen/Qwen2-Audio-7B-Instruct}{Qwen/Qwen2-Audio-7B-Instruct} & \multirow{2}{*}{\citet{chu2024qwen2audiotechnicalreport}}\\
        & Qwen2-Audio-7B & 8B & 32K & \href{https://huggingface.co/Qwen/Qwen2-Audio-7B}{Qwen/Qwen2-Audio-7B} & \\
        \cmidrule{2-6}
        & audio-flamingo-3-hf & 9B & 32K & \href{https://huggingface.co/nvidia/audio-flamingo-3-hf}{nvidia/audio-flamingo-3-hf} & \multirow{1}{*}{\citet{goel2025audioflamingo3advancing}}\\
        \cmidrule{2-6}
        & SeaLLMs-Audio-7B & 8B & 8K & \href{https://huggingface.co/SeaLLMs/SeaLLMs-Audio-7B}{SeaLLMs/SeaLLMs-Audio-7B} & \multirow{1}{*}{\citet{zhang2024seallms3openfoundation}}\\
        \cmidrule{2-6}
        & ultravox-v0\_2 & 8B & 8K & \href{https://huggingface.co/fixie-ai/ultravox-v0_2}{fixie-ai/ultravox-v0\_2} & \multirow{10}{*}{\citet{team2024ultravox}}\\
        & ultravox-v0\_3 & 8B & 8K & \href{https://huggingface.co/fixie-ai/ultravox-v0_3}{fixie-ai/ultravox-v0\_3} & \\
        & ultravox-v0\_3-llama-3\_2-1b & 2B & 128K & \href{https://huggingface.co/fixie-ai/ultravox-v0_3-llama-3_2-1b}{fixie-ai/ultravox-v0\_3-llama-3\_2-1b} & \\
        & ultravox-v0\_4 & 8B & 8K & \href{https://huggingface.co/fixie-ai/ultravox-v0_4}{fixie-ai/ultravox-v0\_4} & \\
        & ultravox-v0\_4\_1-llama-3\_1-8b & 8B & 128K & \href{https://huggingface.co/fixie-ai/ultravox-v0_4_1-llama-3_1-8b}{fixie-ai/ultravox-v0\_4\_1-llama-3\_1-8b} & \\
        & ultravox-v0\_4\_1-mistral-nemo & 13B & 128K & \href{https://huggingface.co/fixie-ai/ultravox-v0_4_1-mistral-nemo}{fixie-ai/ultravox-v0\_4\_1-mistral-nemo} & \\
        & ultravox-v0\_5-llama-3\_2-1b & 2B & 128K & \href{https://huggingface.co/fixie-ai/ultravox-v0_5-llama-3_2-1b}{fixie-ai/ultravox-v0\_5-llama-3\_2-1b} & \\
        & ultravox-v0\_5-llama-3\_1-8b & 8B & 128K & \href{https://huggingface.co/fixie-ai/ultravox-v0_5-llama-3_1-8b}{fixie-ai/ultravox-v0\_5-llama-3\_1-8b} & \\
        & ultravox-v0\_6-llama-3\_1-8b & 8B & 128K & \href{https://huggingface.co/fixie-ai/ultravox-v0_6-llama-3_1-8b}{fixie-ai/ultravox-v0\_6-llama-3\_1-8b} & \\
        & ultravox-v0\_6-qwen-3-32b & 32B & 40K & \href{https://huggingface.co/fixie-ai/ultravox-v0_6-qwen-3-32b}{fixie-ai/ultravox-v0\_6-qwen-3-32b} & \\
    \bottomrule
    \end{tabularx}
    \caption{General information about audio modality baseline models.}
    \label{tab:baselines_models_audio}
\end{table*}
\begin{table*}[t]
    \setlength{\tabcolsep}{3pt}
    \centering
    \small
    \begin{tabularx}{\textwidth}{%
    @{}c%
    p{0.25\linewidth}%
    r%
    r%
    p{0.30\linewidth}%
    l%
    @{}l@{}}
        \toprule
        & \textbf{Model} & \textbf{Parameters} & \textbf{Context length} & \textbf{Hugging Face Hub link} & \textbf{Citation} \\
        \midrule
        & LLaVA-NeXT-Video-7B-hf & 7B & 4K & \href{https://huggingface.co/llava-hf/LLaVA-NeXT-Video-7B-hf}{llava-hf/LLaVA-NeXT-Video-7B-hf} & \multirow{1}{*}{\citet{liu2024llavanext}}\\
        \cmidrule{2-6}
        & Qwen2-VL-2B-Instruct & 2B & 32K & \href{https://huggingface.co/Qwen/Qwen2-VL-2B-Instruct}{Qwen/Qwen2-VL-2B-Instruct} & \multirow{3}{*}{\citet{wang2024qwen2vlenhancingvisionlanguagemodels}}\\
        & Qwen2-VL-7B-Instruct & 7B & 32K & \href{https://huggingface.co/Qwen/Qwen2-VL-7B-Instruct}{Qwen/Qwen2-VL-7B-Instruct} & \\
        & Qwen2-VL-72B-Instruct & 72B & 32K & \href{https://huggingface.co/Qwen/Qwen2-VL-72B-Instruct}{Qwen/Qwen2-VL-72B-Instruct} & \\
        \cmidrule{2-6}
        & Qwen2.5-VL-3B-Instruct & 3B & 128K & \href{https://huggingface.co/Qwen/Qwen2.5-VL-3B-Instruct}{Qwen/Qwen2.5-VL-3B-Instruct} & \multirow{3}{*}{\citet{bai2025qwen25vltechnicalreport}}\\
        & Qwen2.5-VL-7B-Instruct & 7B & 128K & \href{https://huggingface.co/Qwen/Qwen2.5-VL-7B-Instruct}{Qwen/Qwen2.5-VL-7B-Instruct} & \\
        & Qwen2.5-VL-72B-Instruct & 72B & 128K & \href{https://huggingface.co/Qwen/Qwen2.5-VL-72B-Instruct}{Qwen/Qwen2.5-VL-72B-Instruct} & \\
        \cmidrule{2-6}
        & Qwen3-VL-2B-Instruct & 2B & 262K & \href{https://huggingface.co/Qwen/Qwen3-VL-2B-Instruct}{Qwen/Qwen3-VL-2B-Instruct} & \multirow{2}{*}{\citet{yang2025qwen3technicalreport}}\\
        & Qwen3-VL-8B-Instruct & 9B & 262K & \href{https://huggingface.co/Qwen/Qwen3-VL-8B-Instruct}{Qwen/Qwen3-VL-8B-Instruct} & \\
        \cmidrule{2-6}
        & Qwen2.5-Omni-3B & 6B & 32K & \href{https://huggingface.co/Qwen/Qwen2.5-Omni-3B}{Qwen/Qwen2.5-Omni-3B} & \multirow{2}{*}{\citet{xu2025qwen25omnitechnicalreport}} \\
        & Qwen2.5-Omni-7B & 11B & 32K & \href{https://huggingface.co/Qwen/Qwen2.5-Omni-7B}{Qwen/Qwen2.5-Omni-7B} & \\
        \cmidrule{2-6}
        & MiniCPM-o-2\_6 & 9B & 32K & \href{https://huggingface.co/openbmb/MiniCPM-o-2_6}{openbmb/MiniCPM-o-2\_6} & \multirow{1}{*}{\citet{yao2024minicpmvgpt4vlevelmllm}} \\
        \cmidrule{2-6}
        & InternVL3\_5-4B & 5B & 40K & \href{https://huggingface.co/OpenGVLab/InternVL3_5-4B}{OpenGVLab/InternVL3\_5-4B} & \multirow{2}{*}{\citet{wang2025internvl3}} \\
        & InternVL3\_5-2B-Instruct & 2B & 40K & \href{https://huggingface.co/OpenGVLab/InternVL3_5-2B-Instruct}{OpenGVLab/InternVL3\_5-2B-Instruct} & \\
        \cmidrule{2-6}
        & InternVL3-9B-Instruct & 9B & 32K & \href{https://huggingface.co/OpenGVLab/InternVL3-9B}{OpenGVLab/InternVL3-9B} & \multirow{1}{*}{\citet{zhu2025internvl3}} \\
        & InternVL3-9B & 9B & 32K & \href{https://huggingface.co/OpenGVLab/InternVL3-9B}{OpenGVLab/InternVL3-9B} & \multirow{1}{*}{\citet{zhu2025internvl3}} \\
        & InternVL3-2B & 1B & 40K & \href{https://huggingface.co/OpenGVLab/InternVL3-2B}{OpenGVLab/InternVL3-2B} & \\
    \bottomrule
    \end{tabularx}
    \caption{General information about video modality baseline models.}
    \label{tab:baselines_models_video}
\end{table*}

\begin{table*}
    \setlength{\tabcolsep}{1.5pt}
    \centering
    \scriptsize
    \small
    \begin{tabular}{l|c|cccccc}
        \toprule
        \textbf{Model} & \textbf{Total} & \textbf{LabTabVQA} & \textbf{RealVQA} & \textbf{ruCLEVR} & \textbf{ruCommonVQA} & \textbf{ruHHH-Image*} & \textbf{ruMathVQA} \\
        \midrule
        Human Baseline & {0.80} & {0.91} & {0.63} & {0.96} & {0.84} & {0.89} & {0.95} \\
        \midrule

        Qwen3-Omni-30B-A3B-Inst & \textbf{0.55} & \textbf{0.56} / \textbf{0.62} & \textbf{0.31} / 0.67 & \textbf{0.50} / \textbf{0.63} & \textbf{0.56} / \textbf{0.86} & \textbf{0.45} / \textbf{0.57} & 0.01 / \textbf{0.30} \\
        GPT 4 & 0.48 & 0.29 / 0.29 & 0.23 / \textbf{0.70} & 0.31 / 0.53 & 0.27 / 0.85 & 0.39 / 0.49 & 0.01 / 0.17 \\
        Qwen2.5-VL-72B-Inst & 0.41 & 0.53 / 0.60 & 0.21 / 0.65 & 0.40 / 0.58 & 0.43 / 0.80 & 0.35 / 0.44 & 0.02 / 0.23 \\
        Qwen2-VL-72B-Inst & 0.33 & 0.12 / 0.14 & 0.10 / 0.49 & 0.29 / 0.55 & 0.14 / 0.60 & 0.14 / 0.19 & 0.01 / 0.06 \\
        Qwen2.5-VL-32B-Inst & 0.31 & 0.47 / 0.57 & 0.16 / 0.50 & 0.33 / 0.58 & 0.30 / 0.76 & 0.26 / 0.40 & 0.00 / 0.12 \\
        Qwen2.5-VL-7B-Inst & 0.26 & 0.19 / 0.27 & 0.11 / 0.32 & 0.23 / 0.34 & 0.33 / 0.56 & 0.20 / 0.29 & 0.02 / 0.10 \\
        llava-next-110b-hf & 0.24 & 0.06 / 0.10 & 0.09 / 0.29 & 0.11 / 0.20 & 0.31 / 0.58 & 0.12 / 0.18 & 0.01 / 0.01 \\
        Phi-3.5-vision-inst & 0.23 & 0.12 / 0.18 & 0.03 / 0.10 & 0.08 / 0.16 & 0.21 / 0.40 & 0.19 / 0.26 & 0.01 / 0.02 \\
        llava-next-72b-hf & 0.23 & 0.09 / 0.11 & 0.07 / 0.29 & 0.09 / 0.19 & 0.21 / 0.49 & 0.11 / 0.15 & 0.00 / 0.00 \\
        Qwen2.5-Omni-7B & 0.23 & 0.11 / 0.18 & 0.08 / 0.35 & 0.13 / 0.32 & 0.23 / 0.53 & 0.14 / 0.23 & 0.03 / 0.08 \\
        Qwen2-VL-7B-Inst & 0.20 & 0.01 / 0.13 & 0.05 / 0.36 & 0.10 / 0.34 & 0.19 / 0.50 & 0.02 / 0.16 & 0.01 / 0.02 \\
        SmolVLM-Inst & 0.19 & 0.01 / 0.16 & 0.03 / 0.23 & 0.11 / 0.34 & 0.14 / 0.52 & 0.14 / 0.22 & 0.02 / 0.04 \\
        Qwen3-VL-8B-Inst & 0.19 & 0.00 / 0.07 & 0.11 / 0.45 & 0.14 / 0.33 & 0.23 / 0.59 & 0.05 / 0.15 & 0.04 / 0.10 \\
        Phi-4-multimodal-inst & 0.18 & 0.26 / 0.35 & 0.11 / 0.30 & 0.09 / 0.20 & 0.07 / 0.36 & 0.04 / 0.08 & 0.04 / 0.07 \\
        MiniCPM-o-2\_6 & 0.18 & 0.04 / 0.06 & 0.08 / 0.31 & 0.15 / 0.33 & 0.26 / 0.50 & 0.06 / 0.09 & 0.00 / 0.00 \\
        Qwen2.5-Omni-3B & 0.18 & 0.04 / 0.14 & 0.07 / 0.25 & 0.16 / 0.29 & 0.22 / 0.48 & 0.04 / 0.21 & 0.01 / 0.04 \\
        InternVL3-9B & 0.17 & 0.01 / 0.04 & 0.04 / 0.27 & 0.19 / 0.37 & 0.09 / 0.45 & 0.00 / 0.07 & 0.01 / 0.02 \\
        Qwen2-VL-2B-Inst & 0.17 & 0.00 / 0.09 & 0.06 / 0.28 & 0.11 / 0.32 & 0.12 / 0.44 & 0.03 / 0.09 & 0.04 / 0.05 \\
        gemma-3-27b-it & 0.15 & 0.00 / 0.01 & 0.05 / 0.38 & 0.04 / 0.17 & 0.07 / 0.44 & 0.00 / 0.02 & 0.01 / 0.04 \\
        granite-vision-3.3-2b & 0.14 & 0.00 / 0.08 & 0.01 / 0.10 & 0.00 / 0.08 & 0.11 / 0.20 & 0.08 / 0.20 & 0.02 / 0.02 \\
        Qwen2.5-VL-3B-Inst & 0.14 & 0.02 / 0.11 & 0.03 / 0.13 & 0.13 / 0.26 & 0.15 / 0.32 & 0.03 / 0.18 & 0.00 / 0.04 \\
        Qwen3-VL-2B-Inst & 0.12 & 0.00 / 0.04 & 0.06 / 0.35 & 0.11 / 0.27 & 0.14 / 0.47 & 0.05 / 0.08 & 0.01 / 0.04 \\
        llava-1.5-13b-hf & 0.12 & 0.01 / 0.16 & 0.01 / 0.10 & 0.01 / 0.08 & 0.10 / 0.33 & 0.02 / 0.26 & 0.01 / 0.01 \\

    \bottomrule
    \end{tabular}
    \caption{\textbf{Image} modality evaluation results (6 tasks out of 11). All tasks metrics are \texttt{Exact Match   /   Judge Score}. For \textit{ruHHH-Image} dataset the metrics are \texttt{Group Exact Match   /   Group Judge Score}. For Human Baseline aggregated results are provided (average of EM and JudgeScore).}
    \label{tab:baselines_image}
\end{table*}
\begin{table*}
    \setlength{\tabcolsep}{2pt}
    \centering
    \scriptsize
    \small
    \begin{tabular}{l|c|ccccc}
        \toprule
        \textbf{Model} & \textbf{Total} & \textbf{ruNaturalScienceVQA} & \textbf{SchoolScienceVQA} & \textbf{UniScienceVQA} & \textbf{WEIRD} & \textbf{ruTiE-Image}\\
        
        \midrule
        Human Baseline & {0.80} & {0.99} & {0.82} & {0.13} & {0.85} & {0.77} \\
        \midrule

        Qwen3-Omni-30B-A3B-Inst & \textbf{0.55} & \textbf{0.77} / \textbf{0.85} & \textbf{0.64} / \textbf{0.70} & 0.11 / 0.34 & \textbf{0.70} / 0.78 & 0.63 / 0.64 \\
        GPT 4 & 0.48 & 0.64 / 0.69 & 0.59 / 0.66 & 0.10 / \textbf{0.40} & 0.69 / \textbf{0.79} & \textbf{0.70} / \textbf{0.73} \\
        Qwen2.5-VL-72B-Inst & 0.41 & 0.01 / 0.05 & 0.24 / 0.32 & 0.11 / 0.25 & 0.65 / 0.76 & 0.63 / 0.69 \\
        Qwen2-VL-72B-Inst & 0.33 & 0.25 / 0.40 & 0.54 / 0.68 & \textbf{0.18} / 0.33 & 0.31 / 0.48 & 0.65 / 0.69 \\
        Qwen2.5-VL-32B-Inst & 0.31 & 0.00 / 0.03 & 0.00 / 0.04 & 0.01 / 0.09 & 0.47 / 0.77 & 0.43 / 0.62 \\
        Qwen2.5-VL-7B-Inst & 0.26 & 0.13 / 0.17 & 0.16 / 0.32 & 0.06 / 0.13 & 0.36 / 0.62 & 0.24 / 0.47 \\
        llava-next-110b-hf & 0.24 & 0.25 / 0.34 & 0.25 / 0.45 & 0.06 / 0.13 & 0.24 / 0.33 & 0.54 / 0.55 \\
        Phi-3.5-vision-inst & 0.23 & 0.32 / 0.53 & 0.22 / 0.33 & 0.07 / 0.11 & 0.41 / 0.60 & 0.33 / 0.35 \\
        llava-next-72b-hf & 0.23 & 0.19 / 0.27 & 0.40 / 0.47 & 0.07 / 0.13 & 0.27 / 0.33 & 0.52 / 0.55 \\
        Qwen2.5-Omni-7B & 0.23 & 0.09 / 0.23 & 0.12 / 0.24 & 0.06 / 0.12 & 0.30 / 0.60 & 0.28 / 0.48 \\
        Qwen2-VL-7B-Inst & 0.20 & 0.04 / 0.38 & 0.03 / 0.52 & 0.09 / 0.19 & 0.08 / 0.38 & 0.11 / 0.58 \\
        SmolVLM-Inst & 0.19 & 0.19 / 0.36 & 0.11 / 0.27 & 0.06 / 0.10 & 0.36 / 0.44 & 0.14 / 0.24 \\
        Qwen3-VL-8B-Inst & 0.19 & 0.03 / 0.08 & 0.14 / 0.20 & 0.10 / 0.22 & 0.08 / 0.17 & 0.39 / 0.41 \\
        Phi-4-multimodal-inst & 0.18 & 0.05 / 0.12 & 0.30 / 0.40 & 0.08 / 0.15 & 0.02 / 0.07 & 0.44 / 0.46 \\
        MiniCPM-o-2\_6 & 0.18 & 0.07 / 0.19 & 0.20 / 0.35 & 0.01 / 0.10 & 0.20 / 0.24 & 0.34 / 0.41 \\
        Qwen2.5-Omni-3B & 0.18 & 0.07 / 0.23 & 0.09 / 0.24 & 0.06 / 0.12 & 0.14 / 0.60 & 0.13 / 0.36 \\
        InternVL3-9B & 0.17 & 0.17 / 0.25 & 0.37 / 0.44 & 0.06 / 0.16 & 0.01 / 0.16 & 0.21 / 0.40 \\
        Qwen2-VL-2B-Inst & 0.17 & 0.04 / 0.32 & 0.04 / 0.38 & 0.08 / 0.14 & 0.09 / 0.31 & 0.16 / 0.44 \\
        gemma-3-27b-it & 0.15 & 0.05 / 0.12 & 0.49 / 0.55 & 0.04 / 0.28 & 0.00 / 0.12 & 0.21 / 0.25 \\
        granite-vision-3.3-2b & 0.14 & 0.27 / 0.38 & 0.08 / 0.22 & 0.04 / 0.07 & 0.20 / 0.52 & 0.20 / 0.26 \\
        Qwen2.5-VL-3B-Inst & 0.14 & 0.02 / 0.26 & 0.04 / 0.22 & 0.05 / 0.11 & 0.09 / 0.50 & 0.08 / 0.35 \\
        Qwen3-VL-2B-Inst & 0.12 & 0.01 / 0.04 & 0.14 / 0.19 & 0.05 / 0.11 & 0.03 / 0.10 & 0.21 / 0.25 \\
        llava-1.5-13b-hf & 0.12 & 0.01 / 0.37 & 0.01 / 0.25 & 0.00 / 0.09 & 0.01 / 0.49 & 0.03 / 0.22 \\
        
    \bottomrule
    \end{tabular}
    \caption{\textbf{Image} modality evaluation results (5 tasks out of 11). All tasks metrics are \texttt{Exact Match   /   Judge Score}. For UniScienceVQA the Human Baseline is crowd, not an expert. For Human Baseline aggregated results are provided (average of EM and JudgeScore).}
    \label{tab:baselines_image1}
\end{table*}

\begin{table*}
    \setlength{\tabcolsep}{3pt}
    \centering
    \scriptsize
    \small
    \begin{tabular}{l|c|cccc}
        \toprule
        \textbf{Model} & \textbf{Total} & \textbf{AQUARIA} & \textbf{ruEnvAQA} & \textbf{ruTiE-Audio} & \textbf{ruSLUn} \\
        
        \midrule
        Human Baseline & {0.895} & {0.98} & {0.95} & {0.75} & {0.91} \\
        \midrule

        Qwen3-Omni-30B-A3B-Instruct & \textbf{0.56} & \textbf{0.69} / \textbf{0.77} & \textbf{0.70} / \textbf{0.78} & 0.43 / 0.44 & \textbf{0.39} / \textbf{0.28} \\
        Qwen2.5-Omni-7B & 0.47 & 0.55 / 0.67 & 0.55 / 0.70 & 0.36 / 0.41 & 0.37 / 0.18 \\
        Qwen2.5-Omni-3B & 0.38 & 0.41 / 0.58 & 0.36 / 0.64 & 0.30 / 0.36 & 0.35 / 0.04 \\
        MiniCPM-o-2\_6 & 0.37 & 0.40 / 0.55 & 0.47 / 0.64 & 0.25 / 0.32 & 0.31 / 0.01 \\
        ultravox-v0\_6-qwen-3-32b & 0.32 & 0.39 / 0.42 & 0.37 / 0.40 & \textbf{0.47} / \textbf{0.51} & 0.00 / 0.00 \\
        ultravox-v0\_5-llama-3\_1-8b & 0.31 & 0.30 / 0.37 & 0.34 / 0.41 & 0.29 / 0.31 & 0.31 / 0.15 \\
        ultravox-v0\_4\_1-llama-3\_1-8b & 0.31 & 0.29 / 0.36 & 0.35 / 0.42 & 0.31 / 0.32 & 0.28 / 0.11 \\
        ultravox-v0\_4 & 0.30 & 0.30 / 0.38 & 0.34 / 0.43 & 0.29 / 0.31 & 0.30 / 0.09 \\
        ultravox-v0\_6-llama-3\_1-8b & 0.30 & 0.29 / 0.36 & 0.26 / 0.40 & 0.30 / 0.32 & 0.31 / 0.16 \\
        ultravox-v0\_3 & 0.28 & 0.29 / 0.35 & 0.35 / 0.43 & 0.27 / 0.29 & 0.22 / 0.04 \\
        ultravox-v0\_4\_1-mistral-nemo & 0.26 & 0.18 / 0.34 & 0.27 / 0.42 & 0.18 / 0.34 & 0.26 / 0.14 \\
        audio-flamingo-3-hf & 0.26 & 0.21 / 0.42 & 0.41 / 0.59 & 0.15 / 0.28 & 0.00 / 0.00 \\
        Qwen2-Audio-7B-Instruct & 0.22 & 0.18 / 0.43 & 0.15 / 0.52 & 0.17 / 0.28 & 0.05 / 0.00 \\
        ultravox-v0\_2 & 0.14 & 0.01 / 0.25 & 0.00 / 0.38 & 0.01 / 0.25 & 0.23 / 0.00 \\
        ultravox-v0\_5-llama-3\_2-1b & 0.12 & 0.04 / 0.25 & 0.08 / 0.29 & 0.05 / 0.23 & 0.00 / 0.00 \\
        Qwen-Audio-Chat & 0.12 & 0.01 / 0.28 & 0.01 / 0.40 & 0.01 / 0.24 & 0.00 / 0.00 \\
        ultravox-v0\_3-llama-3\_2-1b & 0.12 & 0.06 / 0.24 & 0.08 / 0.29 & 0.05 / 0.20 & 0.00 / 0.00 \\
        SeaLLMs-Audio-7B & 0.10 & 0.08 / 0.14 & 0.02 / 0.20 & 0.14 / 0.23 & 0.00 / 0.01 \\
        
    \bottomrule
    \end{tabular}
    \caption{\textbf{Audio} modality evaluation results. All tasks metrics are \texttt{Exact Match  /  Judge Score}. For \textit{ruSLUn} dataset the metrics are \texttt{Intent Exact Match  /  Slot F1 Score}. For Human Baseline aggregated results are provided (average of EM and JudgeScore).}
    \label{tab:baselines_audio}
\end{table*}

\begin{table*}
    \setlength{\tabcolsep}{3pt}
    \centering
    \scriptsize
    \small
    \begin{tabular}{l|c|ccc}
        \toprule
        \textbf{Model} & \textbf{Total} & \textbf{CommonVideoQA} & \textbf{RealVideoQA} & \textbf{ruHHH-Video} \\
        
        \midrule
        Human Baseline & {0.92} & {0.96} & {0.96} & {0.84} \\
        \midrule

        Qwen2.5-VL-72B-Instruct & \textbf{0.63} & \textbf{0.57} / \textbf{0.64} & \textbf{0.65} / \textbf{0.73} & \textbf{0.53} / \textbf{0.63} \\
        Qwen3-VL-8B-Instruct & 0.58 & 0.53 / 0.60 & 0.61 / 0.69 & 0.46 / 0.56 \\
        Qwen2-VL-72B-Instruct & 0.56 & 0.49 / 0.62 & 0.57 / 0.70 & 0.34 / 0.62 \\
        Qwen2.5-VL-7B-Instruct & 0.52 & 0.49 / 0.56 & 0.58 / 0.67 & 0.41 / 0.44 \\
        Qwen2.5-Omni-7B & 0.44 & 0.42 / 0.55 & 0.44 / 0.61 & 0.24 / 0.39 \\
        Qwen2.5-VL-3B-Instruct & 0.43 & 0.39 / 0.49 & 0.47 / 0.60 & 0.25 / 0.37 \\
        Qwen3-VL-2B-Instruct & 0.42 & 0.41 / 0.48 & 0.51 / 0.59 & 0.20 / 0.31 \\
        Qwen3-Omni-30B-A3B-Instruct & 0.41 & 0.00 / 0.00 & 0.64 / 0.72 & 0.48 / \textbf{0.63} \\
        MiniCPM-o-2\_6 & 0.37 & 0.33 / 0.49 & 0.41 / 0.57 & 0.13 / 0.30 \\
        Qwen2.5-Omni-3B & 0.34 & 0.30 / 0.46 & 0.33 / 0.54 & 0.12 / 0.27 \\
        InternVL3-9B-Instruct & 0.32 & 0.27 / 0.30 & 0.25 / 0.32 & 0.33 / 0.44 \\
        InternVL3-9B & 0.32 & 0.28 / 0.31 & 0.26 / 0.31 & 0.32 / 0.42 \\
        Qwen2-VL-7B-Instruct & 0.30 & 0.09 / 0.50 & 0.12 / 0.61 & 0.04 / 0.45 \\
        InternVL3\_5-4B & 0.29 & 0.27 / 0.31 & 0.28 / 0.32 & 0.24 / 0.32 \\
        LLaVA-NeXT-Video-7B-hf & 0.13 & 0.09 / 0.21 & 0.09 / 0.24 & 0.03 / 0.09 \\
        
    \bottomrule
    \end{tabular}
    \caption{\textbf{Video} modality evaluation results. All tasks metrics are \texttt{Exact Match  /  Judge Score}. For \textit{ruHHH-Video} dataset the metrics are \texttt{Group Exact Match  /  Group Judge Score}. For Human Baseline aggregated results are provided (average of EM and JudgeScore).}
    \label{tab:baselines_video}
\end{table*}

The exact results of the finished submissions of the models from \autoref{tab:baselines_models_image} are presented in Tables  \ref{tab:baselines_image}, \ref{tab:baselines_image1}. The results of the models from \autoref{tab:baselines_models_audio} are presented in \autoref{tab:baselines_audio}. The evaluation results of the models from \autoref{tab:baselines_models_video} are stated in \autoref{tab:baselines_video}.

Evaluated on vision (image) modality models are strongest on natural-image semantics --- object/scene understanding, object functions, and everyday knowledge (\realvqa, \rucommonvqa, \weird, \rutieimage) and show decent spatial relations and multi-object reasoning (\realvqa, \ruclevr). Performance drops on diagrammatic/scientific QA with OCR (\unisciencevqa, \runaturalsciencevqa) and tables (\labtabvqa), exposing gaps in text extraction, scheme recognition, and cell-level grounding. Math (\rumathvqa) is brittle: EM lags JS, indicating partially correct solutions that miss the required final answer. Harder compositional and counterfactual reasoning (\realvqa, \rucommonvqa) still separates model tiers, especially smaller checkpoints. Overall, weaknesses concentrate in OCR/diagram/table parsing and structured compositional reasoning, while strengths lie in natural-image commonsense and basic spatial reasoning.

Moving to audio modality evaluations, the relative strengths cluster around acoustic scene understanding and temporal/comparative reasoning over environmental audio (\ruenvaqa), with partial competence in social/interaction cues and topic/scene grounding (\aquaria, \rutieaudio) captured by higher JS. Weaknesses are pronounced in speech recognition (\ruslun, \rutieaudio --- EM), speaker/turn handling (diarization), and format-faithful final answer extraction. Improving ASR robustness, speaker attribution, and constrained decoding or answer templates should convert many high-JS outputs into EM gains.

As for the video modality evaluations, the current relative strengths lie in scene/object recognition and short event recognition. Clear weaknesses persist in temporal perception/localization, action-sequence reasoning, mutual object localization, counting, and cause-and-effect—skills central to \commonvideoqa/\realvideoqa. Low \ruhhhvideo scores further reveal brittleness on ethics-aware interpretation and social context. Practically, improving temporal grounding (longer context windows or frame selection), causal reasoning, and answer-format control (constrained decoding/final-answer extraction) should convert many near-misses (high JS, low EM) into measurable EM gains.

\subsection{Human Baseline Details}
\label{sec:appendix_hb}
Human baseline values were obtained by evaluating the aggregate responses of annotators on control tasks. 
Prior to metric calculation, we conducted an additional identification of annotators who performed labeling tasks with low quality. To identify such annotators during the labeling process, we employed control tasks with automated correctness checking; incorrect answers reduced the annotator's skill score. During post-processing, we filtered out annotators who consistently made errors and whose responses did not align with the majority vote across all their submissions.

To evaluate the average human level we collected crowd-source\footnote{Annotation provided by ABC Elementary platform \url{https://app.elementary.center}}.

For expert tasks, in addition to the basic human baseline, we also performed expert annotation. We aggregate experts' answers with an overlap of 3 for expert human baseline scores \footnote{The observed quantity is a consequence of the limited pool of domain experts in niche specializations, who also had no prior involvement in the annotation of the datasets.}. The experts annotating answers for the human baseline had not been involved in the original dataset creation.

\begin{table*}[ht!]
    \centering
    \small
    \begin{tabular}{@{}lp{0.3\linewidth}p{0.08\linewidth}p{0.09\linewidth}p{0.1\linewidth}p{0.08\linewidth}p{0.08\linewidth}p{0.05\linewidth}@{}}
        \toprule
         & \textbf{Dataset} & \textbf{Overlap} & \textbf{Num samples} & \textbf{Total, \$} & \textbf{Per item, \$} & \textbf{Per hour, \$} & \textbf{IAA}\\
        \midrule
         & {\aquaria} & {5} & {786} & {324.42} & {0.41} & {7.38} & {93.87\%}\\
         & {\commonvideoqa} & {5} & {1200} & {2364.12} & {1.97} & {8.53} & {92.41\%}\\
         & {\labtabvqa} & {5} & {339} & {518.99} & {0.31} & {11.04} & {87.91\%}\\
         & {\realvqa} & {5} & {1 010} & {405.82} & {0.40} & {8.54} & {69.65\%}\\
         & {\realvideoqa} & {5} & {671} & {785.54} & {1.17} & {8.53} & {92.19\%}\\
         & {\ruclevr} & {5} & {2 063} & {1 440.15} & {0.70} & {7.40} & {93.36\%}\\
         & {\rucommonvqa} & {5} & {2 922} & {10 401.71} & {3.56} & {8.54} & {79.24\%}\\
         & {\ruenvaqa} & {5} & {644} & {239.58} & {0.37} & {7.38} & {89.46\%}\\
         & {\ruhhhimage} & {5} & {610} & {276.94} & {0.45} & {8.54} & {90.25\%}\\
         & {\ruhhhvideo} & {5} & {911} & {638.40} & {0.70} & {708} & {91.28\%}\\
         & {\rumathvqa (crowd)} & {5} & {2 975} & {214.71} & {0.07} & {1.10} & {85.01\%}\\
         & {\rumathvqa (expert)} & {5} & {2 975} & {1 363.58} & {2.29} & {1.35} & {83.09\%}\\
         & {\runaturalsciencevqa (crowd)} & {5} & {403} & {119.55} & {0.30} & {6.09} & {90.37\%}\\
         & {\runaturalsciencevqa (expert)} & {3} & {403} & {123.08} & {0.31} & {9.65} & {96.69\%}\\
         & {\ruslun} & {5} & {741} & {133.00} & {0.03} & {2.88} & {81.05\%}\\
         & {\rutieaudio} & {3} & {500} & {65.16} & {0.13} & {3.62} & {81.00\%}\\
         & {\rutieimage} & {3} & {500} & {65.16} & {0.13} & {3.62} & {85.33\%}\\
         & {\schoolsciencevqa (crowd)} & {5} & {1 750} & {1 293.83} & {0.74} & {8.54} & {67.56\%}\\
         & {\schoolsciencevqa (expert)} & {3} & {1 750} & {1 270.70} & {0.73} & {8.94} & {81.23\%}\\
         & {\unisciencevqa} & {5} & {1 150} & {102.4} & {0.09} & {2.41} & {45.19\%}\\
         & {\weird} & {5} & {889} & {109.1} & {0.10} & {9.4} & {90.84\%}\\
        \bottomrule
    \end{tabular}
    \caption{Payrates and total expenses for human baseline annotation.}
    \label{tab:human_baseline_expences}
\end{table*}

The ABC Elementary annotation platform ensures the necessary data anonymity during processing. The hourly compensation offered is above the minimum wage per hour in Russia (see \autoref{tab:human_baseline_expences}). Annotators are made aware of potentially sensitive topics within the data, including politics, societal minorities, and religion. The data collection process undergoes a mandatory quality evaluation, featuring an automated annotation quality check through honeypot tasks.

\end{document}